\documentclass[11pt]{article}

\usepackage[preprint]{acl}

\usepackage{times}
\usepackage{latexsym}
\usepackage[T1]{fontenc}
\usepackage[utf8]{inputenc}
\usepackage{microtype}
\usepackage{inconsolata}
\usepackage{fvextra}
\usepackage{graphicx}
\usepackage{booktabs}
\usepackage{amsmath}
\usepackage{amssymb}
\usepackage{multirow}
\usepackage{array}
\usepackage{xcolor}
\usepackage{colortbl}
\usepackage{subcaption}
\usepackage{pifont}

\newcommand{\best}[1]{\textbf{#1}}
\newcommand{\snd}[1]{\underline{#1}}

\DefineVerbatimEnvironment{PromptBlock}{Verbatim}{
  fontsize=\footnotesize,
  breaklines=true,
  breakanywhere=true,
  breaksymbolleft={},
  breaksymbolright={},
  xleftmargin=0pt
}

\graphicspath{{figures/}}

\title{DeskCraft: Benchmarking Desktop Agents on Professional Workflows and Human-in-the-Loop Collaboration}
\author{
  Wenkai Wang$^{1,\ast}$,
  Tao Xiong$^{1,\ast}$,
  Jingchen Ni$^{2,\ast}$,
  Yunpeng Bao$^{1,\ast}$, \\
  Xiyun Li$^{3}$,
  Tianqi Liu$^{1}$,
  Hongcan Guo$^{4}$,
  Zilong Huang$^{3}$,
  Shengyu Zhang$^{1,\dagger}$ \\[0.5ex]
  $^{1}$Zhejiang University \quad
  $^{2}$Tsinghua University \\
  $^{3}$Tencent \quad
  $^{4}$The University of Hong Kong \\[0.3ex]
  {\small $^{\ast}$Equal contribution. \quad $^{\dagger}$Corresponding author.}
}

\begin{document}
\maketitle
%
\begin{abstract}
    Real-world professional desktop workflows in specialized creative and engineering software unfold over long horizons and often require human-in-the-loop coordination, where agents proactively seek necessary information and users provide additional instructions, clarifications, feedback, or corrections as the task progresses.
    Yet existing desktop GUI benchmarks mostly reduce this setting to short, simplified tasks with all user instructions provided upfront.
    To address this issue, we introduce \textbf{DeskCraft}, a desktop GUI
    benchmark targeting long horizon creative and engineering workflows and proactive human-agent
    collaboration. DeskCraft organizes tasks into a multilevel difficulty taxonomy, with long horizon
    tasks requiring over 50 execution steps, and covers professional creative software across design, video, audio, and 3D creation.
    Furthermore, DeskCraft formalizes human-agent collaboration into an interaction protocol covering \emph{mid-turn} and \emph{post-turn} exchanges.
    Mid-turn interaction captures both agent-initiated clarification 
    under uncertainty and user-initiated interruption during execution, 
    while post-turn interaction accommodates user-driven feedback after the agent   
    signals completion, together spanning the full space of realistic collaboration patterns.
    We evaluate 18 proprietary and open source agents on 538 tasks and find that GPT-5.4 reaches \textbf{31.6\%} on standard tasks and \textbf{27.6\%} on interactive tasks. 
    Further analyses reveal persistent failures in long horizon workflow delivery and proactive clarification. We will open-source all evaluation codes, tasks, and data at \url{https://github.com/mrwwk/DeskCraft}.
\end{abstract}

\section{Introduction}
\label{sec:introduction}

%
%

Frontier multimodal models, such as GPT-5~\citep{openai2025gpt5} and Claude~\citep{anthropic2025claude}, now demonstrate strong capabilities in screen understanding and GUI operation~\citep{ui_tars_2025,agashe2025agents2,wang2025opencua}. This progress points toward a future in which desktop agents can take over substantial portions of routine digital work for their users.

Real-world desktop productivity, however, requires capabilities that 
extend far beyond isolated GUI actions. Professional workflows span 
multiple applications and extended time horizons; a 3D rendering pipeline, 
for instance, transitions from modeling to lighting, rendering, and 
compositing across various tools. Throughout these processes, the user 
iteratively directs the workflow via clarification, correction, and feedback. 
In tandem, the agent must proactively elicit missing information rather 
than relying on assumptions \cite{horvitz1999principles, allen1999mixed}. Deployable desktop agents, therefore, must 
not only sustain long action sequences but also dynamically adapt to
 evolving user intents.

Existing desktop benchmarks \citep{xie2024osworld,bonatti2024winaa,yang2025macosworld} 
successfully evaluate agents in live virtual machines,  but their tasks are largely short, atomic, and specified by predetermined instructions, leaving sustained workflows 
and human-in-the-loop dialogue underexplored. Benchmarks with explicit user interaction are mainly developed for tool-use, enterprise workflows, and mobile assistants \citep{yao2024taubench,xu2024theagentcompany,mobileworld2025}.
In desktop workflows, agents must map each clarification or correction to the current GUI state, revise their plan, and continue from the work already completed.
Consequently, there remains a need for a desktop benchmark that evaluates such interactive, long horizon workflows in live environments.

\begin{figure*}[t]
    \centering
    \includegraphics[width=\textwidth]{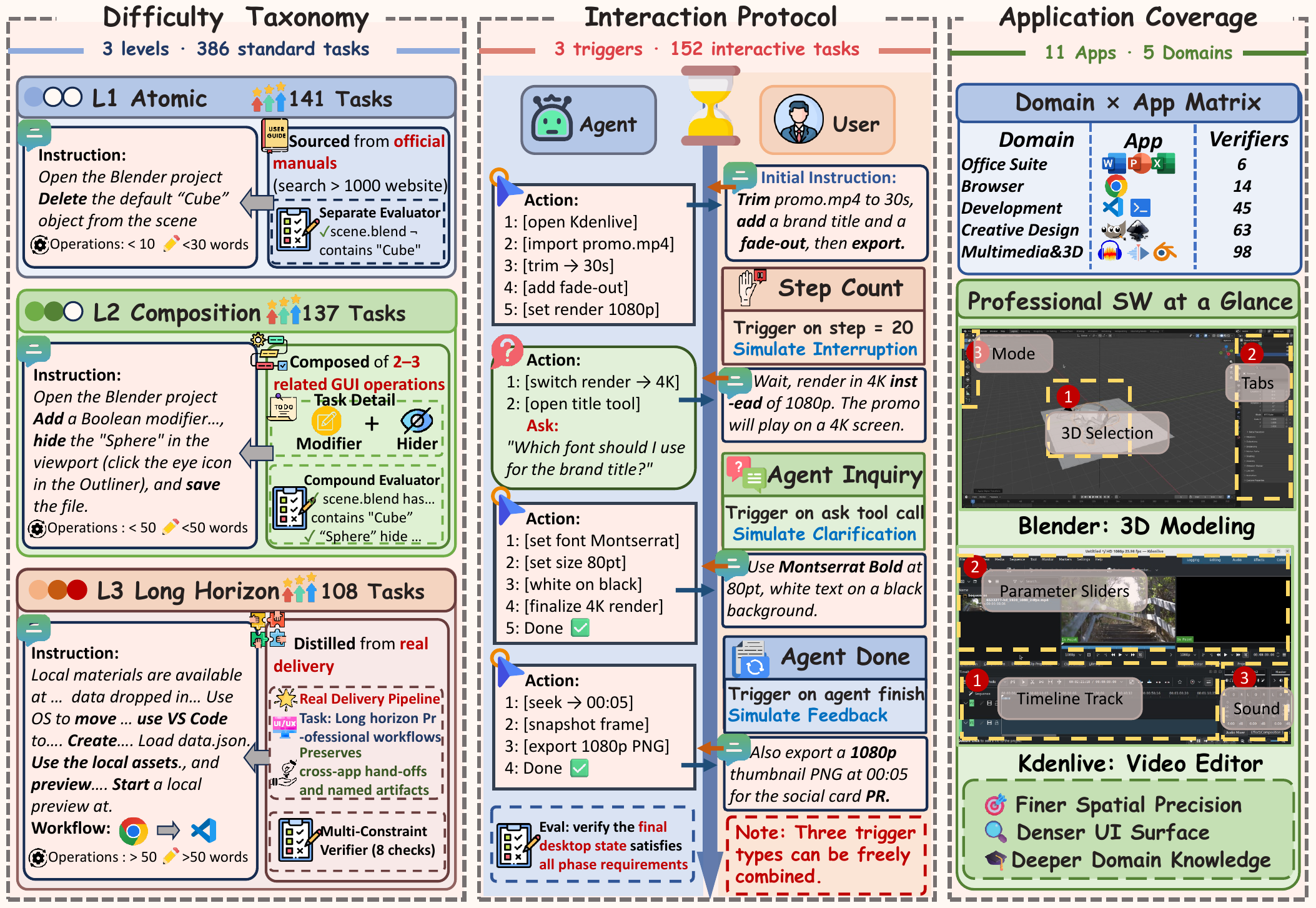}
    \caption{\textbf{Overview of DeskCraft}. \textbf{Left:} 386 standard tasks 
    stratified into L1 atomic, L2 compositional, and L3 long horizon levels,
     with L3 distilled from real delivery pipelines. \textbf{Middle:} 152 interactive 
     tasks driven by three composable triggers (\emph{step count}, \emph{agent inquiry}, 
     \emph{agent done}) that evolve a task through human-agent collaboration. 
     \textbf{Right:} 11 applications across 5 domains, including professional 
     software (e.g., Blender, Kdenlive) that demands finer spatial precision, 
     denser UI, and deeper domain knowledge than prior benchmarks.}
    \label{fig:overview}
\end{figure*}

To bridge this gap, we introduce \textbf{DeskCraft}
(Figure~\ref{fig:overview}), a 538-task desktop
benchmark designed to evaluate agents on long-horizon professional
workflows and human-agent interaction in live desktop environments.
DeskCraft contributes three design components.
\textbf{Diagnostic workflow difficulty.} Desktop tasks impose increasingly complex 
execution demands on GUI agents, ranging from following simple user instructions, 
to composing operations within a task, to sustaining long horizon workflows 
over many steps. DeskCraft defines this progression as an L1/L2/L3 difficulty 
taxonomy (\S\ref{sec:taxonomy}), enabling failures to be diagnosed by the level of 
execution demand they expose. In particular, L3 tasks are distilled from real 
professional scenarios, preserving the dependency structure of actual 
delivery processes rather than synthetically chaining independent operations.
\textbf{Human-agent interaction protocol.} Real desktop collaboration evolves as execution 
proceeds: users may revise goals, while agents 
 may need to request missing information or escalate risky decisions. DeskCraft formalizes this process through three trigger types covering mid-turn and post-turn
interaction. \emph{Mid-turn} triggers fire during execution and comprise two types: agent-initiated clarification and user-initiated interruption. \emph{Post-turn} trigger   
fires after the agent signals completion, allowing users to provide follow-up instructions. Together, these triggers capture a broad range of realistic human-agent collaboration patterns.  
\textbf{Broadened professional software coverage.} Prior
benchmarks concentrate on office suites, leaving professional
creative workflows underexplored. DeskCraft expands evaluation to image design, vector design, video editing, audio production, and 3D rendering, covering workflows that demand
spatial precision and domain-specific tool use.

\begin{table*}[t]
\centering
\begingroup
\footnotesize
\setlength{\tabcolsep}{4pt}
\renewcommand{\arraystretch}{1.06}
\newcommand{\Yes}{\textcolor{green!55!black}{\raisebox{0.1ex}{\large$\boldsymbol{\checkmark}$}}}
\newcommand{\No}{\textcolor{red!75!black}{\raisebox{0.1ex}{\large$\boldsymbol{\times}$}}}
\begin{tabular}{@{}llrcccc@{}}
\toprule
\textbf{Benchmark} & \textbf{Domain} & \textbf{\#Tasks} & \textbf{LH Focus} & \textbf{User Int.} & \textbf{Diff. Lvls.} & \textbf{Eval.} \\
\midrule
OSWorld \citep{xie2024osworld}           & Desktop (Ubuntu)  & 369    & \No       & \No         & \No       & End state \\
WAA \citep{bonatti2024winaa}             & Desktop (Windows) & 154    & \No       & \No         & \No       & End state \\
macOSWorld \citep{yang2025macosworld}    & Desktop (macOS)   & 202    & \No       & \No         & \No       & End state \\
WorldGUI \citep{worldgui2025}            & Desktop (Windows) & 611    & \No       & \No         & \No    & End state \\
VeriGUI \citep{verigui2025}              & Web               & 130    & \Yes      & \No         & \No       & Subtask \\
WebArena \citep{zhou2023webarena}        & Web               & 812    & \Yes      & \No         & \No       & Functional \\
TheAgentCo.\ \citep{xu2024theagentcompany} & Web (Enterprise) & 175 & \Yes   & NPC coworkers & \No     & Ckpt. \\
$\tau$-bench \citep{yao2024taubench}     & API + User        & 165    & \No       & LM user     & \No       & DB goal \\
MobileWorld \citep{mobileworld2025}      & Mobile            & 201    & \Yes      & User sim.   & \No       & State \\
GAIA \citep{mialon2023gaia}              & General QA        & 466    & \No       & \No         & \Yes      & Exact ans. \\
SWE-b.\ Pro \citep{deng2025swebenchpro} & Code              & 1{,}865 & \Yes      & \No         & \No       & Tests \\
\midrule
\textbf{DeskCraft (Ours)}                & \textbf{Desktop (Ubuntu)} & \textbf{538} & \textbf{\Yes} & User-in-loop & \textbf{\Yes} & End state \\
\bottomrule
\end{tabular}
\endgroup
\caption{\textbf{Comparison with representative agent benchmarks} along domain, scale, long horizon focus (LH Focus), user interaction form (User Int.), difficulty stratification (Diff.\ Lvls.), and evaluation granularity (Eval.). LH Focus is marked when multi-step workflows or cross-application dependencies are a central benchmark axis. DeskCraft is the first desktop benchmark to jointly support long horizon professional workflows, a human-in-the-loop protocol, and structured difficulty levels.}
\label{tab:benchmark_comparison}
\end{table*}

Across 538 tasks, the strongest model reaches only $33.8\%$ on standard tasks. On the interactive split, GPT-5.4 reaches $27.6\%$, while Kimi-K2.6 reaches $25.7\%$ under the 100-step setting. Further analysis shows that performance drops sharply on L3 
workflow-level artifact delivery, longer step budgets recover only a small tail 
of additional successes beyond 100 steps, and agents rarely seek clarification 
proactively. These results suggest that the dominant bottleneck has shifted 
from simple instruction execution to sustained workflow planning
 and proactive human-agent coordination.
Our contributions are as follows:
\begin{itemize}
    \item We introduce \textbf{DeskCraft}, a 538-task desktop
    benchmark with an L1/L2/L3 difficulty taxonomy and professional
    workflows spanning image and vector design, video editing, audio
    production, and 3D rendering.
    \item We propose a \textbf{human-agent interaction protocol}
    that models collaboration as phase-based task evolution driven by
    user feedback, agent information seeking, and execution progress.
    \item We evaluate \textbf{18 proprietary and open-source agents}, showing that 
    current models remain far from reliable, exhibit the largest gaps in L3 workflow 
    delivery and proactive clarification, and gain limited additional success
     from longer step budgets.
\end{itemize}

\section{Related Work}
\label{sec:related_work}

\paragraph{Desktop and Long Horizon Benchmarks.}
Desktop GUI benchmarks have established execution verified evaluation 
and expanded across platforms, action interfaces, initial state robustness, 
and professional software grounding~\citep{xie2024osworld,bonatti2024winaa,yang2025macosworld,osworldmcp2025,worldgui2025,screenspotpro2025,uivision2025}.
 However, they still largely focus on single instruction tasks, 
 leaving sustained workflows across multiple desktop applications 
 and user dialogue during execution underexplored. In parallel, 
 long horizon evaluation has advanced in web, GUI trajectory, and 
professional workplace settings, revealing persistent gaps in agents' 
 ability to complete multi step tasks~\citep{zhou2023webarena,verigui2025,xu2024theagentcompany}.
DeskCraft introduces a benchmark of long horizon professional desktop workflows that span multiple applications.

\paragraph{Interactive and Human-in-the-Loop Evaluation.}
Interactive agent evaluation has increasingly moved beyond static single-turn task completion,
 emphasizing dialogue, evolving user intent, and benchmark extensions along new evaluation 
axes~\citep{yao2024taubench,xu2024theagentcompany,mobileworld2025,mialon2023gaia,deng2025swebenchpro,multiswebench2025,zhang2025swebenchlive}.
However, these advances have only limited coverage in desktop environments, where most
benchmarks still evaluate agents under fixed task instructions without mid-execution user feedback~\citep{worldgui2025}.
DeskCraft introduces a Human-in-the-loop protocol for long horizon professional desktop workflows (Table~\ref{tab:benchmark_comparison}).



\section{DeskCraft Benchmark}
\label{sec:benchmark}

DeskCraft is an execution-based desktop benchmark targeting the joint
setting of long horizon workflows, user interaction, and professional
software tasks. This section specifies its task
formulation, L1/L2/L3 difficulty taxonomy, interaction protocol, and
evaluation procedure.

\subsection{Task Definition}
\label{sec:task_definition}
DeskCraft formulates GUI agent evaluation as a phase-conditioned
  control problem in a live desktop environment. A task is defined as
  \[
  \tau = (s_0,\; u_0,\; \Phi,\; \mathcal{E},\; R),
  \]
  where \(s_0\) is the initial desktop state, \(u_0\) is the user's
  instruction, \(\mathcal{E}\) is the desktop environment,
  \(\Phi=(\phi_1,\ldots,\phi_K)\) is an optional sequence of interaction
  phases (\S\ref{sec:interaction_protocol}), and \(R\) is the evaluation
  function. Each phase \(\phi_k=(u_k,g_k)\) pairs a follow-up user
  message with a trigger condition that determines when it is delivered.

  At each step, the agent observes a screenshot \(x_t\) and the active
  instruction, then selects
  \[
  a_t \in \mathcal{A} \cup \{\texttt{DONE},\; \texttt{ASK},\;
  \texttt{FAIL}\},
  \]
  where \(\mathcal{A}\) comprises GUI operations (clicks, keystrokes,
  scrolls). The episode ends when the agent emits \texttt{DONE} or
  \texttt{FAIL}, or when the step budget is reached; \texttt{ASK} does
  not terminate but may activate the next phase, updating the active
  instruction. Standard tasks set \(K{=}0\) (single fixed instruction);
  interactive tasks set \(K{>}0\), allowing the goal to evolve during
  execution. The final score \(R(s_T)\in\{0,1\}\) is computed from the
  resulting desktop state.
\subsection{Difficulty Taxonomy}
\label{sec:taxonomy}

DeskCraft categorizes standard desktop tasks by the execution capability required for success.
\textbf{L1} tasks consist of simple atomic operations, where the agent needs to perform one clearly specified GUI action.
\textbf{L2} tasks are built by composing related L1 tasks and typically involve 2-4 dependent GUI operations. 
\textbf{L3} tasks are long-horizon tasks that pursue a concrete high level objective through 
multiple interrelated subtasks. These tasks are crafted to resemble real world usage   
scenarios, avoiding trivial concatenation of L1-level atomic operations,
and each task is provided with multiple relevant resource files.

The difficulty distribution also varies across applications. Some newly introduced professional software domains
currently include more L1-style atomic tasks, whereas commonly applications contain a higher
proportion of L2/L3 tasks.

\begin{figure}[t]
  \centering
  \includegraphics[width=\columnwidth]{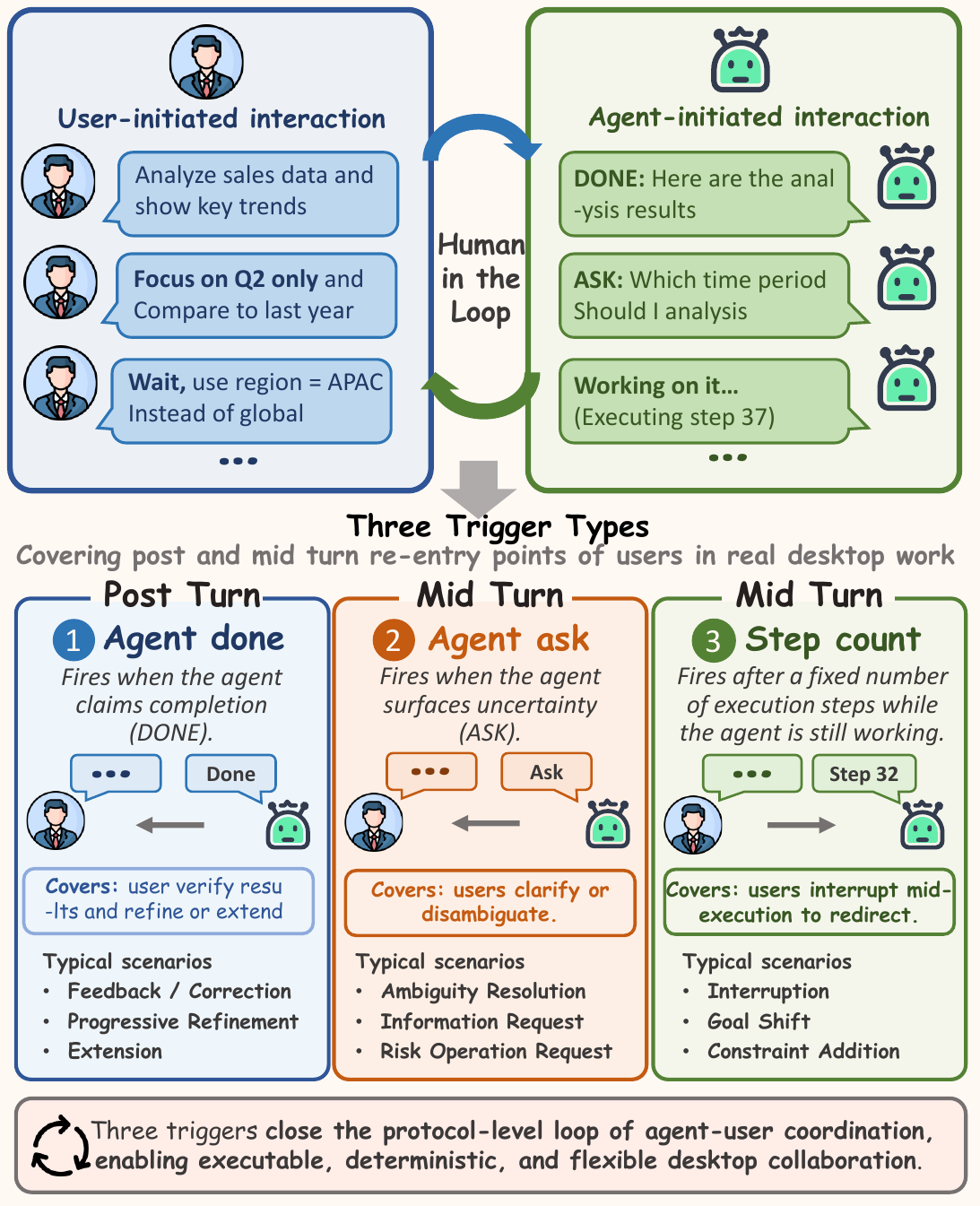}
  \caption{\textbf{DeskCraft interaction protocol.} Three composable triggers
  (\texttt{agent\_done}, \texttt{agent\_ask}, \texttt{step\_count}) define
  when the next user phase enters the session: after completion, on agent
  inquiry, or after a fixed step budget.}
  \label{fig:interaction_triggers}
\end{figure}

\subsection{Interaction Protocol}
\label{sec:interaction_protocol}

In real desktop work, users rarely fix a complete specification upfront; they 
clarify, interrupt, or revise as execution unfolds.  
Yet unconstrained dialogue makes evaluation hard to
reproduce. DeskCraft therefore represents interaction as an executable
phase protocol that captures goal evolution while keeping it
deterministic.

An interactive task consists of a sequence of phases
\(\Phi=(\phi_1,\ldots,\phi_K)\). Each phase \(\phi_k=(u_k,g_k)\)
contains a user message \(u_k\) and a trigger condition
\(g_k(\cdot)\in\{0,1\}\). When \(g_k\) fires, \(u_k\) is
appended to the interaction history and becomes the agent's active
instruction.

\paragraph{Triggers as a closed-loop minimal set.}

\begin{figure*}[t]
  \centering
  \begin{subfigure}[t]{0.31\textwidth}
    \centering
    \includegraphics[width=\linewidth]{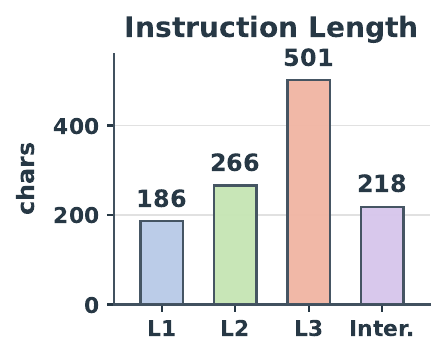}
    \caption{Instruction length.}
    \label{fig:difficulty_instruction_length}
  \end{subfigure}
  \hfill
  \begin{subfigure}[t]{0.31\textwidth}
    \centering
    \includegraphics[width=\linewidth]{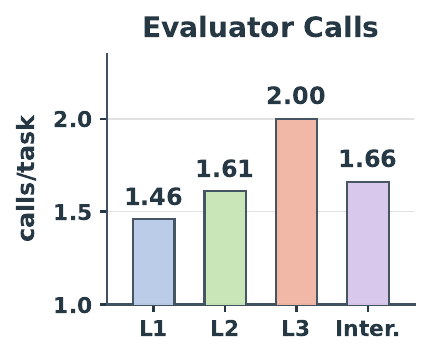}
    \caption{Evaluator calls.}
    \label{fig:difficulty_evaluator_calls}
  \end{subfigure}
  \hfill
  \begin{subfigure}[t]{0.31\textwidth}
    \centering
    \includegraphics[width=\linewidth]{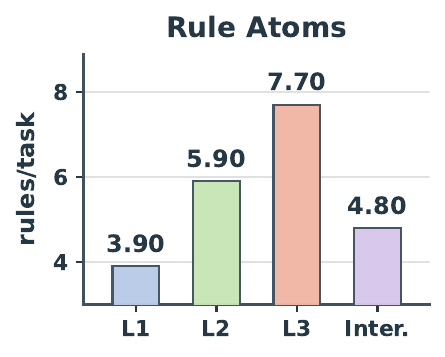}
    \caption{Rule atoms.}
    \label{fig:difficulty_rule_atoms}
  \end{subfigure}
  \caption{\textbf{Difficulty taxonomy statistics.}
  Although DeskCraft defines L1/L2/L3 by required execution capability rather than surface length, the levels align with measurable complexity: instruction length  
   and evaluator calls generally increase from L1 to L3. Some tasks use gold-file comparison for evaluation, involving only a single evaluator call and rule 
  regardless of task complexity. Interactive tasks are shown separately because their complexity is distributed across phase-level user messages. }
  \label{fig:difficulty_taxonomy_validation}
\end{figure*}
DeskCraft closes the human-agent interaction loop with a minimal set of three composable trigger types, covering mid-turn and post-turn interaction. For \emph{mid-turn} interaction, occurring while the agent is still executing:
\textbf{\texttt{agent\_ask}} fires when the agent emits \texttt{ASK} to solicit clarification,
and \textbf{\texttt{step\_count}} fires after a predetermined number of steps to model
user-initiated interruption. For \emph{post-turn} interaction, occurring after the agent signals completion:
\textbf{\texttt{agent\_done}} fires when the agent emits \texttt{DONE}, allowing the user to verify deliverables
and issue follow-up instructions or corrections (Figure~\ref{fig:interaction_triggers}).
Triggers compose freely within a task, enabling phase sequences that interleave them to produce realistic patterns such as
``clarify $\to$ interrupt $\to$ refine.'' Scenario families (ambiguity resolution, interruption, progressive refinement, feedback correction) are analysis
labels for the collaborative ability being tested, not additional trigger types.

\paragraph{User simulator.}
We employ an MLLM as a user simulator. When a predefined trigger fires, 
the simulator issues the next phase goal or responds to an unexpected \texttt{ASK} with   
 clarification, ensuring deterministic user interaction without trajectory drift. 
 If the agent has not completed the previous phase, the simulator still advances 
 to the next phase instruction to keep evaluation progressing; meanwhile, 
 the MLLM produces a judgment based on the current screenshot and agent output to    
 assess whether the previous phase was successfully completed. Whether a task ultimately succeeds is determined by the final desktop state.
 Full prompt template is given in Appendix~\ref{app:simulator}.


\subsection{Execution-Based Verification}
\label{sec:evaluation}

DeskCraft evaluates task success by verifying the resulting desktop state. 
We build a domain-aware verifier library for professional software. DeskCraft verifiers
extract structured state from project files or application runtimes 
and apply rule-based checks over the extracted fields, 
enabling deterministic evaluation of both long-horizon and interactive tasks. 
\section{Benchmark Construction}
\label{sec:construction}

We construct DeskCraft as 538 desktop tasks grounded in
realistic work and verified by automatic execution-based evaluators.
This section summarizes our task sourcing, difficulty annotation and
quality control, and dataset statistics.

\subsection{Task Sourcing}
\label{sec:sourcing}

For each of the 11 supported applications, we systematically collect operation
workflows from official documentation sites and online tutorials, yielding 224 
reference sources that collectively define a capability matrix of 120{+}      
operation categories. We sample tasks to ensure no two within the same application 
test the same atomic feature, producing 386 standard tasks backed by 300{+}
evaluator functions. L3 tasks follow a \emph{workflow distillation}
pipeline: we identify real professional workflows from documentation 
and tutorials, decompose each into a self-contained task 
with named inputs and a verifiable deliverable.

Across the full 538-task dataset, tasks use 279 unique asset files spanning 19 formats,
sourced through two channels: (1)~downloaded from public repositories; (2)~manually authored by annotators to fulfill specific task requirements.
The remaining 152 interactive tasks are derived by pairing selected L2/L3 workflows with typed triggers from the interaction protocol
 (\S\ref{sec:interaction_protocol}), covering both user-driven lifecycle management and agent-driven information acquisition.

\subsection{Evaluator Function Quality Control}
Each task comprises an instruction, a VM configuration, and an                                                                                                      
execution-based evaluator. For each application domain, practitioners                                                                                               
first draft a task design document specifying verification strategies; an LLM then generates
the evaluator functions; finally, a human and LLM dual review
checks the evaluator function correctness.



\subsection{Dataset Statistics}
\label{sec:statistics}

\begin{figure}[t]
\centering
\includegraphics[width=0.95\columnwidth]{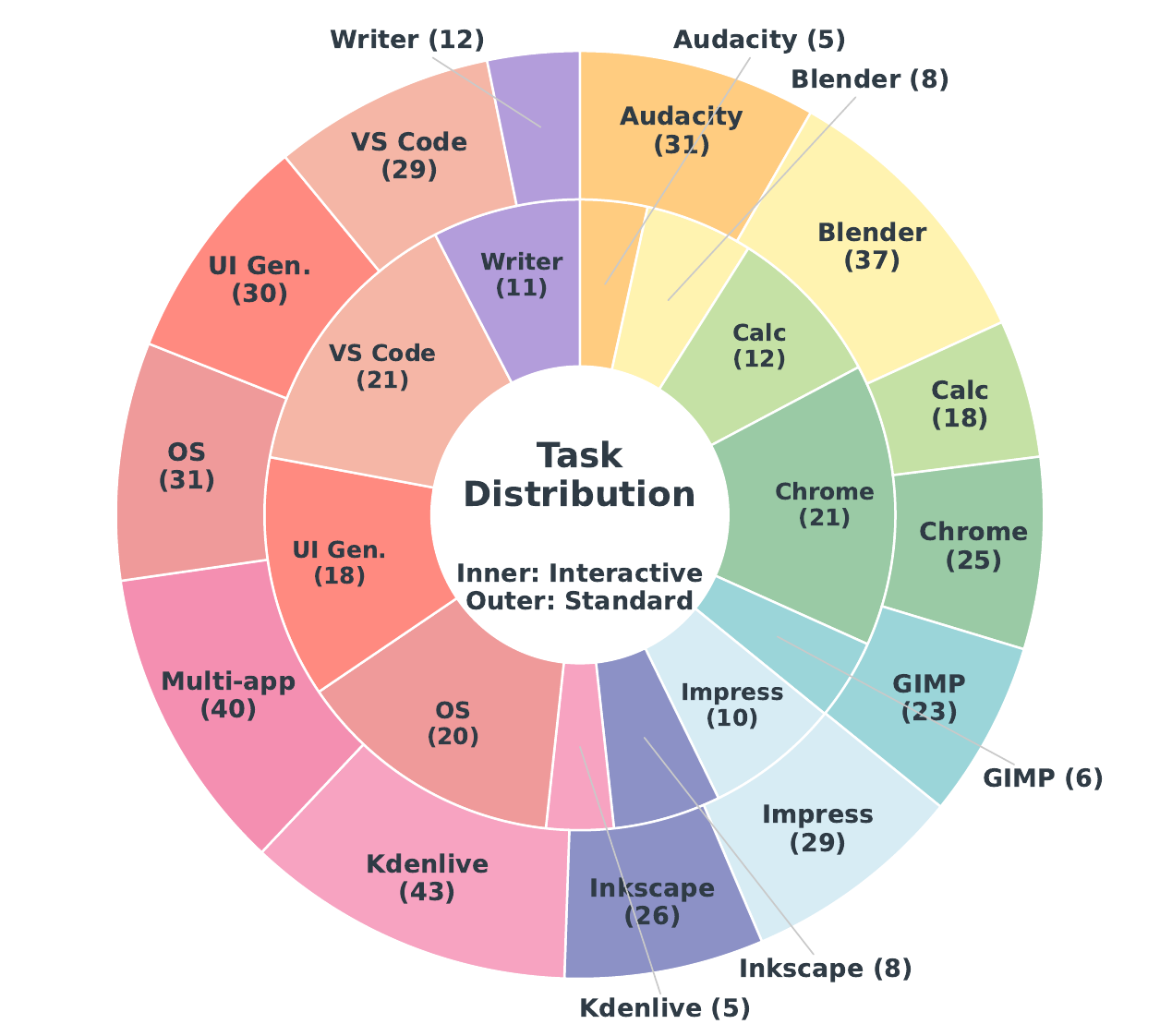}
\caption{\textbf{Per application task count} for the standard (outer ring) and                                                                                               
interactive (inner ring) splits, covering 11 applications and a multi-app                                                                                           
workflow category.}
\label{fig:software_distribution_pies}
\end{figure}

Figure~\ref{fig:software_distribution_pies} shows DeskCraft's task distribution                                                                                     
. The standard split is balanced                                                                                     
across L1/L2/L3 difficulty levels, which are defined by execution capability 
and correlate with measurable complexity signals: median instruction 
length rises from 186 to 501 characters,
average evaluator calls increase from 1.46 to 2.00, and average rule atoms
grow from 3.9 to 7.7 across levels
(Figure~\ref{fig:difficulty_taxonomy_validation}).
The interactive split contributes
403 phase-level user messages spanning scenario families such as progressive
refinement and requirement change.

\section{Experiment}
\label{sec:setup}

In this section, we conduct experiments to address the following research questions:
\begin{itemize}
    \item \textbf{RQ1:} How well do current GUI agents perform on professional desktop workflows under standard and interactive settings?
    \item \textbf{RQ2:} How much additional performance can a strong GUI agent recover under longer action horizons (300 steps)?
    \item \textbf{RQ3:} How do task success and execution length change as desktop workflows become more difficult from L1 to L3?
    \item \textbf{RQ4:} How well do current GUI agents collaborate with humans during interactive desktop workflows?
  \end{itemize}


\subsection{Experiment Settings}
\label{sec:agents}


\paragraph{Evaluated agents.}
We evaluate three families of models on DeskCraft:
(i) \textbf{proprietary frontier models} (GPT-5.4 \cite{openai2025gpt5}, Kimi-K2.6 \cite{team2026kimi});
(ii) \textbf{open-source generalist VLMs} (Qwen3-VL 8B/32B/235B-A22B \cite{ai2025qwen3},
Qwen3.5 9B/35B-A3B/397B-A17B \cite{qwen3.5}, Qwen3.6 35B-A3B \cite{qwen36_35b_a3b}); and
(iii) \textbf{open-source CUA foundation models} specialized for GUI use
(EvoCUA 8B/32B \cite{xue2026evocua}, GUI-Owl-1.5 8B/32B \cite{xu2026mobile}, OpenCUA 7B/32B \cite{wang2026opencua}, OS-Atlas-Pro 7B \cite{wu2025atlas}, UI-TARS 1.5 7B \cite{ui_tars_2025}).
This selection lets us compare proprietary frontier agents, open-source
generalist models, and GUI-specialized foundations while probing the roles
of model scale and domain-specific training in desktop agent performance.

For interactive tasks, we instantiate the simulator
 with \textbf{Kimi-K2.5} as a fixed
backbone across all evaluated agents. The full prompt template is given in
Appendix~\ref{app:simulator}.



\subsection{Overall Performance under Standard and Interactive Settings (RQ1)}
\label{sec:res_overall}
To answer RQ1, we evaluate current GUI agents on the \textbf{Standard} and
\textbf{Interactive} splits of DeskCraft. Table~\ref{tab:main_results}
reports per application and overall task-level success rates. We further
analyze repeated-run reliability for Kimi-K2.6 using, where pass@k counts a task as successful if any of $k$ runs succeeds, and pass$^k$ requires all $k$ runs to succeed.
We make the following observations:

%

\begin{table*}[t]
\centering
\begingroup
\footnotesize
\setlength{\tabcolsep}{3pt}
\renewcommand{\arraystretch}{1.10}

\newcolumntype{C}{>{\centering\arraybackslash}p{0.62cm}}  
\newcolumntype{A}{>{\columncolor{gray!8}\centering\arraybackslash}p{0.70cm}}  
\newcommand{\rh}[1]{\rotatebox{60}{\textbf{#1}}}          

\resizebox{\textwidth}{!}{%
\begin{tabular}{l | C C C C C C C C C C C C C | A | C C C C C C C C C C C C C | A}
\toprule
\multirow{2}{*}{\textbf{Agent}}
  & \multicolumn{14}{c|}{\textbf{Standard} (386 tasks)}
  & \multicolumn{14}{c}{\textbf{Interactive} (152 tasks)} \\
\cmidrule(lr){2-15}\cmidrule(lr){16-29}
 & \rh{Writer} & \rh{Calc} & \rh{Impress}
 & \rh{Chrome} & \rh{VS Code}
 & \rh{GIMP} & \rh{Inkscape}
 & \rh{Kdenlive} & \rh{Audacity} & \rh{Blender}
 & \rh{UI Gen}
 & \rh{Multi-app}
 & \rh{OS}
 & \textbf{Avg.}
 & \rh{Writer} & \rh{Calc} & \rh{Impress}
 & \rh{Chrome} & \rh{VS Code}
 & \rh{GIMP} & \rh{Inkscape}
 & \rh{Kdenlive} & \rh{Audacity} & \rh{Blender}
 & \rh{UI Gen}
 & \rh{Multi-app}
 & \rh{OS}
 & \textbf{Avg.} \\
\midrule

\rowcolor{gray!10}
\multicolumn{29}{l}{\textit{\textbf{Proprietary frontier models}}} \\
\midrule
GPT-5.4             & \best{50.0} & \snd{22.2} & \best{6.9} & \best{30.3} & \best{50.0} & \best{17.4} & \snd{50.0} & \best{27.9} & \snd{32.3} & \snd{54.1} & \snd{10.0} & \best{30.2} & \snd{32.3} & \snd{31.6}
                    & \best{81.8} & 0.0 & \best{10.0} & 0.0 & \best{25.0} & 0.0 & 0.0 & 0.0 & 20.0 & 0.0 & \snd{22.2} & \snd{48.0} & \best{73.3} & \best{27.6} \\
Kimi-K2.6           & 8.3 & \best{27.8} & \snd{3.7} & \snd{28.0} & \snd{37.9} & \snd{13.0} & \best{65.4} & \best{27.9} & \best{48.4} & \best{67.6} & \best{18.5} & \snd{25.6} & \best{41.4} & \best{33.8}
                    & \snd{54.5} & 0.0 & 0.0 & \snd{7.1} & \snd{12.5} & 0.0 & \best{28.6} & 0.0 & \best{40.0} & 0.0 & \best{33.3} & 44.0 & \snd{60.0} & \snd{25.7} \\
Kimi-K2.5           & \snd{33.3} & 11.1 & 3.4 & 0.0 & 20.0 & \snd{13.0} & 46.2 & \snd{25.6} & 25.8 & 43.2 & 6.7 & 9.3 & 29.0 & 20.3
                    & \best{81.8} & 0.0 & 0.0 & \best{9.1} & 4.8 & 0.0 & 0.0 & 0.0 & \snd{25.0} & 0.0 & 0.0 & \best{52.0} & 50.0 & 24.0 \\
\midrule

\rowcolor{gray!10}
\multicolumn{29}{l}{\textit{\textbf{Open-source generalist VLMs}}} \\
\midrule
Qwen3-VL-8B         & 25.0 & 11.1 & 3.4 & 0.0 & 0.0 & 0.0 & 0.0 & 2.3 & 0.0 & 2.7 & 6.7 & 5.0 & 0.0 & 3.2
                    & 27.3 & 0.0 & 0.0 & 0.0 & 0.0 & 0.0 & 0.0 & 0.0 & 0.0 & 0.0 & 0.0 & 0.0 & 0.0 & 2.1 \\
Qwen3-VL-32B        & 8.3 & 5.6 & 3.4 & 20.0 & 6.9 & 4.3 & 0.0 & 11.6 & 6.7 & 5.4 & 3.6 & 10.0 & 0.0 & 6.8
                    & \snd{54.5} & 0.0 & \best{10.0} & 0.0 & 4.8 & 0.0 & 0.0 & 0.0 & 0.0 & 0.0 & 0.0 & 0.0 & 10.0 & 6.9 \\
Qwen3-VL-235B-A22B  & 8.3 & 11.1 & 0.0 & 12.0 & 0.0 & 4.3 & 0.0 & 2.3 & 0.0 & 8.1 & 0.0 & 7.5 & 6.5 & 4.3
                    & 45.5 & 0.0 & 0.0 & 4.8 & 0.0 & 0.0 & 0.0 & 0.0 & \best{40.0} & 0.0 & 0.0 & 0.0 & 5.0 & 6.2 \\
Qwen3.5-9B          & 8.3 & 11.1 & 3.4 & 20.0 & 14.3 & 4.3 & 11.5 & 7.1 & 6.5 & 8.1 & 3.7 & 5.0 & 13.8 & 8.7
                    & 36.4 & 0.0 & 0.0 & 0.0 & 0.0 & 0.0 & 0.0 & 0.0 & 20.0 & 0.0 & 5.6 & 0.0 & 0.0 & 4.2 \\
Qwen3.5-35B-A3B     & 8.3 & 0.0 & \best{6.9} & 13.0 & 7.1 & 8.7 & 11.5 & 18.6 & 12.9 & 14.3 & 0.0 & 17.9 & 12.9 & 11.1
                    & 45.5 & 0.0 & 0.0 & 4.8 & 0.0 & 0.0 & 0.0 & 0.0 & 20.0 & 0.0 & \snd{22.2} & 0.0 & 35.0 & 12.4 \\
Qwen3.5-397B-A17B   & 8.3 & 5.6 & 3.4 & 20.0 & 10.3 & 4.3 & 11.5 & 16.3 & 9.7 & 18.9 & 4.5 & 15.0 & 26.7 & 12.9
                    & 36.4 & 0.0 & \best{10.0} & 4.8 & 4.8 & 0.0 & 0.0 & 0.0 & 20.0 & 0.0 & 11.1 & 0.0 & 35.0 & 11.7 \\
Qwen3.6-35B-A3B     & 8.3 & 0.0 & \best{6.9} & 12.0 & 6.9 & 0.0 & 19.2 & 18.6 & 3.2 & 18.9 & 0.0 & 10.0 & 16.7 & 10.2
                    & 45.5 & 0.0 & 0.0 & 0.0 & 0.0 & 0.0 & 0.0 & 0.0 & 0.0 & 0.0 & 5.6 & 0.0 & 20.0 & 6.9 \\
\midrule

\rowcolor{gray!10}
\multicolumn{29}{l}{\textit{\textbf{Open-source CUA foundation models}}} \\
\midrule
UI-TARS-1.5-7B      & 0.0 & 0.0 & 3.4 & 6.1 & 10.0 & 0.0 & 3.8 & 2.3 & 0.0 & 5.4 & 0.0 & 4.7 & 0.0 & 3.1
                    & 0.0 & 0.0 & 0.0 & 0.0 & 0.0 & 0.0 & 0.0 & 0.0 & 0.0 & 0.0 & 0.0 & 0.0 & 0.0 & 0.0 \\
EvoCUA-8B           & 8.3 & 5.6 & 3.4 & 15.2 & 13.3 & 0.0 & 7.7 & 2.3 & 0.0 & 5.4 & \snd{10.0} & 9.3 & 9.7 & 7.0
                    & 0.0 & 0.0 & 0.0 & 0.0 & 0.0 & 0.0 & 0.0 & 0.0 & 0.0 & 0.0 & 0.0 & 4.0 & 0.0 & 0.7 \\
EvoCUA-32B          & 0.0 & 5.6 & 3.4 & 24.2 & 13.3 & 0.0 & 23.1 & 18.6 & 6.5 & 10.8 & 0.0 & 9.3 & 19.4 & 11.4
                    & 0.0 & 0.0 & 0.0 & 0.0 & 0.0 & 0.0 & 0.0 & 0.0 & 0.0 & 0.0 & 0.0 & 0.0 & 0.0 & 0.0 \\
GUI-Owl-1.5-8B-Instruct& 0.0 & 0.0 & 0.0 & 6.1 & 3.3 & 4.3 & 3.8 & 4.7 & 0.0 & 10.8 & 0.0 & 7.0 & 6.5 & 4.1
                    & 0.0 & 0.0 & 0.0 & 0.0 & 0.0 & 0.0 & 0.0 & 0.0 & 0.0 & 0.0 & 0.0 & 8.0 & 0.0 & 1.3 \\
GUI-Owl-1.5-32B-Instruct& 0.0 & 0.0 & 3.4 & 3.0 & 6.7 & 4.3 & 0.0 & 9.3 & 3.2 & 8.1 & 0.0 & 4.7 & 3.2 & 4.1
                    & 0.0 & 0.0 & 0.0 & 0.0 & 0.0 & 0.0 & 0.0 & 0.0 & 0.0 & 0.0 & 0.0 & 4.0 & 0.0 & 0.7 \\
OpenCUA-7B          & 0.0 & 0.0 & 3.4 & 18.2 & 13.3 & 0.0 & 3.8 & 4.7 & 6.5 & 2.7 & 0.0 & 4.7 & 3.2 & 5.2
                    & 0.0 & 0.0 & 0.0 & 0.0 & 0.0 & 0.0 & 0.0 & 0.0 & 0.0 & 0.0 & 0.0 & 4.0 & 0.0 & 0.7 \\
OpenCUA-32B         & 0.0 & 0.0 & 3.4 & 9.1 & 16.7 & 8.7 & 11.5 & 14.0 & 12.9 & 13.5 & 0.0 & 9.3 & 12.9 & 9.6
                    & 0.0 & 0.0 & 0.0 & 0.0 & 0.0 & 0.0 & 0.0 & 0.0 & 0.0 & 0.0 & 0.0 & 0.0 & 0.0 & 0.0 \\
OS-Atlas-Pro-7B     & 0.0 & 5.6 & 0.0 & 0.0 & 0.0 & 0.0 & 0.0 & 0.0 & 0.0 & 0.0 & 0.0 & 0.0 & 0.0 & 0.3
                    & 0.0 & 0.0 & 0.0 & 0.0 & 0.0 & 0.0 & 0.0 & 0.0 & 0.0 & 0.0 & 0.0 & 0.0 & 0.0 & 0.0 \\
\bottomrule
\end{tabular}%
}
\endgroup
\caption{%
\textbf{Per-application success rate on DeskCraft.}
We report task success rate (SR, \%) for each agent on the
\textbf{Standard} split (386 tasks) and the \textbf{Interactive}
split (152 tasks). The two \textbf{Avg.}\ columns report overall
task-level SR within each regime. \best{Bold} = best per column;
\snd{underline} = runner-up.
}
\label{tab:main_results}
\end{table*}

\textbf{Obs.\ding{182}} \textbf{Current GUI agents achieve limited overall success on DeskCraft.}
The best average success rates remain below 35\%: 
Kimi-K2.6 achieves the highest Standard performance at 33.8\%,
while GPT-5.4 achieves the highest Interactive performance at 27.6\%.
GPT-5.4 reaches 31.6\% on Standard, and Kimi-K2.6 reaches 25.7\% on Interactive.
Most open-source generalist VLMs and GUI-specialized foundation models are substantially lower, indicating that
DeskCraft still leaves substantial room for improvement.

\begin{figure}[t]
    \centering
    \includegraphics[width=\columnwidth]{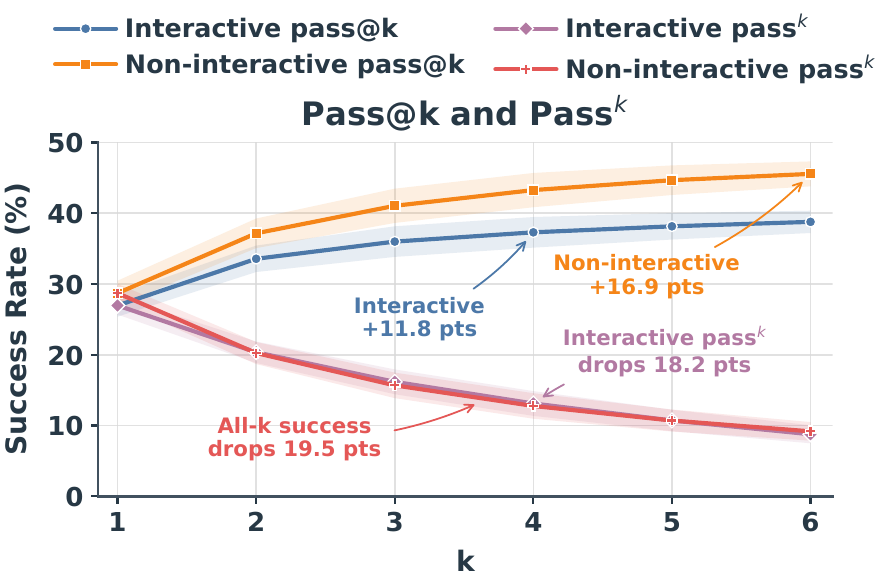}
    \caption{\textbf{Pass@k and pass$^k$ trends for Kimi-K2.6.}
    Pass@k evaluation requires multiple independent runs per task, we
    compute these metrics on \textbf{a subset of tasks.}
    Pass@k increases with larger $k$, whereas pass$^k$ decreases as the
    requirement shifts from at least one successful attempt to consistent
    success across all attempts.}
    \label{fig:kimi_k26_passk_line}
\end{figure}

\textbf{Obs.\ding{183}} \textbf{Multiple attempts raise upper-bound success, but run-to-run reliability remains weak.}
Figure~\ref{fig:kimi_k26_passk_line} shows that Kimi-K2.6 benefits from
multiple trials on both settings. Since pass@k requires repeated independent
rollouts for each task, we report pass@k and pass$^k$ on a representative task
subset. On this subset, Standard pass@k rises from 28.7\% at
$k{=}1$ to 45.6\% at $k{=}6$, and Interactive pass@k rises from 27.0\% to
38.8\%. However, pass$^k$ drops as $k$ increases. This gap
shows that current GUI agents often succeed only intermittently across
repeated executions of the same workflow, rather than solving it robustly.

\subsection{Long-Horizon 300-Step Budget Analysis (RQ2)}
\label{sec:res_step_budget}

\begin{figure}[t]
    \centering
    \includegraphics[width=\columnwidth]{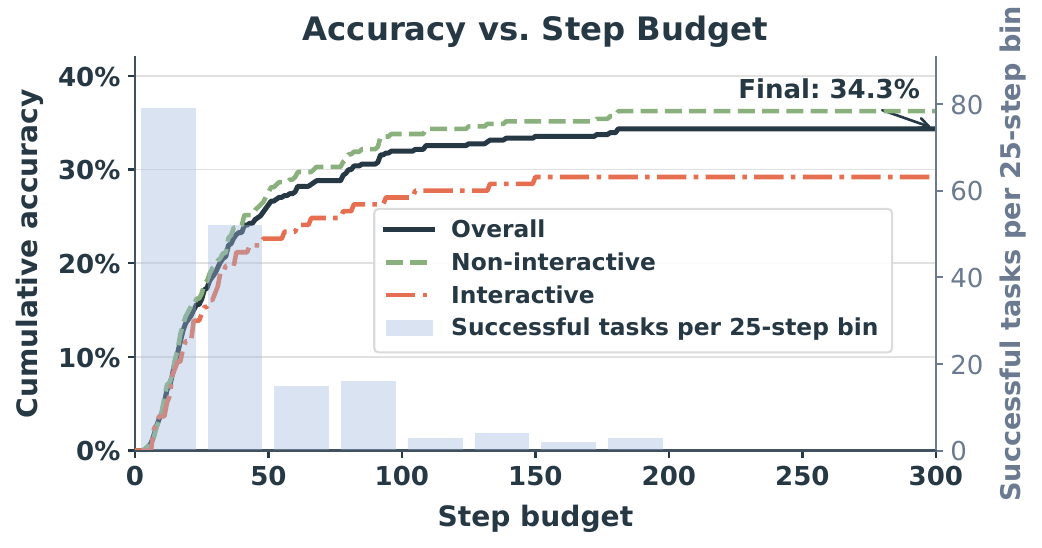}
    \caption{\textbf{Cumulative accuracy of Kimi K2.6 under increasing step budgets.}
    A task contributes to the accuracy at a given budget only if it is completed
    successfully within that number of steps.}
    \label{fig:kimi_k26_step_pareto}
\end{figure}

To answer RQ2, we analyze Kimi-K2.6 under progressively larger action
budgets. Figure~\ref{fig:kimi_k26_step_pareto} reports cumulative success:
a task is counted at a given budget only if the model completes it
successfully within that number of steps.

\textbf{Obs.\ding{184}} \textbf{Longer action budgets reveal additional capability beyond the
100-step regime.}
Kimi-K2.6 benefits substantially as the budget increases toward 100 steps:
overall success rises from 17.0\% at 25 steps to 34.3\% at 100 steps.
The model continues to complete some tasks after the conventional 100-step
horizon: standard success reaches 34.9\% at 150 steps and 35.7\% at 181 steps. In
absolute terms, the extended budget adds 13 more successful tasks after the
100-step point, including four tasks completed after
150 steps. No additional successful completion appears beyond 200 steps in
our run. These results suggest that sub-100-step evaluations can miss a
small but meaningful tail of long-horizon capabilities.

\subsection{Difficulty-Level Capability Degradation (RQ3)}
\label{sec:res_difficulty}
\begin{figure*}[t]
  \centering
  \begin{subfigure}[t]{0.31\textwidth}
    \centering
    \includegraphics[width=\linewidth]{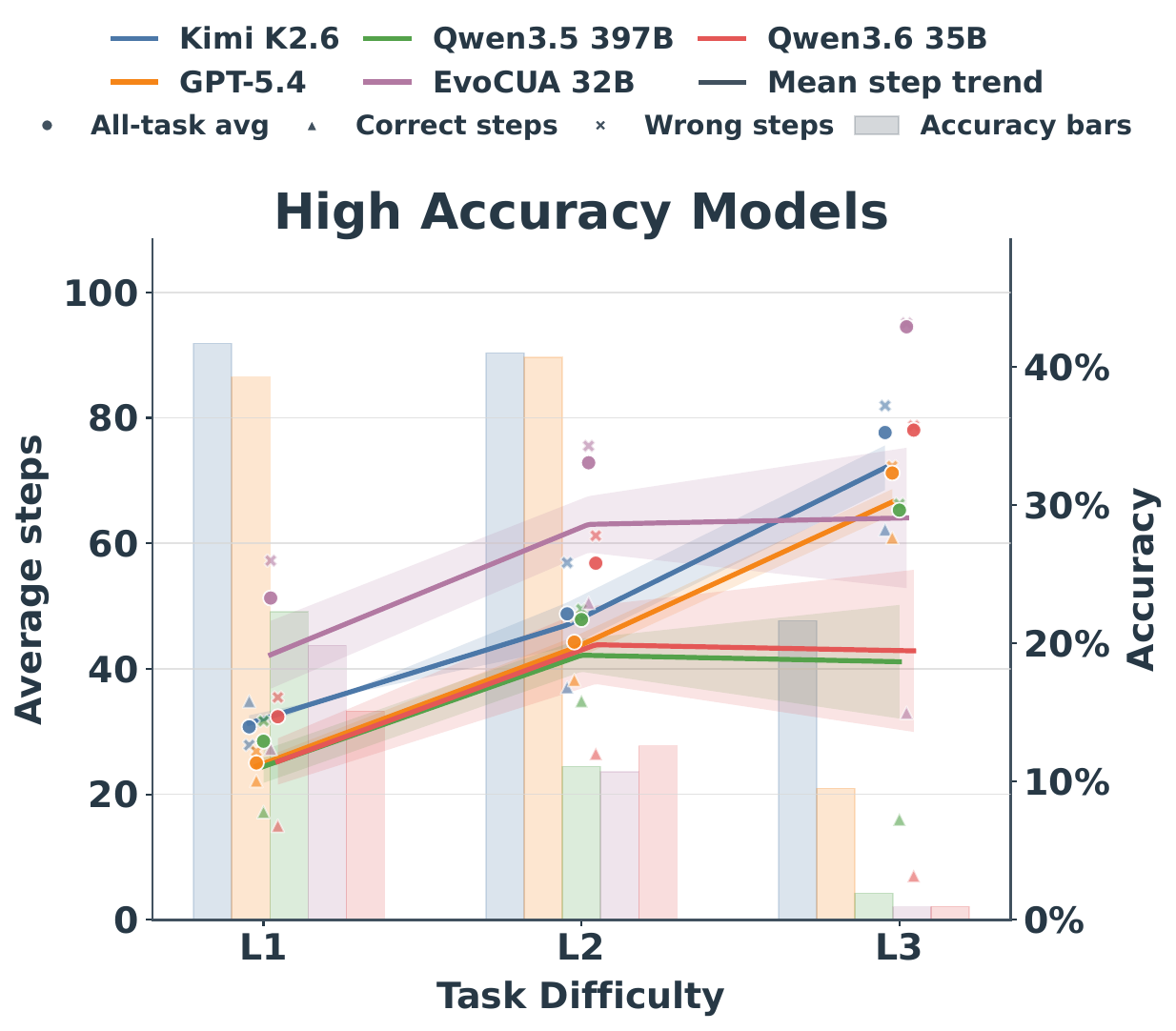}
    \caption{Leading models.}
    \label{fig:run_length_accuracy_high}
  \end{subfigure}
  \hfill
  \begin{subfigure}[t]{0.31\textwidth}
    \centering
    \includegraphics[width=\linewidth]{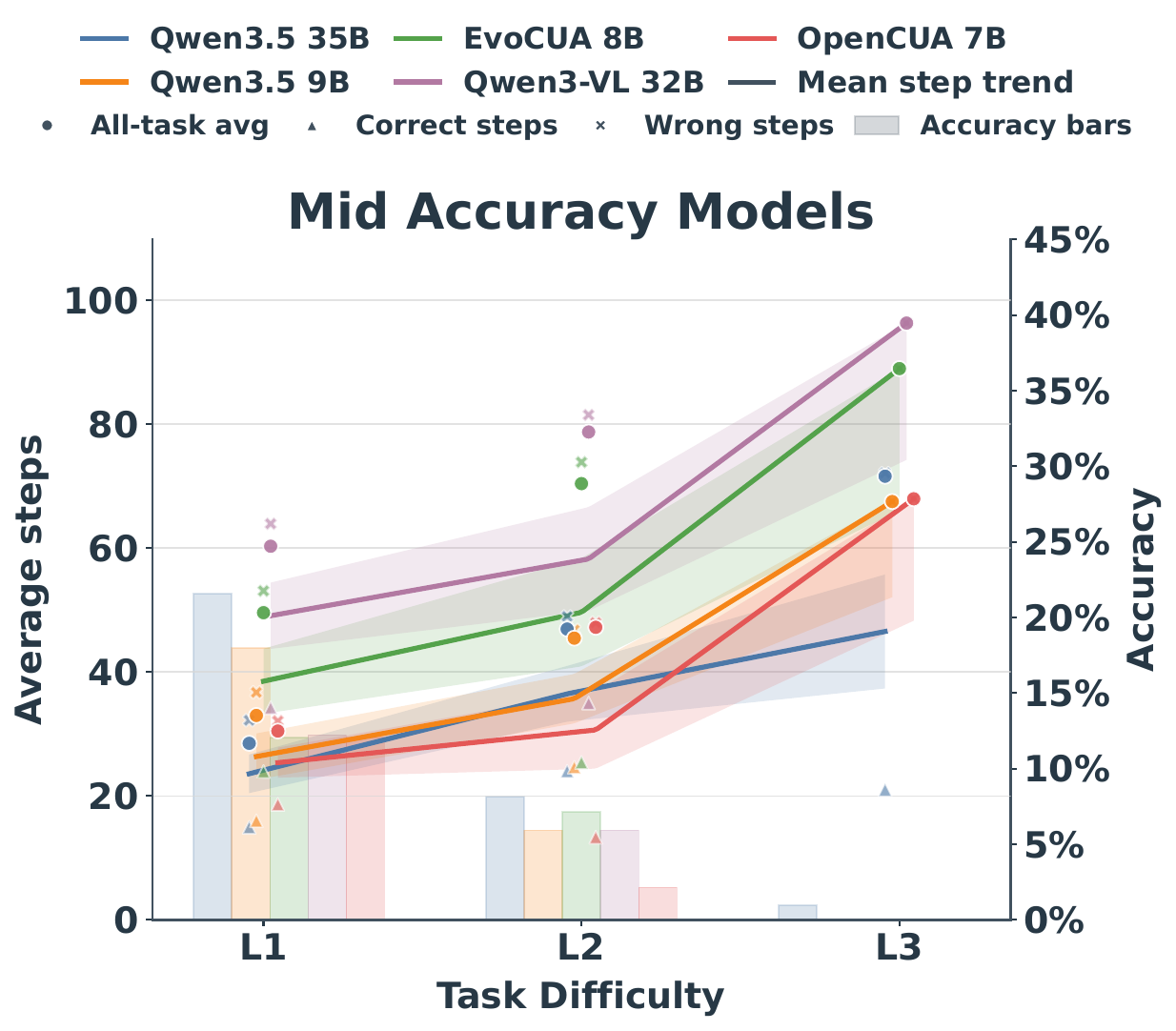}
    \caption{Competitive models.}
    \label{fig:run_length_accuracy_mid}
  \end{subfigure}
  \hfill
  \begin{subfigure}[t]{0.31\textwidth}
    \centering
    \includegraphics[width=\linewidth]{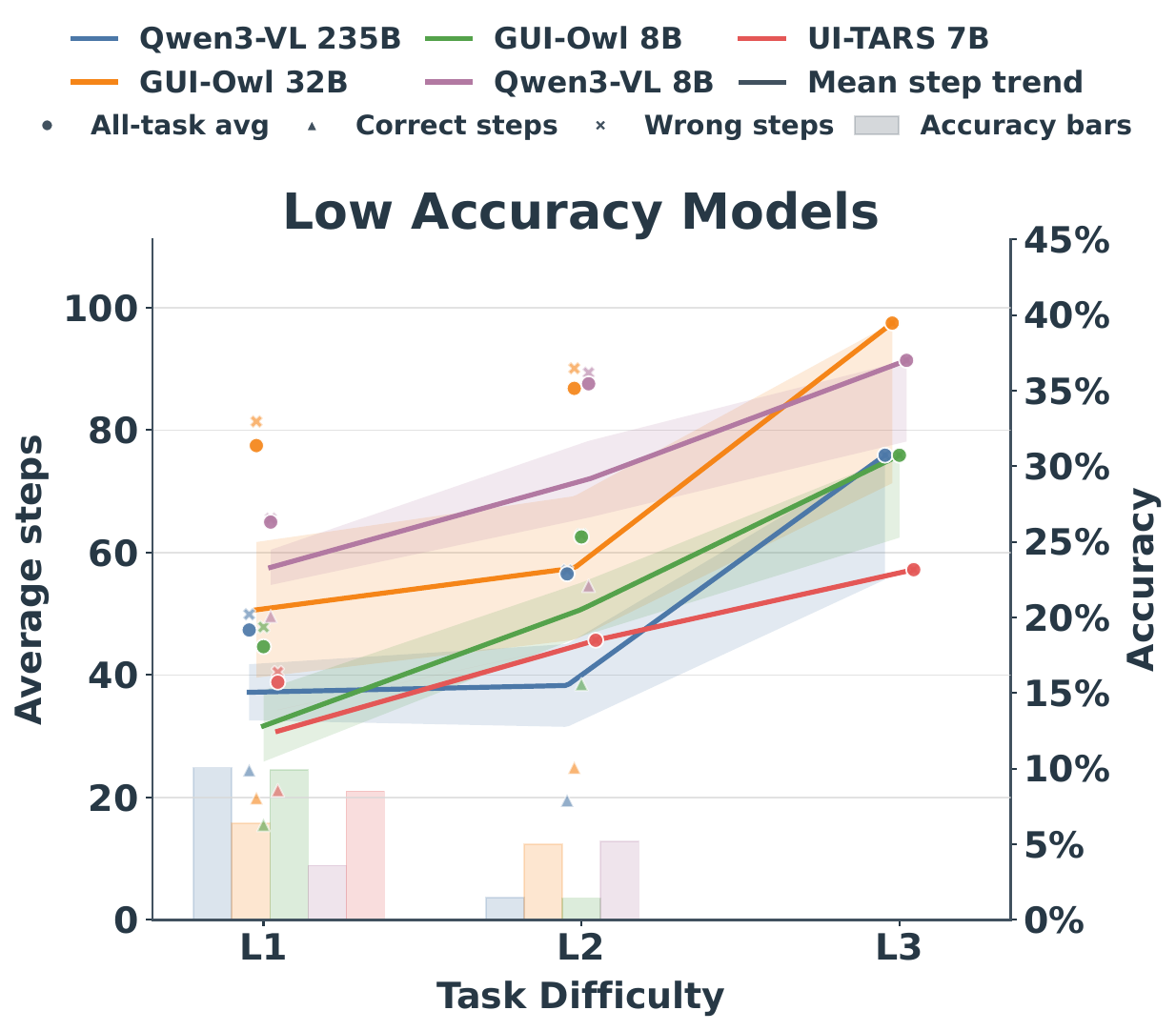}
    \caption{Emerging models.}
    \label{fig:run_length_accuracy_low}
  \end{subfigure}
  \caption{\textbf{Run-length and accuracy trends across L1, L2, and L3 tasks.} Lines show the mean of correct- and wrong-task step counts, markers show all/correct/wrong step averages, and bars show per-level accuracy.}
  \label{fig:difficulty_degradation_cua}
  \end{figure*}

  To answer RQ3, we analyze how performance changes across DeskCraft's
  three difficulty levels. Figure~\ref{fig:difficulty_degradation_cua}
  shows both success rates and run lengths across levels.

  \textbf{Obs.\ding{185}} \textbf{Accuracy drops as task difficulty increases.}
  Across model families, success rates decline from L1/L2 to L3, with the main
  cliff typically appearing at L3. For example, EvoCUA-32B drops from 19.9\%
  (L1) to 10.7\% (L2) and 1.0\% (L3). Stronger general-purpose agents also
  remain limited on L3: Kimi-K2.6 declines from 41.0\% (L2) to 21.6\% (L3), and
  GPT-5.4 from 40.7\% (L2) to 9.5\% (L3).

  \textbf{Obs.\ding{186}} \textbf{Higher difficulty is associated with longer runs.}
  Average run length generally increases from L1 to L3 for both successful and
  failed trajectories. For instance, Kimi-K2.6 rises from 30.8 steps on L1 to
  48.8 on L2 and 77.7 on L3, while GPT-5.4 rises from 25.0 to 44.3 to 71.2.
  This suggests that harder desktop workflows are not only less accurate, but
  also less efficient, reflecting persistent weaknesses in long-horizon planning
  and state management.

\subsection{Human-in-the-Loop Collaboration Analysis (RQ4)}
\label{sec:res_hitl_collaboration}

To answer RQ4, we group Interactive tasks by their
human-in-the-loop collaboration mode and compare task success rates for
Kimi-K2.6 and GPT-5.4. The full label distribution is reported in
Appendix~\ref{app:hitl_modes}.

\begin{figure}[t]
    \centering
    \includegraphics[width=\columnwidth]{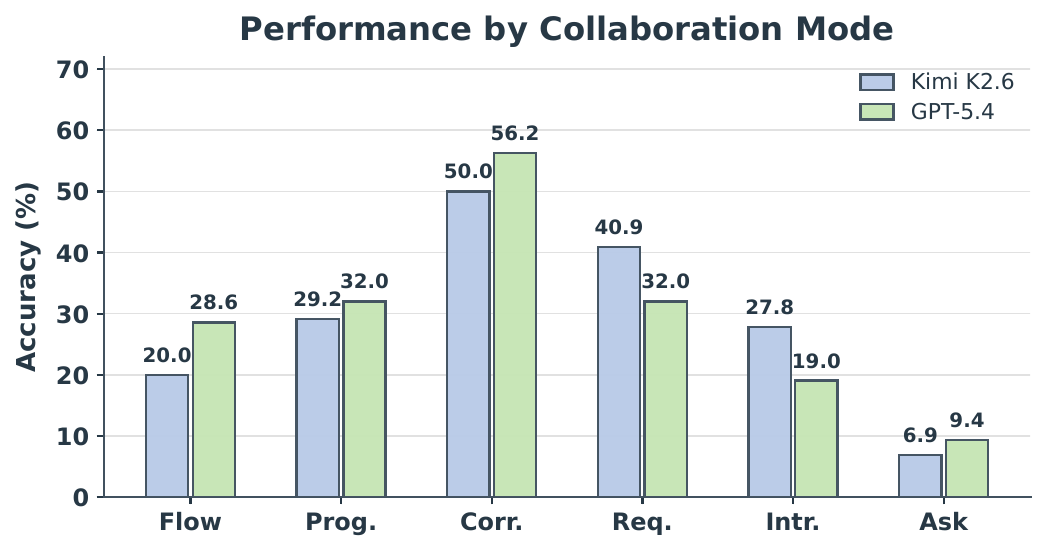}
    \caption{\textbf{Success rates by human-in-the-loop collaboration mode}
    (Flow/Prog./Corr./Req./Intr./Ask denote workflow, progressive refinement,
    correction, requirement change, interruption, and clarification).}
    \label{fig:kimi_k26_collaboration_mode_accuracy}
\end{figure}

\textbf{Obs.\ding{187}} \textbf{Explicit revision feedback is easier to handle than interrupted workflows.}
Kimi-K2.6 and GPT-5.4 perform best on correction/feedback tasks, where the user provides concrete revision guidance.
Performance is lower on interruption tasks, suggesting
that current agents use explicit local feedback more accurately than
mid-workflow changes that require replanning.
This pattern indicates that agents are better at making bounded local edits
than at preserving execution state and repairing a plan after the workflow is disrupted.

\textbf{Obs.\ding{188}} \textbf{Agents rarely ask for clarification when goals are underspecified.}
Ask-style tasks have the lowest success rates for both Kimi-K2.6 and
GPT-5.4. Thus, exposing an Ask channel is not
sufficient; current agents often proceed without requesting the missing
information needed for successful execution.
The dominant failure mode appears to be over-commitment to an initial guess.

\section{Conclusion}
\label{sec:conclusion}


We introduced \textbf{DeskCraft}, a 538-task execution-based benchmark 
for desktop GUI agents, featuring an L1/L2/L3 difficulty taxonomy, an
executable interaction protocol, and professional workflow 
coverage beyond existing desktop benchmarks. Across standard and interactive settings, experiments show that 
current agents remain far from robust on long-horizon and interactive tasks, 
with substantial weaknesses in workflow completion, replanning under intervention, and proactive
clarification. Additional steps recover a small tail of successes. By making these challenges explicit and measurable, DeskCraft provides a concrete basis for
evaluating progress on realistic desktop agents.

\section*{Limitations}
\label{sec:limitations}

DeskCraft expands desktop-agent evaluation to longer workflows, interactive collaboration, and professional software, but it still has several scope
boundaries. First, although the benchmark includes both English and Chinese instructions, its language coverage is still partial rather than fully
multilingual. Second, the interaction protocol uses scripted user messages to ensure reproducibility and controlled comparison across agents; this
makes evaluation stable, but it cannot capture the full diversity and unpredictability of real human collaboration.
Finally, DeskCraft is a fixed benchmark release with a finite set of applications, workflows, and step budgets, so it should be viewed as a incomplete slice of real-world desktop work.
%

\bibliography{custom}
\clearpage
\appendix


\section{Interaction Protocol Implementation Details}
\label{app:interaction_example}

We illustrate the interactive execution logic through the way DeskCraft injects
user messages into the Kimi GUI agent. At the start of an interactive task, the
instruction for Phase 1 is used as the initial task request. If any later phase
uses the \texttt{agent\_asks} trigger, the runtime switches Kimi into an
interactive mode before rollout so that the agent can explicitly request user
clarification when needed.

This interactive mode extends the agent prompt with a dedicated
\texttt{call\_user} tool and a short behavioral suffix:

\begin{PromptBlock}
- {
    "name": "call_user",
    "description": "Ask the user for clarification when the instruction
      is ambiguous, incomplete, or updated mid-task.",
    "parameters": {
      "type": "object",
      "properties": {
        "question": {
          "type": "string",
          "description": "A short, specific question for the user."
        }
      },
      "required": ["question"]
    }
  }

This is an interactive session.
- If the instruction is ambiguous or missing details, call `call_user`
  to ask a precise clarification question.
- If the user provides an update or changes the requirement later,
  incorporate it and continue from the current desktop state.
- Do not pretend the user already answered if they have not.
\end{PromptBlock}

After each agent turn, the interaction handler checks whether the current phase
trigger has fired. For \texttt{agent\_done} and \texttt{step\_count} triggers,
the phase index is advanced before the next user utterance is generated, so the
simulator sees the next phase goal instead of repeating the previous one. For
\texttt{agent\_asks}, the handler extracts the \texttt{call\_user} question,
passes it to the simulator, and treats the simulator response as the next user
message.

The resulting user message is delivered to Kimi through
\texttt{receive\_user\_message}. Kimi stores the message in two places. First,
it is placed in a \emph{pending} buffer that is consumed at the next
\texttt{predict} call and injected as a highest-priority turn-local update:

\begin{PromptBlock}
The following message is newly added this turn and should be treated
as highest-priority update.
[User Additional Message]:
{message}
\end{PromptBlock}

Second, the same message is appended to a bounded persistent history so that
previous user constraints remain visible in later steps. Older messages are
inserted into the task instruction as:

\begin{PromptBlock}
Follow all persistent user requirements below unless a newer
requirement explicitly supersedes an older one.
[Persistent User Requirement 1]:
{message_1}
[Persistent User Requirement 2]:
{message_2}
\end{PromptBlock}

Messages injected in the current turn are removed from the persistent block for
that same call to avoid duplication. If Kimi emits \texttt{call\_user}, the
parser converts it to a non-executable \texttt{CALL\_USER} signal, skips
environment execution for that turn, and lets the simulator produce a response
before rollout resumes from the same desktop state. In this way, the full
interaction protocol can be viewed as a loop over four stages: execute the
current desktop action, check whether the authored phase trigger fires,
generate the next user utterance if needed, and inject that utterance back into
the agent context for the next step.

\section{User Simulator Prompt Template}
\label{app:simulator}

\label{app:simulator_prompt}
\textbf{Simulator Prompt}
For interactive tasks, DeskCraft uses an LLM-based user simulator to generate
the next user utterance while keeping the interaction tied to the authored
phase protocol. The simulator is conditioned on the task scenario, user
persona, completed phases, current phase goal, optional next phase goal, the
agent's latest reply, and the current screenshot. Its system prompt template is:
\begin{table*}[t]
    \centering
    \small
    \caption{\textbf{Distribution of human-in-the-loop collaboration modes in the
    Interactive split.} Each task is assigned one primary mode for non-overlapping
    analysis.}
    \label{tab:collaboration_mode_distribution}
    \begin{tabular}{@{}lrrp{0.43\textwidth}@{}}
    \toprule
    Collaboration mode & Primary & Any-label & Main capability tested \\
    \midrule
    Progressive refinement & 50 (32.9\%) & 85 (55.9\%) &
    Incorporating staged user requirements across phases \\
    Ambiguity / clarification & 32 (21.1\%) & 34 (22.4\%) &
    Asking for missing information before execution \\
    Requirement change & 25 (16.4\%) & 55 (36.2\%) &
    Adapting to changed or newly added constraints \\
    Interruption & 21 (13.8\%) & 22 (14.5\%) &
    Replanning after mid-course user intervention \\
    Correction / feedback & 17 (11.2\%) & 20 (13.2\%) &
    Revising an existing or attempted result \\
    Multi-step workflow & 7 (4.6\%) & 26 (17.1\%) &
    Continuing staged delivery without strong correction or ambiguity \\
    \bottomrule
    \end{tabular}
    \end{table*}
\begin{PromptBlock}
You are roleplaying as a realistic computer user.
You are trying to complete a task on a computer, and an AI
assistant is helping operate the screen for you.

## Current Scenario
{scenario_description}
## Your Persona
- Expertise level: {expertise_level}
- Communication style: {communication_style}
{completed_phases_section}
## Current Phase Goal (Phase {current_phase_number} of {total_phases})
You need to ask the AI assistant to do the following:
{current_phase_instruction}
{next_phase_section}

## Rules
1. Speak like a real user. Do not use overly precise technical
   terms unless your persona is a professional user.
2. Use the screenshot to judge whether the AI assistant has
   completed the current requirement.
3. If a "Next Phase Goal" is provided above, naturally ask for
   that requirement next. Do not invent new requests on your own.
4. If the current phase is complete and there is no next phase
   goal, indicate that the whole task is finished and do not add
   any new requests.
5. Keep the conversation natural and coherent, like a real person
   chatting with an AI assistant.
6. Your `message` should follow the language implied by the scenario
   and current instruction. If the task context is in Chinese, reply
   in Chinese; if it is in English, reply in English.
7. In normal cases, always set `action` to `new_instruction`.
8. If the AI assistant has not completed the current phase, keep the
   interaction in the same phase: set `phase_complete` to false and
   use `message` to restate or correct the current requirement.
9. If the AI assistant has completed the current phase and there is a
   next phase goal, set `phase_complete` to true and use `message` to
   naturally express that next phase goal.
10. If the current phase expects the AI assistant to ask the user a
    question, answer that question directly and naturally. In that
    case, use `clarify` and set `phase_complete` to true.
11. If the AI assistant explicitly asks the user a question unexpectedly,
    you may use `clarify`, and in that case `phase_complete` must be
    false.

## Output Format
You must output valid JSON with the following fields:
{
    "action": "new_instruction" or "clarify",
    "message": "What you want to say to the AI assistant",
    "phase_complete": true or false,
    "reason": "When phase_complete is false, explain why the
               current phase is not complete"
}

Meaning of `action`:
- "new_instruction": The default and normal case. Use it for both
  correcting the current phase requirement and expressing the next
  phase requirement.
- "clarify": Only use this when the AI assistant explicitly asks the
  user a question unexpectedly. Do not use it otherwise.
\end{PromptBlock}

When the trigger type is \texttt{agent\_asks} and the GUI agent explicitly
asks a question, the simulator receives an additional instruction telling it
to answer the question directly, set \texttt{action} to \texttt{clarify}, and
mark the phase as complete. If the agent calls the user unexpectedly on a
phase that is not authored as \texttt{agent\_asks}, the simulator is instead
instructed to answer briefly without advancing the phase.

\section{Human-in-the-Loop Collaboration Mode Labels}
\label{app:hitl_modes}

Table~\ref{tab:collaboration_mode_distribution} reports the distribution of
human-in-the-loop collaboration labels used for the RQ4 analysis. Each task is
assigned one primary label for non-overlapping success-rate analysis.

The label distribution shows that interactive desktop tasks are often not
single-mode interactions: 91 of 170 tasks contain at least one secondary
collaboration label. We therefore use primary labels for the main
non-overlapping success-rate analysis and use any-label statistics only to
describe overlap among collaboration demands.

\section{Additional Experimental Details}

\clearpage
\onecolumn

\label{app:details}
%

\begin{table*}[t]
\centering
\begingroup
\footnotesize
\setlength{\tabcolsep}{2pt}
\renewcommand{\arraystretch}{0.8}

\newcolumntype{C}{>{\centering\arraybackslash}p{0.62cm}}
\newcolumntype{A}{>{\columncolor{gray!8}\centering\arraybackslash}p{0.70cm}}
\newcommand{\rh}[1]{\rotatebox{60}{\textbf{#1}}}

\resizebox{\textwidth}{!}{%
\begin{tabular}{l c | C C C C C C C C C C C C C | A}
\toprule
\multirow{2}{*}{\textbf{Agent}} & \multirow{2}{*}{\textbf{Lvl.}} & \multicolumn{14}{c}{\textbf{Non-interactive}} \\
\cmidrule(lr){3-16}
 & & \rh{Writer} & \rh{Calc} & \rh{Impress} & \rh{Chrome} & \rh{VS Code} & \rh{GIMP} & \rh{Inkscape} & \rh{Kdenlive} & \rh{Audacity} & \rh{Blender} & \rh{UI Gen} & \rh{Multi-app} & \rh{OS} & \textbf{Avg.} \\
\midrule

\rowcolor{gray!10}
\multicolumn{16}{l}{\textit{\textbf{Proprietary frontier models}}} \\
\midrule
\multirow{3}{*}{GPT-5.4} & L1 & 100.0 & 0.0 & 22.2 & 33.3 & 46.7 & 14.3 & 63.6 & 18.2 & 77.8 & 70.0 & 0.0 & 43.5 & 40.0 & 39.0 \\
 & L2 & 60.0 & 42.9 & 0.0 & 35.3 & 66.7 & 28.6 & 62.5 & 53.8 & 21.4 & 70.6 & 25.0 & 25.0 & 36.4 & 40.7 \\
 & L3 & 20.0 & 10.0 & 0.0 & 0.0 & 33.3 & 11.1 & 14.3 & 12.5 & 0.0 & 10.0 & 0.0 & 0.0 & 20.0 & 9.5 \\
\midrule
\multirow{3}{*}{Kimi-K2.6} & L1 & 50.0 & 100.0 & 11.1 & 10.0 & 33.3 & 42.9 & 72.7 & 18.2 & 88.9 & 80.0 & 50.0 & 34.8 & 60.0 & 41.7 \\
 & L2 & 60.0 & 42.9 & 0.0 & 33.3 & 37.5 & 28.6 & 62.5 & 53.8 & 50.0 & 76.5 & 33.3 & 25.0 & 27.3 & 41.0 \\
 & L3 & 0.0 & 10.0 & 0.0 & 66.7 & 50.0 & 33.3 & 57.1 & 12.5 & 0.0 & 40.0 & 0.0 & 0.0 & 37.5 & 21.6 \\
\midrule
\rowcolor{gray!10}
\multicolumn{16}{l}{\textit{\textbf{Open-source generalist VLMs}}} \\
\midrule
\multirow{3}{*}{Qwen3-VL-235B-A22B} & L1 & 50.0 & 100.0 & 0.0 & 30.0 & 0.0 & 14.3 & 0.0 & 4.5 & 0.0 & 30.0 & 0.0 & 8.7 & 20.0 & 10.1 \\
 & L2 & 0.0 & 14.3 & 0.0 & 0.0 & 0.0 & 0.0 & 0.0 & 0.0 & 0.0 & 0.0 & 0.0 & 8.3 & 0.0 & 1.5 \\
 & L3 & 0.0 & 0.0 & 0.0 & 0.0 & 0.0 & 0.0 & 0.0 & 0.0 & 0.0 & 0.0 & 0.0 & 0.0 & 0.0 & 0.0 \\
\midrule
\multirow{3}{*}{Qwen3-VL-32B} & L1 & 50.0 & 0.0 & 11.1 & 30.0 & 6.7 & 14.3 & 0.0 & 9.1 & 22.2 & 20.0 & 10.0 & 13.0 & 0.0 & 12.2 \\
 & L2 & 0.0 & 14.3 & 0.0 & 16.7 & 11.1 & 0.0 & 0.0 & 23.1 & 0.0 & 0.0 & 0.0 & 8.3 & 0.0 & 5.9 \\
 & L3 & 0.0 & 0.0 & 0.0 & 0.0 & 0.0 & 0.0 & 0.0 & 0.0 & 0.0 & 0.0 & 0.0 & 0.0 & 0.0 & 0.0 \\
\midrule
\multirow{3}{*}{Qwen3-VL-8B} & L1 & 50.0 & 0.0 & 11.1 & 0.0 & 0.0 & 0.0 & 0.0 & 0.0 & 0.0 & 10.0 & 0.0 & 8.7 & 0.0 & 3.6 \\
 & L2 & 40.0 & 28.6 & 0.0 & 0.0 & 0.0 & 0.0 & 0.0 & 7.7 & 0.0 & 0.0 & 16.7 & 0.0 & 0.0 & 5.2 \\
 & L3 & 0.0 & 0.0 & 0.0 & 0.0 & 0.0 & 0.0 & 0.0 & 0.0 & 0.0 & 0.0 & 0.0 & 0.0 & 0.0 & 0.0 \\
\midrule
\multirow{3}{*}{Qwen3.5-35B-A3B} & L1 & 0.0 & 0.0 & 22.2 & 20.0 & 13.3 & 28.6 & 27.3 & 13.6 & 44.4 & 30.0 & 0.0 & 26.1 & 30.0 & 21.6 \\
 & L2 & 20.0 & 0.0 & 0.0 & 8.3 & 11.1 & 0.0 & 0.0 & 38.5 & 0.0 & 11.8 & 0.0 & 8.3 & 0.0 & 8.1 \\
 & L3 & 0.0 & 0.0 & 0.0 & 0.0 & 0.0 & 0.0 & 0.0 & 0.0 & 0.0 & 0.0 & 0.0 & 0.0 & 10.0 & 1.0 \\
\midrule
\multirow{3}{*}{Qwen3.5-397B-A17B} & L1 & 0.0 & 0.0 & 11.1 & 30.0 & 20.0 & 14.3 & 18.2 & 18.2 & 11.1 & 60.0 & 0.0 & 26.1 & 40.0 & 22.3 \\
 & L2 & 20.0 & 14.3 & 0.0 & 16.7 & 11.1 & 0.0 & 12.5 & 23.1 & 14.3 & 5.9 & 8.3 & 0.0 & 18.2 & 11.1 \\
 & L3 & 0.0 & 0.0 & 0.0 & 0.0 & 0.0 & 0.0 & 0.0 & 0.0 & 0.0 & 0.0 & 0.0 & 0.0 & 20.0 & 1.9 \\
\midrule
\multirow{3}{*}{Qwen3.5-9B} & L1 & 0.0 & 100.0 & 11.1 & 30.0 & 26.7 & 14.3 & 27.3 & 9.1 & 22.2 & 30.0 & 0.0 & 8.7 & 30.0 & 18.0 \\
 & L2 & 20.0 & 14.3 & 0.0 & 16.7 & 11.1 & 0.0 & 0.0 & 7.7 & 0.0 & 0.0 & 8.3 & 0.0 & 9.1 & 5.9 \\
 & L3 & 0.0 & 0.0 & 0.0 & 0.0 & 0.0 & 0.0 & 0.0 & 0.0 & 0.0 & 0.0 & 0.0 & 0.0 & 0.0 & 0.0 \\
\midrule
\multirow{3}{*}{Qwen3.6-35B-A3B} & L1 & 0.0 & 0.0 & 22.2 & 10.0 & 13.3 & 0.0 & 18.2 & 9.1 & 11.1 & 50.0 & 0.0 & 13.0 & 30.0 & 15.1 \\
 & L2 & 20.0 & 0.0 & 0.0 & 16.7 & 11.1 & 0.0 & 37.5 & 46.2 & 0.0 & 11.8 & 0.0 & 8.3 & 9.1 & 12.6 \\
 & L3 & 0.0 & 0.0 & 0.0 & 0.0 & 0.0 & 0.0 & 0.0 & 0.0 & 0.0 & 0.0 & 0.0 & 0.0 & 10.0 & 1.0 \\
\midrule
\rowcolor{gray!10}
\multicolumn{16}{l}{\textit{\textbf{Open-source CUA foundation models}}} \\
\midrule
\multirow{3}{*}{EvoCUA-32B} & L1 & 0.0 & 0.0 & 11.1 & 16.7 & 20.0 & 0.0 & 45.5 & 18.2 & 22.2 & 40.0 & 0.0 & 17.4 & 30.0 & 19.9 \\
 & L2 & 0.0 & 14.3 & 0.0 & 35.3 & 11.1 & 0.0 & 12.5 & 30.8 & 0.0 & 0.0 & 0.0 & 0.0 & 18.2 & 10.7 \\
 & L3 & 0.0 & 0.0 & 0.0 & 0.0 & 0.0 & 0.0 & 0.0 & 0.0 & 0.0 & 0.0 & 0.0 & 0.0 & 10.0 & 1.0 \\
\midrule
\multirow{3}{*}{EvoCUA-8B} & L1 & 0.0 & 0.0 & 11.1 & 16.7 & 20.0 & 0.0 & 18.2 & 0.0 & 0.0 & 20.0 & 30.0 & 13.0 & 10.0 & 12.1 \\
 & L2 & 20.0 & 14.3 & 0.0 & 17.6 & 11.1 & 0.0 & 0.0 & 7.7 & 0.0 & 0.0 & 0.0 & 8.3 & 18.2 & 7.1 \\
 & L3 & 0.0 & 0.0 & 0.0 & 0.0 & 0.0 & 0.0 & 0.0 & 0.0 & 0.0 & 0.0 & 0.0 & 0.0 & 0.0 & 0.0 \\
\midrule
\multirow{3}{*}{GUI-Owl-1.5-32B-Instruct} & L1 & 0.0 & 0.0 & 11.1 & 0.0 & 6.7 & 14.3 & 0.0 & 4.5 & 11.1 & 20.0 & 0.0 & 8.7 & 0.0 & 6.4 \\
 & L2 & 0.0 & 0.0 & 0.0 & 5.9 & 11.1 & 0.0 & 0.0 & 23.1 & 0.0 & 5.9 & 0.0 & 0.0 & 9.1 & 5.0 \\
 & L3 & 0.0 & 0.0 & 0.0 & 0.0 & 0.0 & 0.0 & 0.0 & 0.0 & 0.0 & 0.0 & 0.0 & 0.0 & 0.0 & 0.0 \\
\midrule
\multirow{3}{*}{GUI-Owl-1.5-8B-Instruct} & L1 & 0.0 & 0.0 & 0.0 & 8.3 & 0.0 & 14.3 & 9.1 & 9.1 & 0.0 & 40.0 & 0.0 & 13.0 & 20.0 & 9.9 \\
 & L2 & 0.0 & 0.0 & 0.0 & 5.9 & 11.1 & 0.0 & 0.0 & 0.0 & 0.0 & 0.0 & 0.0 & 0.0 & 0.0 & 1.4 \\
 & L3 & 0.0 & 0.0 & 0.0 & 0.0 & 0.0 & 0.0 & 0.0 & 0.0 & 0.0 & 0.0 & 0.0 & 0.0 & 0.0 & 0.0 \\
\midrule
\multirow{3}{*}{OpenCUA-7B} & L1 & 0.0 & 0.0 & 11.1 & 33.3 & 20.0 & 0.0 & 9.1 & 9.1 & 22.2 & 10.0 & 0.0 & 8.7 & 10.0 & 12.1 \\
 & L2 & 0.0 & 0.0 & 0.0 & 11.8 & 11.1 & 0.0 & 0.0 & 0.0 & 0.0 & 0.0 & 0.0 & 0.0 & 0.0 & 2.1 \\
 & L3 & 0.0 & 0.0 & 0.0 & 0.0 & 0.0 & 0.0 & 0.0 & 0.0 & 0.0 & 0.0 & 0.0 & 0.0 & 0.0 & 0.0 \\
\midrule
\multirow{3}{*}{OS-Atlas-Pro-7B} & L1 & 0.0 & 0.0 & 0.0 & 0.0 & 0.0 & 0.0 & 0.0 & 0.0 & 0.0 & 0.0 & 0.0 & 0.0 & 0.0 & 0.0 \\
 & L2 & 0.0 & 14.3 & 0.0 & 0.0 & 0.0 & 0.0 & 0.0 & 0.0 & 0.0 & 0.0 & 0.0 & 0.0 & 0.0 & 0.7 \\
 & L3 & 0.0 & 0.0 & 0.0 & 0.0 & 0.0 & 0.0 & 0.0 & 0.0 & 0.0 & 0.0 & 0.0 & 0.0 & 0.0 & 0.0 \\
\midrule
\multirow{3}{*}{UI-TARS-1.5-7B} & L1 & 0.0 & 0.0 & 11.1 & 16.7 & 20.0 & 0.0 & 9.1 & 4.5 & 0.0 & 20.0 & 0.0 & 8.7 & 0.0 & 8.5 \\
 & L2 & 0.0 & 0.0 & 0.0 & 0.0 & 0.0 & 0.0 & 0.0 & 0.0 & 0.0 & 0.0 & 0.0 & 0.0 & 0.0 & 0.0 \\
 & L3 & 0.0 & 0.0 & 0.0 & 0.0 & 0.0 & 0.0 & 0.0 & 0.0 & 0.0 & 0.0 & 0.0 & 0.0 & 0.0 & 0.0 \\
\bottomrule
\end{tabular}%
}
\endgroup
\caption{%
\textbf{Per-software success rate on non-interactive DeskCraft tasks by difficulty level.}
Each model is expanded into three rows (\textbf{L1}, \textbf{L2}, \textbf{L3}). The \textbf{Avg.} column is the weighted success rate within that difficulty level.
}
\label{tab:non_interactive_per_software_level_results}
\end{table*}


\begin{table*}[t]
\centering
\begingroup
\scriptsize
\setlength{\tabcolsep}{1.5pt}
\renewcommand{\arraystretch}{0.72}

\newcolumntype{C}{>{\centering\arraybackslash}p{0.58cm}}
\newcolumntype{A}{>{\columncolor{gray!8}\centering\arraybackslash}p{0.66cm}}
\newcommand{\rh}[1]{\rotatebox{60}{\textbf{#1}}}

\resizebox{\textwidth}{!}{%
\begin{tabular}{l c | C C C C C C C C C C C C C | A}
\toprule
\multirow{2}{*}{\textbf{Agent}} & \multirow{2}{*}{\textbf{Lvl.}} & \multicolumn{14}{c}{\textbf{Non-interactive Avg. Steps}} \\
\cmidrule(lr){3-16}
 & & \rh{Writer} & \rh{Calc} & \rh{Impress} & \rh{Chrome} & \rh{VS Code} & \rh{GIMP} & \rh{Inkscape} & \rh{Kdenlive} & \rh{Audacity} & \rh{Blender} & \rh{UI Gen} & \rh{Multi-app} & \rh{OS} & \textbf{Avg.} \\
\midrule

\rowcolor{gray!10}
\multicolumn{16}{l}{\textit{\textbf{Proprietary frontier models}}} \\
\midrule
\multirow{3}{*}{GPT-5.4} & L1 & 20.5 & 6.0 & 31.4 & 20.0 & 30.9 & 39.0 & 17.5 & 21.6 & 42.4 & 12.7 & 44.2 & 23.4 & 6.0 & 25.0 \\
 & L2 & 31.0 & 16.4 & 33.4 & 19.5 & 64.0 & 133.9 & 37.2 & 68.2 & 56.1 & 32.6 & 64.2 & 31.8 & 12.7 & 44.3 \\
 & L3 & 65.0 & 67.2 & 61.0 & 64.0 & 48.5 & 124.8 & 62.9 & 75.8 & 78.6 & 128.2 & 66.2 & 52.9 & 16.8 & 71.2 \\
\midrule
\multirow{3}{*}{Kimi-K2.6} & L1 & 77.5 & 11.0 & 40.3 & 16.4 & 24.2 & 38.0 & 15.3 & 18.2 & 54.9 & 19.0 & 97.4 & 22.9 & 20.0 & 30.8 \\
 & L2 & 35.4 & 8.6 & 85.1 & 21.8 & 31.6 & 131.7 & 42.0 & 31.0 & 54.6 & 38.9 & 86.3 & 44.7 & 40.4 & 48.8 \\
 & L3 & 100.8 & 50.3 & 108.5 & 33.0 & 52.8 & 142.3 & 56.7 & 62.8 & 90.6 & 86.4 & 95.2 & 45.0 & 52.5 & 77.7 \\
\midrule
\rowcolor{gray!10}
\multicolumn{16}{l}{\textit{\textbf{Open-source generalist VLMs}}} \\
\midrule
\multirow{3}{*}{Qwen3-VL-235B-A22B} & L1 & 125.5 & 3.0 & 36.3 & 19.2 & 60.7 & 23.6 & 20.6 & 44.3 & 31.9 & 25.7 & 91.0 & 66.5 & 54.7 & 47.3 \\
 & L2 & 21.4 & 38.0 & 37.8 & 30.2 & 51.2 & 89.3 & 31.8 & 96.3 & 49.2 & 69.9 & 63.5 & 72.8 & 44.3 & 56.5 \\
 & L3 & 49.2 & 80.7 & 83.9 & 53.3 & 68.5 & 101.7 & 47.7 & 100.0 & 42.0 & 95.7 & 99.0 & 95.5 & 36.4 & 75.9 \\
\midrule
\multirow{3}{*}{Qwen3-VL-32B} & L1 & 146.0 & 4.0 & 58.9 & 60.5 & 67.7 & 48.0 & 44.8 & 41.9 & 49.8 & 36.2 & 95.9 & 73.4 & 73.1 & 60.3 \\
 & L2 & 75.0 & 52.7 & 62.4 & 82.9 & 86.7 & 103.1 & 80.1 & 70.8 & 68.1 & 82.6 & 67.5 & 96.4 & 91.5 & 78.8 \\
 & L3 & 98.4 & 95.6 & 100.1 & 100.0 & 100.3 & 94.2 & 93.0 & 91.6 & 93.0 & 105.4 & 106.5 & 94.0 & 83.4 & 96.3 \\
\midrule
\multirow{3}{*}{Qwen3-VL-8B} & L1 & 64.0 & 3.0 & 58.7 & 46.0 & 89.1 & 48.6 & 27.5 & 49.1 & 69.4 & 66.3 & 100.0 & 79.8 & 73.0 & 65.0 \\
 & L2 & 53.8 & 57.0 & 83.1 & 86.3 & 74.9 & 101.0 & 78.8 & 92.2 & 81.8 & 105.5 & 86.8 & 104.7 & 91.6 & 87.6 \\
 & L3 & 59.4 & 78.6 & 83.4 & 78.0 & 100.0 & 101.2 & 79.6 & 87.1 & 100.2 & 100.0 & 100.0 & 104.9 & 98.0 & 91.4 \\
\midrule
\multirow{3}{*}{Qwen3.5-35B-A3B} & L1 & 28.5 & 5.0 & 23.4 & 19.5 & 25.4 & 27.4 & 22.6 & 20.9 & 36.1 & 24.0 & 45.1 & 40.3 & 26.7 & 28.5 \\
 & L2 & 22.0 & 40.1 & 45.4 & 30.4 & 58.0 & 76.4 & 32.9 & 40.5 & 45.0 & 60.5 & 68.3 & 40.4 & 37.2 & 46.9 \\
 & L3 & 50.4 & 65.3 & 60.5 & 26.3 & 75.2 & 89.4 & 74.9 & 68.4 & 70.9 & 115.6 & 82.1 & 58.0 & 56.7 & 71.6 \\
\midrule
\multirow{3}{*}{Qwen3.5-397B-A17B} & L1 & 39.0 & 10.0 & 32.0 & 23.3 & 23.3 & 22.0 & 15.5 & 14.0 & 47.1 & 16.5 & 76.3 & 30.7 & 30.6 & 28.5 \\
 & L2 & 33.2 & 24.7 & 47.2 & 34.8 & 68.3 & 75.7 & 43.6 & 44.8 & 42.6 & 47.6 & 71.1 & 40.5 & 46.0 & 47.9 \\
 & L3 & 43.8 & 46.9 & 57.2 & 30.7 & 75.7 & 91.1 & 49.4 & 66.6 & 64.4 & 93.5 & 99.0 & 45.9 & 59.6 & 65.3 \\
\midrule
\multirow{3}{*}{Qwen3.5-9B} & L1 & 32.5 & 3.0 & 23.1 & 12.3 & 32.1 & 17.4 & 13.8 & 20.4 & 48.1 & 43.3 & 107.2 & 29.5 & 36.9 & 33.0 \\
 & L2 & 27.8 & 14.0 & 27.4 & 25.4 & 52.0 & 58.9 & 37.8 & 60.1 & 54.3 & 54.3 & 65.8 & 46.3 & 35.2 & 45.5 \\
 & L3 & 37.0 & 55.6 & 53.6 & 46.3 & 62.7 & 76.0 & 63.4 & 58.6 & 61.4 & 96.8 & 124.8 & 50.0 & 67.0 & 67.5 \\
\midrule
\multirow{3}{*}{Qwen3.6-35B-A3B} & L1 & 36.0 & 6.0 & 30.9 & 33.6 & 45.1 & 36.3 & 40.9 & 24.5 & 38.7 & 12.8 & 50.3 & 27.9 & 26.8 & 32.4 \\
 & L2 & 29.6 & 42.3 & 62.0 & 45.8 & 64.7 & 88.3 & 42.8 & 55.7 & 62.9 & 68.3 & 66.2 & 43.0 & 51.5 & 56.8 \\
 & L3 & 50.2 & 79.9 & 76.4 & 82.7 & 59.8 & 93.2 & 80.9 & 95.0 & 70.0 & 99.2 & 90.0 & 61.5 & 61.4 & 78.0 \\
\midrule
\rowcolor{gray!10}
\multicolumn{16}{l}{\textit{\textbf{Open-source CUA foundation models}}} \\
\midrule
\multirow{3}{*}{EvoCUA-32B} & L1 & 114.0 & 6.0 & 62.9 & 52.6 & 54.9 & 56.9 & 46.0 & 32.6 & 82.2 & 55.7 & 85.4 & 41.0 & 26.2 & 51.3 \\
 & L2 & 69.0 & 41.4 & 55.0 & 55.6 & 77.8 & 96.6 & 77.9 & 99.1 & 82.8 & 95.6 & 55.1 & 69.8 & 55.4 & 72.8 \\
 & L3 & 79.8 & 94.3 & 95.8 & 126.2 & 91.3 & 103.3 & 78.3 & 111.8 & 98.0 & 101.1 & 96.9 & 85.5 & 75.3 & 94.5 \\
\midrule
\multirow{3}{*}{EvoCUA-8B} & L1 & 95.0 & 4.0 & 53.9 & 48.8 & 67.6 & 51.1 & 34.9 & 30.9 & 80.9 & 50.6 & 83.0 & 43.2 & 23.3 & 49.6 \\
 & L2 & 80.4 & 63.9 & 66.6 & 62.9 & 81.9 & 92.1 & 64.0 & 84.2 & 79.9 & 83.6 & 59.9 & 47.8 & 53.5 & 70.4 \\
 & L3 & 101.0 & 103.2 & 85.7 & 118.5 & 92.3 & 85.8 & 68.3 & 120.5 & 82.6 & 108.4 & 102.8 & 47.1 & 59.3 & 89.0 \\
\midrule
\multirow{3}{*}{GUI-Owl-1.5-32B-Instruct} & L1 & 100.0 & 100.0 & 39.4 & 78.0 & 85.8 & 86.6 & 74.2 & 73.3 & 74.8 & 64.1 & 100.0 & 85.6 & 72.8 & 77.5 \\
 & L2 & 81.8 & 95.1 & 87.9 & 74.6 & 81.3 & 100.0 & 79.8 & 76.2 & 85.9 & 97.4 & 100.0 & 93.4 & 78.9 & 86.8 \\
 & L3 & 100.0 & 87.3 & 100.0 & 65.8 & 100.0 & 100.0 & 99.9 & 100.0 & 100.0 & 100.0 & 100.0 & 100.0 & 100.0 & 97.5 \\
\midrule
\multirow{3}{*}{GUI-Owl-1.5-8B-Instruct} & L1 & 43.5 & 3.0 & 51.4 & 22.4 & 51.1 & 19.3 & 22.2 & 35.2 & 72.6 & 25.5 & 93.8 & 55.3 & 43.1 & 44.6 \\
 & L2 & 64.4 & 26.7 & 41.8 & 38.6 & 64.7 & 89.7 & 33.5 & 58.5 & 85.7 & 56.9 & 91.5 & 73.5 & 79.1 & 62.5 \\
 & L3 & 55.6 & 61.5 & 88.4 & 41.2 & 89.5 & 89.0 & 46.9 & 100.0 & 60.9 & 100.0 & 100.0 & 61.2 & 60.6 & 75.9 \\
\midrule
\multirow{3}{*}{OpenCUA-7B} & L1 & 45.5 & 4.0 & 33.3 & 21.6 & 20.0 & 33.7 & 15.6 & 31.3 & 38.3 & 29.1 & 54.5 & 35.3 & 25.4 & 30.5 \\
 & L2 & 25.2 & 37.3 & 24.2 & 27.4 & 53.9 & 75.3 & 36.0 & 78.2 & 47.8 & 66.9 & 50.7 & 51.8 & 19.4 & 47.2 \\
 & L3 & 64.2 & 45.4 & 79.4 & 44.5 & 64.7 & 88.1 & 73.3 & 91.8 & 59.2 & 92.7 & 90.9 & 50.0 & 27.5 & 68.0 \\
\midrule
\multirow{3}{*}{OS-Atlas-Pro-7B} & L1 & 2.0 & 100.0 & 57.4 & 37.5 & 30.3 & 19.6 & 10.5 & 17.6 & 5.0 & 28.5 & 6.4 & 18.7 & 52.8 & 24.9 \\
 & L2 & 2.4 & 29.0 & 2.9 & 21.5 & 21.2 & 2.0 & 2.6 & 11.4 & 2.7 & 45.1 & 14.5 & 41.8 & 46.1 & 21.2 \\
 & L3 & 2.0 & 12.2 & 13.0 & 27.8 & 26.5 & 4.1 & 18.3 & 29.6 & 2.6 & 32.6 & 15.0 & 53.5 & 52.4 & 22.7 \\
\midrule
\multirow{3}{*}{UI-TARS-1.5-7B} & L1 & 66.0 & 6.0 & 36.8 & 21.2 & 30.0 & 27.0 & 37.1 & 33.5 & 55.4 & 24.6 & 59.6 & 39.3 & 72.3 & 38.8 \\
 & L2 & 52.4 & 51.1 & 49.1 & 40.4 & 41.1 & 43.9 & 38.6 & 14.2 & 48.4 & 38.1 & 68.9 & 59.0 & 60.3 & 45.7 \\
 & L3 & 77.2 & 62.5 & 59.2 & 33.2 & 35.3 & 61.4 & 68.6 & 16.0 & 44.1 & 62.2 & 49.8 & 56.6 & 95.2 & 57.2 \\
\bottomrule
\end{tabular}%
}
\endgroup
\caption{%
\textbf{Per-software average run length on non-interactive DeskCraft tasks by difficulty level.}
Each model is expanded into three rows (\textbf{L1}, \textbf{L2}, \textbf{L3}).
Values are average executed steps computed from \texttt{results/summary\_json\_collection/non\_interactive}.
The \textbf{Avg.} column is the weighted average run length within that difficulty level.
}
\label{tab:non_interactive_per_software_average_run_length_results}
\end{table*}


\begin{table*}[t]
\centering
\begingroup
\scriptsize
\setlength{\tabcolsep}{1.5pt}
\renewcommand{\arraystretch}{0.72}

\newcolumntype{C}{>{\centering\arraybackslash}p{0.58cm}}
\newcolumntype{A}{>{\columncolor{gray!8}\centering\arraybackslash}p{0.66cm}}
\newcommand{\rh}[1]{\rotatebox{60}{\textbf{#1}}}

\resizebox{\textwidth}{!}{%
\begin{tabular}{l c | C C C C C C C C C C C C C | A}
\toprule
\multirow{2}{*}{\textbf{Agent}} & \multirow{2}{*}{\textbf{Lvl.}} & \multicolumn{14}{c}{\textbf{Non-interactive Correct Steps}} \\
\cmidrule(lr){3-16}
 & & \rh{Writer} & \rh{Calc} & \rh{Impress} & \rh{Chrome} & \rh{VS Code} & \rh{GIMP} & \rh{Inkscape} & \rh{Kdenlive} & \rh{Audacity} & \rh{Blender} & \rh{UI Gen} & \rh{Multi-app} & \rh{OS} & \textbf{Avg.} \\
\midrule

\rowcolor{gray!10}
\multicolumn{16}{l}{\textit{\textbf{Proprietary frontier models}}} \\
\midrule
\multirow{3}{*}{GPT-5.4} & L1 & 20.5 & -- & 9.0 & 12.2 & 34.6 & 20.0 & 16.1 & 33.8 & 41.0 & 11.0 & -- & 21.7 & 4.8 & 22.1 \\
 & L2 & 36.0 & 20.0 & -- & 13.3 & 38.7 & 147.5 & 36.2 & 59.7 & 43.0 & 31.2 & 54.0 & 38.3 & 6.0 & 38.2 \\
 & L3 & 81.0 & 38.0 & -- & -- & 43.5 & 192.0 & 38.0 & 56.0 & -- & 109.0 & -- & -- & 4.0 & 60.9 \\
\midrule
\multirow{3}{*}{Kimi-K2.6} & L1 & -- & 11.0 & 4.0 & 12.0 & 29.4 & 50.7 & 17.4 & 13.5 & 45.8 & 21.6 & 141.6 & 15.5 & 21.7 & 34.8 \\
 & L2 & 33.0 & 8.0 & -- & 11.0 & 21.0 & 153.0 & 38.2 & 24.7 & 61.6 & 35.8 & 43.2 & 19.3 & 25.3 & 37.1 \\
 & L3 & -- & 35.0 & -- & 43.5 & 57.3 & 130.3 & 67.2 & 31.0 & -- & 73.5 & -- & -- & 9.0 & 62.2 \\
\midrule
\rowcolor{gray!10}
\multicolumn{16}{l}{\textit{\textbf{Open-source generalist VLMs}}} \\
\midrule
\multirow{3}{*}{Qwen3-VL-235B-A22B} & L1 & 194.0 & 3.0 & -- & 11.3 & -- & 8.0 & -- & 10.0 & -- & 14.7 & -- & 16.5 & 8.0 & 24.4 \\
 & L2 & -- & 5.0 & -- & -- & -- & -- & -- & -- & -- & -- & -- & 34.0 & -- & 19.5 \\
 & L3 & -- & -- & -- & -- & -- & -- & -- & -- & -- & -- & -- & -- & -- & -- \\
\midrule
\multirow{3}{*}{Qwen3-VL-32B} & L1 & 190.0 & -- & 5.0 & 15.7 & 19.0 & 7.0 & -- & 22.5 & 35.5 & 23.5 & 102.0 & 16.3 & -- & 34.2 \\
 & L2 & -- & 100.0 & -- & 16.0 & 26.0 & -- & -- & 30.7 & -- & -- & -- & 30.0 & -- & 35.0 \\
 & L3 & -- & -- & -- & -- & -- & -- & -- & -- & -- & -- & -- & -- & -- & -- \\
\midrule
\multirow{3}{*}{Qwen3-VL-8B} & L1 & 102.0 & -- & 5.0 & -- & -- & -- & -- & -- & -- & 100.0 & -- & 20.5 & -- & 49.6 \\
 & L2 & 65.0 & 100.0 & -- & -- & -- & -- & -- & 29.0 & -- & -- & 11.5 & -- & -- & 54.6 \\
 & L3 & -- & -- & -- & -- & -- & -- & -- & -- & -- & -- & -- & -- & -- & -- \\
\midrule
\multirow{3}{*}{Qwen3.5-35B-A3B} & L1 & -- & -- & 4.0 & 8.0 & 11.0 & 10.5 & 15.3 & 15.0 & 27.5 & 13.7 & -- & 17.3 & 11.7 & 14.9 \\
 & L2 & 29.0 & -- & -- & 7.0 & 14.0 & -- & -- & 29.0 & -- & 22.0 & -- & 25.0 & -- & 24.0 \\
 & L3 & -- & -- & -- & -- & -- & -- & -- & -- & -- & -- & -- & -- & 21.0 & 21.0 \\
\midrule
\multirow{3}{*}{Qwen3.5-397B-A17B} & L1 & -- & -- & 3.0 & 17.3 & 17.0 & 10.0 & 11.5 & 13.8 & 41.0 & 17.7 & -- & 24.2 & 11.8 & 17.2 \\
 & L2 & 37.0 & 3.0 & -- & 57.0 & 12.0 & -- & 33.0 & 38.3 & 29.5 & 30.0 & 14.0 & -- & 53.0 & 34.9 \\
 & L3 & -- & -- & -- & -- & -- & -- & -- & -- & -- & -- & -- & -- & 16.0 & 16.0 \\
\midrule
\multirow{3}{*}{Qwen3.5-9B} & L1 & -- & 3.0 & 4.0 & 13.3 & 18.2 & 9.0 & 20.0 & 7.5 & 31.5 & 17.7 & -- & 18.0 & 14.7 & 16.0 \\
 & L2 & 39.0 & 1.0 & -- & 12.0 & 12.0 & -- & -- & 31.0 & -- & -- & 63.0 & -- & 27.0 & 24.6 \\
 & L3 & -- & -- & -- & -- & -- & -- & -- & -- & -- & -- & -- & -- & -- & -- \\
\midrule
\multirow{3}{*}{Qwen3.6-35B-A3B} & L1 & -- & -- & 5.0 & 6.0 & 22.5 & -- & 19.5 & 8.5 & 27.0 & 15.6 & -- & 14.3 & 16.7 & 15.0 \\
 & L2 & 22.0 & -- & -- & 27.0 & 10.0 & -- & 29.7 & 30.3 & -- & 21.0 & -- & 26.0 & 25.0 & 26.5 \\
 & L3 & -- & -- & -- & -- & -- & -- & -- & -- & -- & -- & -- & -- & 7.0 & 7.0 \\
\midrule
\rowcolor{gray!10}
\multicolumn{16}{l}{\textit{\textbf{Open-source CUA foundation models}}} \\
\midrule
\multirow{3}{*}{EvoCUA-32B} & L1 & -- & -- & 22.0 & 40.0 & 16.3 & -- & 36.2 & 23.0 & 50.0 & 21.2 & -- & 15.2 & 31.0 & 27.2 \\
 & L2 & -- & 64.0 & -- & 51.0 & 24.0 & -- & 83.0 & 59.0 & -- & -- & -- & -- & 22.5 & 50.5 \\
 & L3 & -- & -- & -- & -- & -- & -- & -- & -- & -- & -- & -- & -- & 33.0 & 33.0 \\
\midrule
\multirow{3}{*}{EvoCUA-8B} & L1 & -- & -- & 5.0 & 12.5 & 24.7 & -- & 17.0 & -- & -- & 10.5 & 58.0 & 18.3 & 19.0 & 23.9 \\
 & L2 & 44.0 & 42.0 & -- & 22.0 & 15.0 & -- & -- & 35.0 & -- & -- & -- & 16.0 & 18.0 & 25.4 \\
 & L3 & -- & -- & -- & -- & -- & -- & -- & -- & -- & -- & -- & -- & -- & -- \\
\midrule
\multirow{3}{*}{GUI-Owl-1.5-32B-Instruct} & L1 & -- & -- & 5.0 & -- & 9.0 & 6.0 & -- & 7.0 & 26.0 & 11.5 & -- & 51.5 & -- & 19.9 \\
 & L2 & -- & -- & -- & 15.0 & 15.0 & -- & -- & 23.7 & -- & 55.0 & -- & -- & 18.0 & 24.9 \\
 & L3 & -- & -- & -- & -- & -- & -- & -- & -- & -- & -- & -- & -- & -- & -- \\
\midrule
\multirow{3}{*}{GUI-Owl-1.5-8B-Instruct} & L1 & -- & -- & -- & 6.0 & -- & 8.0 & 18.0 & 8.5 & -- & 15.2 & -- & 16.7 & 28.5 & 15.5 \\
 & L2 & -- & -- & -- & 51.0 & 26.0 & -- & -- & -- & -- & -- & -- & -- & -- & 38.5 \\
 & L3 & -- & -- & -- & -- & -- & -- & -- & -- & -- & -- & -- & -- & -- & -- \\
\midrule
\multirow{3}{*}{OpenCUA-7B} & L1 & -- & -- & 4.0 & 26.2 & 17.3 & -- & 12.0 & 7.5 & 22.5 & 23.0 & -- & 12.5 & 36.0 & 18.6 \\
 & L2 & -- & -- & -- & 13.0 & 14.0 & -- & -- & -- & -- & -- & -- & -- & -- & 13.3 \\
 & L3 & -- & -- & -- & -- & -- & -- & -- & -- & -- & -- & -- & -- & -- & -- \\
\midrule
\multirow{3}{*}{OS-Atlas-Pro-7B} & L1 & -- & -- & -- & -- & -- & -- & -- & -- & -- & -- & -- & -- & -- & -- \\
 & L2 & -- & 1.0 & -- & -- & -- & -- & -- & -- & -- & -- & -- & -- & -- & 1.0 \\
 & L3 & -- & -- & -- & -- & -- & -- & -- & -- & -- & -- & -- & -- & -- & -- \\
\midrule
\multirow{3}{*}{UI-TARS-1.5-7B} & L1 & -- & -- & 5.0 & 13.0 & 27.7 & -- & 16.0 & 7.0 & -- & 41.5 & -- & 17.0 & -- & 21.2 \\
 & L2 & -- & -- & -- & -- & -- & -- & -- & -- & -- & -- & -- & -- & -- & -- \\
 & L3 & -- & -- & -- & -- & -- & -- & -- & -- & -- & -- & -- & -- & -- & -- \\
\bottomrule
\end{tabular}%
}
\endgroup
\caption{%
\textbf{Per-software average run length on successful non-interactive DeskCraft tasks by difficulty level.}
Each model is expanded into three rows (\textbf{L1}, \textbf{L2}, \textbf{L3}).
Values are average executed steps computed from \texttt{results/summary\_json\_collection/non\_interactive}.
The \textbf{Avg.} column is the weighted average run length within that difficulty level.
`--` means the model has no successful tasks in that software/bucket combination at that difficulty level.
}
\label{tab:non_interactive_per_software_correct_run_length_results}
\end{table*}


\begin{table*}[t]
\centering
\begingroup
\scriptsize
\setlength{\tabcolsep}{1.5pt}
\renewcommand{\arraystretch}{0.72}

\newcolumntype{C}{>{\centering\arraybackslash}p{0.58cm}}
\newcolumntype{A}{>{\columncolor{gray!8}\centering\arraybackslash}p{0.66cm}}
\newcommand{\rh}[1]{\rotatebox{60}{\textbf{#1}}}

\resizebox{\textwidth}{!}{%
\begin{tabular}{l c | C C C C C C C C C C C C C | A}
\toprule
\multirow{2}{*}{\textbf{Agent}} & \multirow{2}{*}{\textbf{Lvl.}} & \multicolumn{14}{c}{\textbf{Non-interactive Wrong Steps}} \\
\cmidrule(lr){3-16}
 & & \rh{Writer} & \rh{Calc} & \rh{Impress} & \rh{Chrome} & \rh{VS Code} & \rh{GIMP} & \rh{Inkscape} & \rh{Kdenlive} & \rh{Audacity} & \rh{Blender} & \rh{UI Gen} & \rh{Multi-app} & \rh{OS} & \textbf{Avg.} \\
\midrule

\rowcolor{gray!10}
\multicolumn{16}{l}{\textit{\textbf{Proprietary frontier models}}} \\
\midrule
\multirow{3}{*}{GPT-5.4} & L1 & -- & 6.0 & 37.9 & 23.9 & 27.8 & 42.2 & 20.0 & 18.9 & 47.5 & 16.7 & 44.2 & 24.8 & 6.8 & 26.9 \\
 & L2 & 23.5 & 13.8 & 33.4 & 22.9 & 114.7 & 128.4 & 39.0 & 78.0 & 59.6 & 36.0 & 67.6 & 29.6 & 16.6 & 48.4 \\
 & L3 & 61.0 & 70.4 & 61.0 & 64.0 & 51.0 & 116.4 & 67.0 & 78.6 & 78.6 & 130.3 & 66.2 & 52.9 & 20.0 & 72.3 \\
\midrule
\multirow{3}{*}{Kimi-K2.6} & L1 & 77.5 & -- & 44.9 & 16.9 & 21.6 & 28.5 & 9.7 & 19.3 & 128.0 & 8.5 & 53.2 & 26.9 & 17.5 & 27.9 \\
 & L2 & 36.0 & 9.0 & 85.1 & 27.1 & 38.0 & 123.2 & 48.3 & 38.3 & 47.6 & 49.0 & 107.9 & 53.1 & 46.0 & 56.9 \\
 & L3 & 100.8 & 52.0 & 108.5 & 12.0 & 48.3 & 148.3 & 42.7 & 67.3 & 90.6 & 95.0 & 95.2 & 45.0 & 78.6 & 81.9 \\
\midrule
\rowcolor{gray!10}
\multicolumn{16}{l}{\textit{\textbf{Open-source generalist VLMs}}} \\
\midrule
\multirow{3}{*}{Qwen3-VL-235B-A22B} & L1 & 57.0 & -- & 36.3 & 22.6 & 60.7 & 26.2 & 20.6 & 45.9 & 31.9 & 30.4 & 91.0 & 71.3 & 66.4 & 49.9 \\
 & L2 & 21.4 & 43.5 & 37.8 & 30.2 & 51.2 & 89.3 & 31.8 & 96.3 & 49.2 & 69.9 & 63.5 & 76.3 & 44.3 & 57.1 \\
 & L3 & 49.2 & 80.7 & 83.9 & 53.3 & 68.5 & 101.7 & 47.7 & 100.0 & 42.0 & 95.7 & 99.0 & 95.5 & 36.4 & 75.9 \\
\midrule
\multirow{3}{*}{Qwen3-VL-32B} & L1 & 102.0 & 4.0 & 65.6 & 79.7 & 71.1 & 54.8 & 44.8 & 43.8 & 53.9 & 39.4 & 95.2 & 82.0 & 73.1 & 64.0 \\
 & L2 & 75.0 & 44.8 & 62.4 & 96.3 & 94.2 & 103.1 & 80.1 & 82.8 & 68.1 & 82.6 & 67.5 & 102.5 & 91.5 & 81.5 \\
 & L3 & 98.4 & 95.6 & 100.1 & 100.0 & 100.3 & 94.2 & 93.0 & 91.6 & 93.0 & 105.4 & 106.5 & 94.0 & 83.4 & 96.3 \\
\midrule
\multirow{3}{*}{Qwen3-VL-8B} & L1 & 26.0 & 3.0 & 65.4 & 46.0 & 89.1 & 48.6 & 27.5 & 49.1 & 69.4 & 62.6 & 100.0 & 85.4 & 73.0 & 65.5 \\
 & L2 & 46.3 & 39.8 & 83.1 & 86.3 & 74.9 & 101.0 & 78.8 & 97.4 & 81.8 & 105.5 & 101.8 & 104.7 & 91.6 & 89.4 \\
 & L3 & 59.4 & 78.6 & 83.4 & 78.0 & 100.0 & 101.2 & 79.6 & 87.1 & 100.2 & 100.0 & 100.0 & 104.9 & 98.0 & 91.4 \\
\midrule
\multirow{3}{*}{Qwen3.5-35B-A3B} & L1 & 28.5 & 5.0 & 29.0 & 22.4 & 27.6 & 34.2 & 25.4 & 21.8 & 43.0 & 28.4 & 45.1 & 48.5 & 33.1 & 32.2 \\
 & L2 & 20.2 & 40.1 & 45.4 & 32.5 & 63.5 & 76.4 & 32.9 & 47.8 & 45.0 & 65.6 & 68.3 & 41.8 & 37.2 & 49.0 \\
 & L3 & 50.4 & 65.3 & 60.5 & 26.3 & 75.2 & 89.4 & 74.9 & 68.4 & 70.9 & 115.6 & 82.1 & 58.0 & 60.7 & 72.1 \\
\midrule
\multirow{3}{*}{Qwen3.5-397B-A17B} & L1 & 39.0 & 10.0 & 35.6 & 25.9 & 24.9 & 24.0 & 16.4 & 14.1 & 47.9 & 14.8 & 76.3 & 33.1 & 43.2 & 31.7 \\
 & L2 & 32.2 & 28.3 & 47.2 & 30.4 & 75.4 & 75.7 & 45.1 & 46.7 & 44.8 & 48.7 & 76.3 & 40.5 & 44.4 & 49.5 \\
 & L3 & 43.8 & 46.9 & 57.2 & 30.7 & 75.7 & 91.1 & 49.4 & 66.6 & 64.4 & 93.5 & 99.0 & 45.9 & 70.5 & 66.3 \\
\midrule
\multirow{3}{*}{Qwen3.5-9B} & L1 & 32.5 & -- & 25.5 & 11.9 & 37.1 & 18.8 & 11.5 & 21.6 & 52.9 & 54.3 & 107.2 & 30.6 & 46.4 & 36.7 \\
 & L2 & 25.0 & 16.2 & 27.4 & 28.1 & 57.0 & 58.9 & 37.8 & 62.5 & 54.3 & 54.3 & 66.0 & 46.3 & 36.0 & 46.8 \\
 & L3 & 37.0 & 55.6 & 53.6 & 46.3 & 62.7 & 76.0 & 63.4 & 58.6 & 61.4 & 96.8 & 124.8 & 50.0 & 67.0 & 67.5 \\
\midrule
\multirow{3}{*}{Qwen3.6-35B-A3B} & L1 & 36.0 & 6.0 & 38.3 & 36.7 & 48.5 & 36.3 & 45.7 & 26.1 & 40.1 & 10.0 & 50.3 & 29.9 & 31.1 & 35.5 \\
 & L2 & 31.5 & 42.3 & 62.0 & 49.5 & 71.5 & 88.3 & 50.6 & 77.4 & 62.9 & 74.6 & 66.2 & 44.5 & 54.2 & 61.2 \\
 & L3 & 50.2 & 79.9 & 76.4 & 82.7 & 59.8 & 93.2 & 80.9 & 95.0 & 70.0 & 99.2 & 90.0 & 61.5 & 67.4 & 78.7 \\
\midrule
\rowcolor{gray!10}
\multicolumn{16}{l}{\textit{\textbf{Open-source CUA foundation models}}} \\
\midrule
\multirow{3}{*}{EvoCUA-32B} & L1 & 114.0 & 6.0 & 68.0 & 55.1 & 64.6 & 56.9 & 54.2 & 34.8 & 91.4 & 78.7 & 85.4 & 46.4 & 24.1 & 57.2 \\
 & L2 & 69.0 & 37.7 & 55.0 & 58.1 & 84.5 & 96.6 & 77.1 & 116.9 & 82.8 & 95.6 & 55.1 & 69.8 & 62.7 & 75.5 \\
 & L3 & 79.8 & 94.3 & 95.8 & 126.2 & 91.3 & 103.3 & 78.3 & 111.8 & 98.0 & 101.1 & 96.9 & 85.5 & 80.0 & 95.1 \\
\midrule
\multirow{3}{*}{EvoCUA-8B} & L1 & 95.0 & 4.0 & 60.0 & 56.1 & 78.3 & 51.1 & 38.9 & 30.9 & 80.9 & 60.6 & 93.7 & 47.0 & 23.8 & 53.1 \\
 & L2 & 89.5 & 67.5 & 66.6 & 71.6 & 90.2 & 92.1 & 64.0 & 88.3 & 79.9 & 83.6 & 59.9 & 50.6 & 61.3 & 73.9 \\
 & L3 & 101.0 & 103.2 & 85.7 & 118.5 & 92.3 & 85.8 & 68.3 & 120.5 & 82.6 & 108.4 & 102.8 & 47.1 & 59.3 & 89.0 \\
\midrule
\multirow{3}{*}{GUI-Owl-1.5-32B-Instruct} & L1 & 100.0 & 100.0 & 43.8 & 78.0 & 91.3 & 100.0 & 74.2 & 76.5 & 80.9 & 77.2 & 100.0 & 88.8 & 72.8 & 81.4 \\
 & L2 & 81.8 & 95.1 & 87.9 & 78.4 & 89.6 & 100.0 & 79.8 & 91.9 & 85.9 & 100.0 & 100.0 & 93.4 & 85.0 & 90.1 \\
 & L3 & 100.0 & 87.3 & 100.0 & 65.8 & 100.0 & 100.0 & 99.9 & 100.0 & 100.0 & 100.0 & 100.0 & 100.0 & 100.0 & 97.5 \\
\midrule
\multirow{3}{*}{GUI-Owl-1.5-8B-Instruct} & L1 & 43.5 & 3.0 & 51.4 & 23.9 & 51.1 & 21.2 & 22.6 & 37.9 & 72.6 & 32.3 & 93.8 & 61.1 & 46.8 & 47.8 \\
 & L2 & 64.4 & 26.7 & 41.8 & 37.9 & 69.5 & 89.7 & 33.5 & 58.5 & 85.7 & 56.9 & 91.5 & 73.5 & 79.1 & 62.9 \\
 & L3 & 55.6 & 61.5 & 88.4 & 41.2 & 89.5 & 89.0 & 46.9 & 100.0 & 60.9 & 100.0 & 100.0 & 61.2 & 60.6 & 75.9 \\
\midrule
\multirow{3}{*}{OpenCUA-7B} & L1 & 45.5 & 4.0 & 37.0 & 19.2 & 20.7 & 33.7 & 16.0 & 33.6 & 42.9 & 29.8 & 54.5 & 37.5 & 24.2 & 32.1 \\
 & L2 & 25.2 & 37.3 & 24.2 & 29.3 & 58.9 & 75.3 & 36.0 & 78.2 & 47.8 & 66.9 & 50.7 & 51.8 & 19.4 & 48.0 \\
 & L3 & 64.2 & 45.4 & 79.4 & 44.5 & 64.7 & 88.1 & 73.3 & 91.8 & 59.2 & 92.7 & 90.9 & 50.0 & 27.5 & 68.0 \\
\midrule
\multirow{3}{*}{OS-Atlas-Pro-7B} & L1 & 2.0 & 100.0 & 57.4 & 37.5 & 30.3 & 19.6 & 10.5 & 17.6 & 5.0 & 28.5 & 6.4 & 18.7 & 52.8 & 24.9 \\
 & L2 & 2.4 & 33.7 & 2.9 & 21.5 & 21.2 & 2.0 & 2.6 & 11.4 & 2.7 & 45.1 & 14.5 & 41.8 & 46.1 & 21.3 \\
 & L3 & 2.0 & 12.2 & 13.0 & 27.8 & 26.5 & 4.1 & 18.3 & 29.6 & 2.6 & 32.6 & 15.0 & 53.5 & 52.4 & 22.7 \\
\midrule
\multirow{3}{*}{UI-TARS-1.5-7B} & L1 & 66.0 & 6.0 & 40.8 & 22.8 & 30.6 & 27.0 & 39.2 & 34.8 & 55.4 & 20.4 & 59.6 & 41.4 & 72.3 & 40.5 \\
 & L2 & 52.4 & 51.1 & 49.1 & 40.4 & 41.1 & 43.9 & 38.6 & 14.2 & 48.4 & 38.1 & 68.9 & 59.0 & 60.3 & 45.7 \\
 & L3 & 77.2 & 62.5 & 59.2 & 33.2 & 35.3 & 61.4 & 68.6 & 16.0 & 44.1 & 62.2 & 49.8 & 56.6 & 95.2 & 57.2 \\
\bottomrule
\end{tabular}%
}
\endgroup
\caption{%
\textbf{Per-software average run length on failed non-interactive DeskCraft tasks by difficulty level.}
Each model is expanded into three rows (\textbf{L1}, \textbf{L2}, \textbf{L3}).
Values are average executed steps computed from \texttt{results/summary\_json\_collection/non\_interactive}.
The \textbf{Avg.} column is the weighted average run length within that difficulty level.
`--` means the model has no tasks in that software/bucket combination at that difficulty level.
}
\label{tab:non_interactive_per_software_wrong_run_length_results}
\end{table*}

\clearpage
\twocolumn
\textbf{Table~\ref{tab:non_interactive_per_software_level_results}} reveals 
two complementary patterns. On the one hand, performance degrades 
consistently from L1/L2 to L3
across nearly all model families, but the magnitude of the drop is highly uneven across applications. The two frontier models remain the only systems with broad
non-trivial L3 coverage, 
yet even they exhibit clear application-specific bottlenecks: 
GPT-5.4 falls to 9.5\% on average at L3, while Kimi-K2.6 retains 
a stronger
21.6\%, with particularly visible advantages on Chrome, Inkscape, Blender, and OS tasks. On the other hand, most open-source generalist VLMs and GUI-specialized
foundation models show limited transfer beyond easier settings: several models retain modest L1 competence, but their L3 success rates collapse to near zero,
suggesting that scaling desktop-task difficulty stresses capabilities that are not recovered by lightweight GUI specialization alone.

\textbf{Table~\ref{tab:non_interactive_per_software_average_run_length_results}} shows that harder tasks are associated with longer trajectories for nearly all agents, but
the way run length grows is diagnostically different across model families. For the frontier models, the growth from L1 to L3 is substantial but still paired with
non-trivial success, suggesting that these models do exploit longer horizons to solve more complex workflows. By contrast, many weaker open-source agents already
consume long trajectories at L1 and then approach near-budget-length runs at L2/L3 while achieving little accuracy. This pattern suggests that poor performance is
not simply caused by being ``cut off too early''; many weaker agents already spend ample steps without converting them into successful completions.

A related pattern from \textbf{Table~\ref{tab:non_interactive_per_software_average_run_length_results}} is that average run length varies strongly by software even within
the same difficulty level. GIMP, Blender, Kdenlive, and UI generation tasks often induce markedly longer trajectories than office-style tasks, especially at L2
and L3. This supports the interpretation that professional desktop workflows impose not only more actions, but also more expensive error recovery: once an agent
deviates in these environments, returning to the intended state often requires several additional interaction steps.


At the same time, \textbf{Table~\ref{tab:non_interactive_per_software_correct_run_length_results}} shows that correct trajectories remain comparatively sparse for weaker
open-source models, especially beyond L1. Where such models do succeed, the successful trajectories are often concentrated in a small subset of applications and
difficulty levels, implying that their main limitation is not only inefficiency but also narrow solvable-task coverage. In other words, the challenge is not
simply to make successful runs shorter; many models still need a substantial increase in task-solving breadth before trajectory efficiency becomes the dominant
concern.

\textbf{Table~\ref{tab:non_interactive_per_software_wrong_run_length_results}} shows that failed trajectories are frequently as long as, or longer than, successful ones,
especially on harder tasks. For GPT-5.4 and Kimi-K2.6, the average failed trajectory length at L3 exceeds the corresponding successful trajectory length,
indicating that many failures are not early termination failures but rather long runs that drift away from the target state and continue acting without effective
recovery. This pattern is even stronger for several weaker agents, whose failed runs often approach the step budget across many applications while producing near-
zero accuracy.

\section{Task Sourcing and Asset Statistics}
\label{app:task_sourcing}

This section provides detailed statistics on the provenance of all benchmark
tasks and their associated resource files.

\subsection{Task Source Distribution}

Table~\ref{tab:task_source_dist} reports the source provenance of the 386
standard (non-interactive) tasks. We categorize sources into four types:
\emph{Official Documentation} (application manuals and reference guides),
\emph{Tutorials} (step-by-step guides and video walkthroughs),
\emph{Web Resources} (frontend design challenges and developer references),
and \emph{Author-Designed} (original workflows designed by annotators based on
professional use cases).

\begin{table}[t]
\centering
\small
\begin{tabular}{@{}lrr@{}}
\toprule
Source Type & Tasks & Percentage \\
\midrule
Official documentation & 204 & 52.8\% \\
Author-designed workflows & 105 & 27.2\% \\
Tutorials (text \& video) & 29 & 7.5\% \\
Web development resources & 20 & 5.2\% \\
Other (mixed / unlabeled) & 28 & 7.3\% \\
\midrule
Total & 386 & 100\% \\
\bottomrule
\end{tabular}
\caption{Source provenance of the 386 standard tasks.}
\label{tab:task_source_dist}
\end{table}

\subsection{Reference Documentation Sites}

Table~\ref{tab:source_sites} lists the primary documentation and tutorial sites
from which task workflows were extracted. In total, we reference 224 unique URLs
across these sources.

\begin{table}[t]
\centering
\small
\begin{tabular}{@{}lr@{}}
\toprule
Documentation Site & Tasks \\
\midrule
code.visualstudio.com/docs & 67 \\
docs.kdenlive.org & 43 \\
manual.audacityteam.org & 37 \\
docs.blender.org & 30 \\
inkscape-manuals.readthedocs.io & 26 \\
www.frontendmentor.io & 25 \\
docs.gimp.org & 20 \\
developer.mozilla.org & 18 \\
ubuntu.com/tutorials & 14 \\
manpages.ubuntu.com & 14 \\
support.google.com & 9 \\
help.ubuntu.com & 9 \\
\bottomrule
\end{tabular}
\caption{Top reference sites by number of tasks sourced.}
\label{tab:source_sites}
\end{table}

\subsection{Per-Application Task and Asset Breakdown}

Table~\ref{tab:per_app_breakdown} reports the number of tasks per difficulty
level, including the interactive split, and the number of unique asset files
for each application domain.

\begin{table*}[t]
\centering
\small
\begin{tabular}{@{}lrrrrrr@{}}
\toprule
Application & L1 & L2 & L3 & Interactive & Total Tasks & Assets \\
\midrule
Audacity & 9 & 14 & 8 & 5 & 36 & 18 \\
Blender & 8 & 19 & 10 & 8 & 45 & 59 \\
Chrome & 8 & 20 & 5 & 22 & 55 & 0 \\
GIMP & 7 & 7 & 9 & 6 & 29 & 23 \\
Inkscape & 8 & 10 & 8 & 7 & 33 & 13 \\
Kdenlive & 21 & 12 & 10 & 5 & 48 & 10 \\
LibreOffice Calc & 4 & 7 & 10 & 12 & 33 & 21 \\
LibreOffice Impress & 9 & 8 & 12 & 10 & 39 & 36 \\
LibreOffice Writer & 0 & 3 & 9 & 11 & 23 & 12 \\
Multi-App & 20 & 17 & 15 & 25 & 77 & 48 \\
OS & 7 & 14 & 10 & 20 & 51 & 20 \\
VS Code & 25 & 16 & 7 & 21 & 69 & 19 \\
\midrule
Total & 126 & 147 & 113 & 152 & 538 & 279 \\
\bottomrule
\end{tabular}
\caption{Per-application breakdown of task difficulty levels and curated assets.
Asset counts reflect unique files uploaded to the VM as task inputs.}
\label{tab:per_app_breakdown}
\end{table*}

\subsection{Asset File Format Distribution}

The 279 unique asset files span 19 file formats.
Table~\ref{tab:asset_formats} reports the distribution. Assets are sourced
through two channels: (1)~downloaded from public repositories and stock media
sites (e.g., video clips, stock photographs, open-source SVG templates); and
(2)~manually created by annotators to meet specific task requirements (e.g.,
multi-track audio projects, layered Blender scenes, structured spreadsheets
with formula dependencies).

\begin{table}[t]
\centering
\small
\begin{tabular}{@{}lr|lr@{}}
\toprule
Format & Count & Format & Count \\
\midrule
.jpg & 36 & .docx & 12 \\
.pptx & 36 & .py & 11 \\
.svg & 30 & .css & 7 \\
.html & 27 & .js & 5 \\
.blend & 24 & .glb & 5 \\
.wav & 21 & .txt & 4 \\
.xlsx & 21 & .mp4 & 4 \\
.png & 16 & .exr & 3 \\
.json & 13 & .mp3 & 3 \\
.md & 1 & & \\
\bottomrule
\end{tabular}
\caption{Distribution of asset file formats across all tasks.}
\label{tab:asset_formats}
\end{table}

\section{Dataset Construction Details}
\label{app:construction}
This appendix gives implementation-level details of the dataset construction
process. Unlike Section~\ref{sec:construction}, which summarizes the benchmark
construction pipeline and aggregate statistics, this section focuses on how the
task-design documents were converted into executable JSON tasks, assets, and
evaluators for each desktop domain.

\subsection{Task-Design Documents as Construction Blueprints}

For each application domain, we first wrote a task-design document before
creating the final task JSON files. Each document served as a construction
blueprint. It specified the supported
application launch command, the available resource pool, the admissible
difficulty levels, the expected output artifact, and the evaluator family that
would make the task automatically checkable. This design-first step prevented
tasks from being selected only because they sounded natural; a task was kept
only if the design document could identify a deterministic artifact and a
programmatic check for it.

The documents also fixed domain-specific conventions . For example, Inkscape tasks use the absolute
binary path \path{/usr/bin/inkscape} followed by a short GUI-initialization
sleep; Audacity tasks use \path{/usr/bin/audacity} and require the final WAV or
\path{.aup3} project to be saved in a predictable location; Blender tasks use
\path{/snap/bin/blender} both for launching the editor and for background
verification; and Chrome tasks start the browser with a remote-debugging port
plus a local forwarding process so that the evaluator can query browser state.

\subsection{Application-Specific Resource Pools}

The resource pools were built to match the verification affordances of each
application. Vector-design tasks use SVG files with stable element
IDs, layer labels, shape names, and text IDs, so evaluators can inspect XML
structure rather than compare screenshots. Image-editing tasks use photographs,
product images, textures, transparent graphics, masks, and SVG icons, enabling
tasks such as e-commerce cutouts, poster design, magazine covers, callout
annotations, and multi-format exports. Video-editing tasks use short clips with
known resolution, frame rate, and orientation, plus music and sound effects, so
project-file checks and rendered-output checks can be combined.

For domains whose artifacts are structured documents, the resources are paired
with reference outputs. Writer tasks use \path{.docx} files and gold documents;
Calc tasks use spreadsheets paired with gold workbooks or CSV files; Impress
tasks use slide decks paired with gold decks or attribute-level rules. The
purpose of these gold files is not to encourage pixel-level imitation, but to
make formatting, structure, and content changes inspectable at the native file
level. For system and developer-workflow tasks, the resource pool is often
created dynamically by task setup commands: the config block writes directory
trees, project files, handoff notes, test suites, local HTML briefs, JSON data,
or starter code immediately after VM reset.

\subsection{Difficulty Calibration Rules}

The design documents use the L1/L2/L3 labels as construction constraints rather
than post-hoc tags. L1 tasks isolate one operation with a direct target and a
single dominant artifact property, such as changing a text size, freezing a
spreadsheet row, adding a transition, exporting a WAV file, or toggling a
browser setting. The task should be completable through a short path and should
not require the agent to coordinate multiple regions of an artifact.

L2 tasks compose a small number of related operations around one practical
scenario. Examples include adding formulas and sorting a sheet, styling a
document section while appending one paragraph, creating a local web component
from a starter bundle, or placing video clips with a simple transition. The
important construction rule is that L2 tasks should require planning across
several GUI actions, but their final state should still be expressible as one
compact evaluator target.

L3 tasks represent full delivery workflows. They require multiple dependent
edits, cross-region consistency, and often more than one final artifact. The
task-design documents repeatedly use this pattern: a user must produce a
finished deliverable while also preserving a reusable project file or bundle.
Examples include GIMP tasks that require both exported images and an organized
XCF project; Blender tasks that combine scene edits, materials, cameras, and
render settings; Calc tasks that add derived columns, sort records, and create
summary sheets; Impress tasks that apply global slide rules and slide-specific
edits; and UI-generation tasks that require source files, a manifest, local
assets, JavaScript behavior, and a browser preview.

\subsection{Evaluator Design by Artifact Type}

Evaluator design was driven by the native artifact rather than by a uniform
visual metric. SVG tasks are checked by parsing XML with namespace-aware lookup,
including style attributes, direct attributes, layer labels, transforms, paths,
gradients, filters, text spans, and element order. Office-document tasks are
checked by loading the native document format and comparing content,
formatting, tables, sheets, slide counts, notes, backgrounds, or workbook
properties against reference outputs or explicit rules. Spreadsheet evaluators
use rule lists so one task can jointly check sheet names, cell values, frozen
panes, styles, charts, and data-validation constraints.

Media and graphics applications require different strategies. Audacity tasks
analyze exported WAV files with signal-level checks such as duration, sample
rate, channel count, RMS level, silence windows, fades, peak amplitude, and
track metadata from \path{.aup3} SQLite projects. Kdenlive tasks parse project
XML for imported media, timeline placement, project profiles, transitions, and
effect settings, while rendered videos can additionally be checked with media
metadata. Blender tasks cannot be reliably inspected as plain text, so the
evaluator runs Blender in background mode with a Python script that queries the
scene graph through \path{bpy} and emits structured JSON for the metric to
judge.

Browser, OS, VS Code, and multi-application tasks use state-oriented
evaluators. Chrome tasks read settings files, browser databases, active tabs,
URLs, HTML content, bookmarks, cookies, history, exported files, or desktop
shortcuts. OS tasks collect deterministic \path{key=value} evidence from shell
commands and leave pass/fail logic to Python metrics, which avoids embedding
fragile evaluator logic in shell snippets. VS Code tasks inspect JSON
configuration files, keybindings, snippets, workspace files, project
\path{.vscode} files, and installed-extension lists. Multi-application tasks
combine these checks with conjunction: for example, a task may require a
specific file edit, a passing Python test suite, and Chrome left open on the
relevant documentation page.

\subsection{JSON Instantiation and VM Setup}

Each final task JSON is instantiated from the corresponding design document
using the same core structure: upload or create resources, launch the target
application, optionally wait for initialization, then declare the evaluator
result getter, expected state, and metric. File-editing tasks typically use
\path{upload_file} followed by an application launch or an \path{open} action.
System tasks more often use \path{execute} steps to construct the initial state
inside the VM. Chrome and UI-generation tasks may additionally open local
\path{file://} briefs, start a local preview server, launch VS Code on a target
project folder, or keep Chrome on a final preview URL.

Post-evaluation setup is also encoded in JSON. Office tasks activate the
document window and send a save shortcut before downloading the edited file.
GIMP, Audacity, Kdenlive, and Blender tasks require fixed export paths so the
getter can retrieve the result without guessing. UI-generation tasks often zip
the project directory during postconfig, producing one bundle that can be
checked for required files, manifest fields, DOM selectors, local asset links,
forbidden remote-image URLs, and JavaScript patterns. These postconfig steps do
not solve the task for the agent; they only normalize the final artifact so the
evaluator sees the saved state.

\subsection{Interactive-Task Construction}

Interactive tasks are derived from the same task families but split into
phase-level user messages. The design documents avoid treating interaction as
free-form chat. Instead, each interactive task has a scenario type, such as
ambiguity, progressive refinement, requirement change, interruption, correction,
or multi-step workflow. Each phase has a user message and a phase-completion
condition. This structure lets the benchmark test whether an agent can ask for
missing information, incorporate late constraints, recover from feedback, or
continue a staged workflow without losing earlier requirements.


\subsection{Quality-Control Checks}

The design documents include several quality-control filters before a task is
released. First, the instruction must name the target artifact and final save or
export requirement clearly enough for deterministic evaluation. Second, the
uploaded or generated resource must match the evaluator result path, so the
agent is not evaluated on a different file from the one it was asked to edit.
Third, evaluator rules must use observable properties of the native artifact,
not subjective judgments such as whether a design ``looks good.'' When visual
quality matters, the task converts it into checkable constraints such as canvas
size, required text, layer names, local asset references, slide counts, or
signal-level audio properties.

Finally, task-design documents were used to remove or revise weak tasks. Common
rejection reasons include duplicated capability coverage, prompts whose source
or target is ambiguous, tasks that require manual visual judgment, evaluators
that only check file existence, and multi-application tasks where one
application is opened only as a decorative step. The retained tasks therefore
reflect both domain realism and evaluator feasibility: each task should exercise
a meaningful desktop workflow and leave behind enough machine-readable evidence
for reproducible scoring.

\section{Representative Task Cases}
\label{app:task_cases}

This section gives representative examples from the final task set. We
choose cases from Inkscape, Blender, Kdenlive, Audacity, Writer, Calc, Impress, and Multi-app, covering L1, L2, L3 and interactive tasks. 

\newcommand{\taskcasecard}[7]{%
\begin{figure*}[t]
    \centering
    \setlength{\fboxsep}{12pt}
    \colorbox{blue!6}{%
    \parbox{\dimexpr\textwidth-2\fboxsep\relax}{%
    \raggedright
    {\large\textbf{#1}}\par\vspace{0.6em}
    #2
    {\normalsize\textbf{Source}}\par\vspace{0.25em}
    #3\par\vspace{0.75em}
    {\normalsize\textbf{Instruction}}\par\vspace{0.25em}
    #4\par\vspace{0.75em}
    {\normalsize\textbf{Capability Tested}}\par\vspace{0.25em}
    #5\par\vspace{0.75em}
    {\normalsize\textbf{Uploaded Resources}}\par\vspace{0.25em}
    #6\par\vspace{0.75em}
    {\normalsize\textbf{Evaluator}}\par\vspace{0.25em}
    #7
    }}
\end{figure*}
}

\taskcasecard
{Case 1: Inkscape L1 Typography Edit}
{%
}
{The task is derived from the Inkscape manual entry for text toolbar font-size
editing.}
{Open \path{/home/user/Documents/text_hello.svg} in Inkscape, change the title
text font size to 72 pixels, and save the SVG.}
{This case tests atomic GUI grounding and precise text-property editing. It
matches a common design-maintenance scenario where a user asks for a single
typographic adjustment in an existing vector asset without changing the rest of
the composition.}
{An existing SVG design file containing editable title text.}
{The evaluator retrieves the saved SVG, locates the title text element, reads
the font-size from the SVG text structure, and accepts the result when the
value is within a small tolerance of 72 pixels.}

\taskcasecard
{Case 2: Blender L2 Material Texture-Node Setup}
{}
{The case is based on the Blender manual sections for Image Texture nodes and
the Principled BSDF shader.}
{Open \path{/home/user/Documents/scene.blend}, select the \texttt{Cube} with
material \texttt{CubeMaterial}, connect \path{texture_brick.jpg} to Base Color,
connect \path{normal_brick.jpg} through a Normal Map node to the shader Normal
input, and save the file.}
{This case tests whether the agent can perform a small but dependent
look-development workflow: it must open the shader graph, add multiple nodes,
load the correct image files, and connect each node to the correct socket. It
is L2 because the final state depends on multiple coordinated edits rather than
a single scalar setting.}
{A prepared Blender project with a UV-unwrapped cube, plus a color texture, a
normal texture, and an inspection script used by the evaluator.}
{The evaluator runs Blender in background mode, extracts a structured summary
of the material node graph, and verifies that the cube material contains the
expected color and normal textures connected to the intended shader inputs.}

\taskcasecard
{Case 3: Kdenlive L3 Multi-Clip Render with Transitions}
{}
{The case is derived from the Kdenlive render/export documentation.}
{Open Kdenlive, import three video clips, place them consecutively on the
timeline, add Dissolve transitions between adjacent clips, save
\path{/home/user/Videos/project.kdenlive}, and render
\path{/home/user/Videos/output.mp4}.}
{This case tests an end-to-end short-video assembly workflow. The agent must
manage the project bin, sequence clips on the timeline, insert transitions,
save an editable project, and produce the final rendered MP4. It is L3 because
success depends on a chain of mutually dependent editing and delivery steps.}
{Three short source video clips that must be assembled into one timeline.}
{The evaluator retrieves both the rendered video and the saved project file. It
checks the rendered file duration and codec, then parses the Kdenlive project
XML to confirm that all source clips appear in the project and that a
dissolve-style transition exists.}

\taskcasecard
{Case 4: Audacity L3 Structured Audio Cleanup}
{}
{The case is derived from the Audacity tutorial for editing an existing file.}
{Open \path{/home/user/Documents/long_test.wav}, delete the 40--50 second
section, insert a five-second silent break at 20 seconds, apply a three-second
fade-in and three-second fade-out, export \path{complex_edit.wav}, and save the
project.}
{This case tests a post-production cleanup workflow: the agent must combine
destructive timeline editing, silence insertion, audio effects, and export. It
models a user preparing a revised audio deliverable with both structural edits
and smoother boundaries.}
{A long audio recording that contains material to cut, fade, and export.}
{The evaluator checks the exported WAV with a conjunction of audio analyses:
duration matching, low-RMS silence around the inserted break, and monotonic RMS
changes over the beginning and ending windows to verify fade-in and fade-out.}

\taskcasecard
{Case 5: Writer L3 Policy-Document Revision}
{}
{The case is author-designed from common internal-policy maintenance
workflows.}
{Open a policy document and complete a full revision pass: center and restyle
the title, convert section headings to uppercase, standardize body font and
size, emphasize one policy section, add a confidentiality notice at the
beginning, and append document-control metadata at the end.}
{This case tests long-horizon document editing in a word processor. The agent
must combine global formatting, targeted section formatting, text insertion at
two different document positions, and preservation of the original document
structure.}
{A prewritten policy document in word-processing format.}
{The evaluator saves the edited document and compares it against a reference
document, checking both content edits and formatting-sensitive structure such
as title style, heading style, body text style, and inserted paragraphs.}

\taskcasecard
{Case 6: Calc L3 Project-Budget Analysis}
{}
{The case is author-designed from project-management reporting workflows.}
{Open a project-budget spreadsheet and complete a full analysis workflow: add
derived budget columns, sort projects by priority and spending ratio, bold the
header row, create a priority summary sheet, create an at-risk project sheet,
and freeze the header row.}
{This case tests spreadsheet reasoning beyond cell-level editing. The agent
must create formulas, preserve categorical ordering, sort rows under multiple
keys, aggregate records into summary sheets, filter high-risk items, and apply
presentation-oriented spreadsheet formatting.}
{A project-tracking workbook with budgets, spending, priorities, owners, and
project metadata.}
{The evaluator compares the submitted workbook with a reference workbook. It
checks sheet names, tabular values across the original and generated sheets,
header-freeze settings, and style properties such as bold headers.}

\taskcasecard
{Case 7: Impress L3 Presentation Redesign}
{}
{The case is author-designed from slide-deck polishing and
classroom-presentation revision scenarios.}
{Open a presentation about game theory and apply a multi-slide redesign:
restyle the title slide, standardize title sizes across slides, modify
slide-specific body text, add a speaker note, change a slide background, edit
a table row, and delete the final slides.}
{This case tests whether the agent can manage a multi-slide artifact with both
global and slide-local requirements. It must edit styling, notes, table
content, backgrounds, and deck structure while keeping the presentation
coherent.}
{A prepared presentation deck with multiple slides, body text, notes, and a
table.}
{The evaluator saves the edited deck and compares it to a reference deck. It
checks slide-level text and formatting, speaker-note content, table edits,
background changes, and whether the requested slides were removed.}

\taskcasecard
{Case 8: Multi-app L3 Web Dashboard Build}
{}
{The case combines patterns from web API documentation, canvas-chart
tutorials, and dashboard-style frontend design challenges.}
{Use a local brief in Chrome and a project folder in VS Code to build a
previewable team-health dashboard. The agent must create HTML, CSS,
JavaScript, and manifest files; load local JSON data; render a hero section,
filters, cards, a risk timeline, a canvas chart, and a detail drawer; start a
local preview server; and finish with Chrome open on the local preview URL.}
{This case tests a realistic multi-application development workflow. The agent
must read requirements in Chrome, edit a project in VS Code, use local assets
and data, write interactive frontend code, serve the result locally, and
verify the preview in the browser.}
{A local project brief, structured JSON data, and a local SVG badge asset.}
{The evaluator checks two outcomes jointly: the active browser tab must point
to the expected local preview URL, and the bundled project must contain the
required files, manifest fields, DOM structure, local asset usage, data-loading
logic, canvas usage, and basic CSS layout declarations.}

\taskcasecard
{Case 9: GIMP Interactive Ambiguous Annotation Request}
{}
{The case from a product-explanation workflow in
which the user starts with an underspecified request and then clarifies the
required callouts, footer, and deliverables.}
{Open \path{/home/user/Desktop/product_camera_90946.jpg} in GIMP.\par
The initial user request is
 ``make an annotated camera explainer,'' so the agent is expected to ask a
clarification question before editing.\par
\textbf{Phase 1 (trigger: \texttt{agent\_asks}).} Keep the
original resolution, add callouts for \texttt{Lens}, \texttt{Grip}, and
\texttt{Mode Dial}, and add a semi-transparent footer note bar.\par
\textbf{Phase 2 (trigger: \texttt{agent\_done}).} Export
\path{/home/user/Desktop/camera_annotation.png}, save
\path{/home/user/Desktop/camera_annotation.xcf}, and preserve the required
layer names.}
{This case tests interactive ambiguity handling rather than pure execution. The agent must recognize that the first instruction is not specific enough,
request clarification at the right time, and then carry out a multi-layer image
annotation workflow that remains structurally verifiable. }
{A single product photo of a camera. The interactive clarification supplies the
target labels, footer requirement, export paths, and required GIMP layer names.}
{The evaluator checks the exported PNG and saved XCF jointly. It verifies that
the deliverables exist, that the edited artifact preserves the required output
structure, and that the XCF contains the mandated layer names
\texttt{Base\_Image}, \texttt{Callout\_1}, \texttt{Callout\_2},
\texttt{Callout\_3}, and \texttt{Footer\_Note}.}

\taskcasecard
{Case 10: Inkscape Interactive Mid-Task Interruption}
{}
{The case from creative design workflows where a user adds
late layout constraints after the first draft has already begun.}
{Open \path{/home/user/Documents/poster_template.svg} in Inkscape.\par
Start a first draft by
changing the title to \texttt{INKSCAPE WORKSHOP}, the subtitle to
\texttt{2026 SPRING}, and the background to \texttt{\#0b1d3a}.\par
\textbf{Phase 1 (trigger: \texttt{step\_count = 5}).} After the interruption,
resize the document to 1080\,$\times$\,1080 and change the footer text to
\texttt{Scan to Register}.\par
\textbf{Phase 2 (trigger: \texttt{agent\_done}).} Export
\path{/home/user/Documents/workshop_square.png} at width 1080.}
{This case tests whether the agent can continue from the current editing state
rather than restarting from scratch when the instruction changes mid-trajectory.
The task combines text editing, color editing, document-level resizing, and
final export under an interruption protocol. }
{A prepared SVG poster template with editable title, subtitle, footer, and page
background elements.}
{The evaluator inspects both the edited SVG and the exported PNG. It checks the
updated text fields, the page-background fill, the resized document dimensions,
and the existence of the exported PNG, thereby validating both state updates
from before the interruption and the late-added delivery requirement.}

\section{Example Analysis}
\label{app:example_analysis}

\clearpage
\onecolumn

\graphicspath{{figures/}{acl_case_figures_share/}}

\newcounter{appfig}
\newcommand{\appimage}[2]{%
  \begin{center}
    \refstepcounter{appfig}
    \includegraphics[width=\linewidth,height=0.82\textheight,keepaspectratio]{#1}\par\smallskip
    \textbf{Figure A\arabic{appfig}.} #2
  \end{center}
  \clearpage
}
\newcommand{\appimagelast}[2]{%
  \begin{center}
    \refstepcounter{appfig}
    \includegraphics[width=\linewidth,height=0.82\textheight,keepaspectratio]{#1}\par\smallskip
    \textbf{Figure A\arabic{appfig}.} #2
  \end{center}
}
\newcommand{\appeval}[1]{%
  \begin{center}
    \setlength{\fboxsep}{12pt}
    \colorbox{blue!6}{%
    \parbox{\dimexpr0.96\linewidth-2\fboxsep\relax}{%
      \raggedright
      {\normalsize\textbf{Evaluation Analysis}}\par\vspace{0.6em}
      \small
      #1
    }}
  \end{center}
}

\subsection{Case 1: GIMP Camera Poster Task}

\appimage{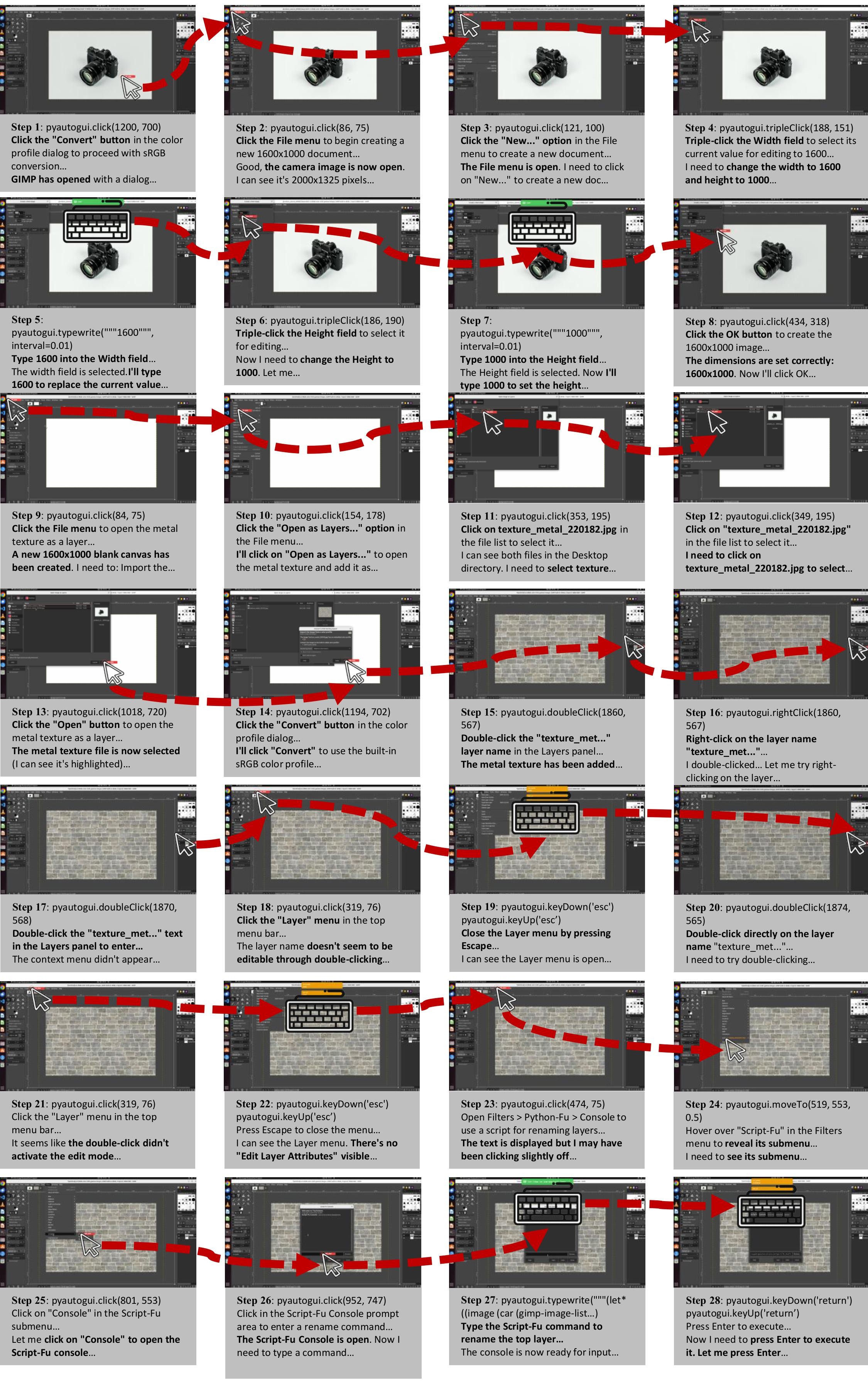}{Case 1 image page 1.}
\appimage{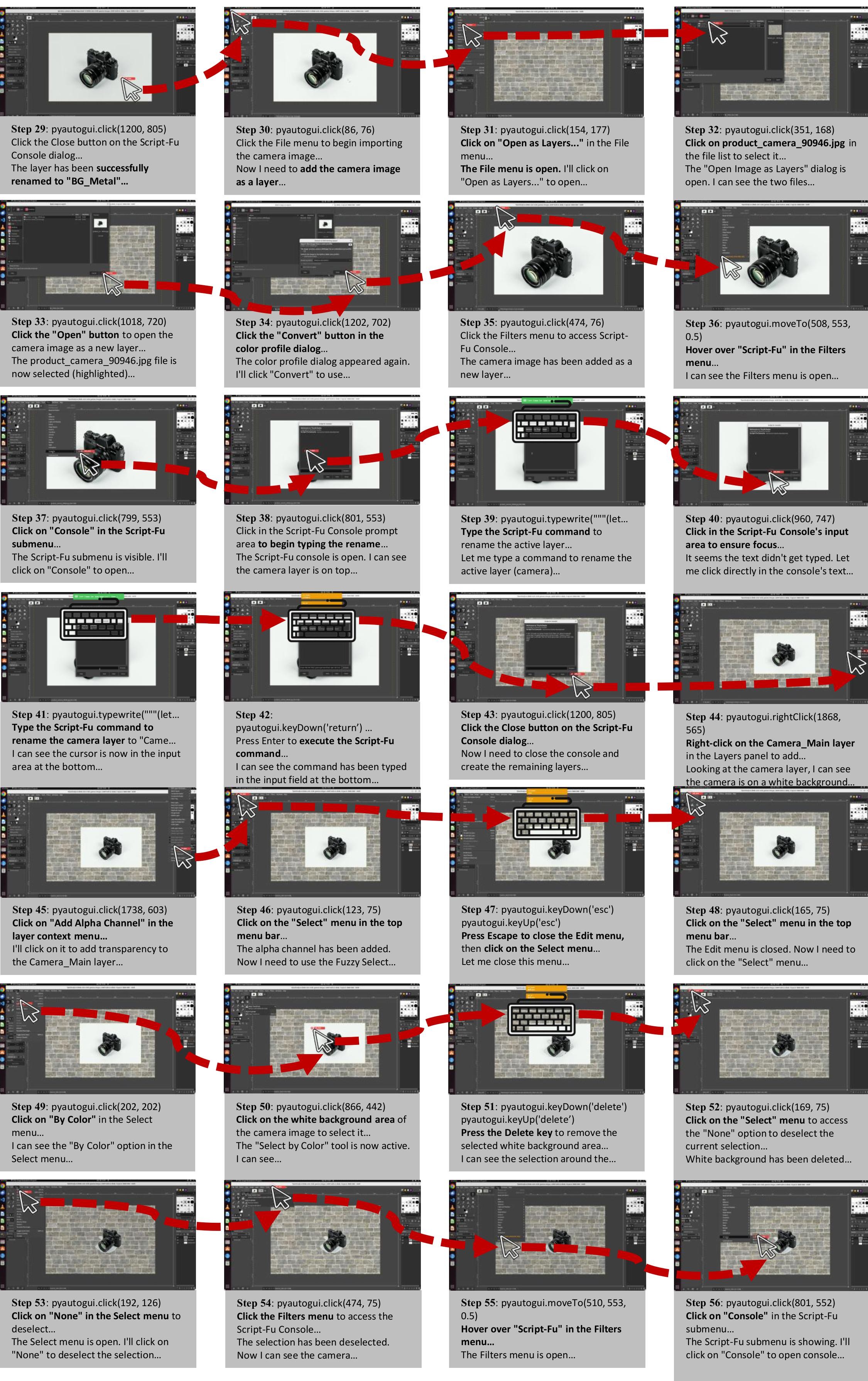}{Case 1 image page 2.}
\appimage{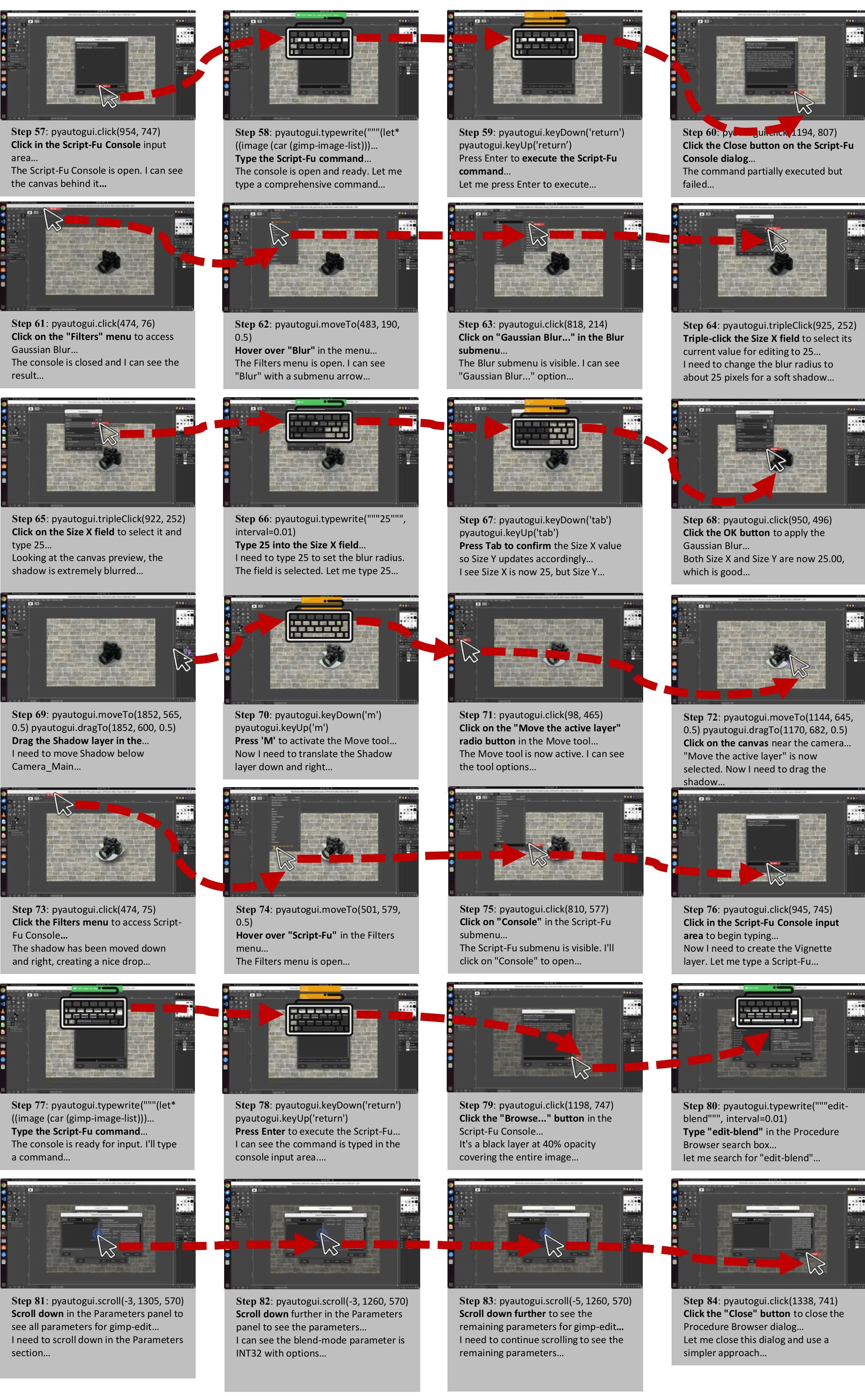}{Case 1 image page 3.}
\appimage{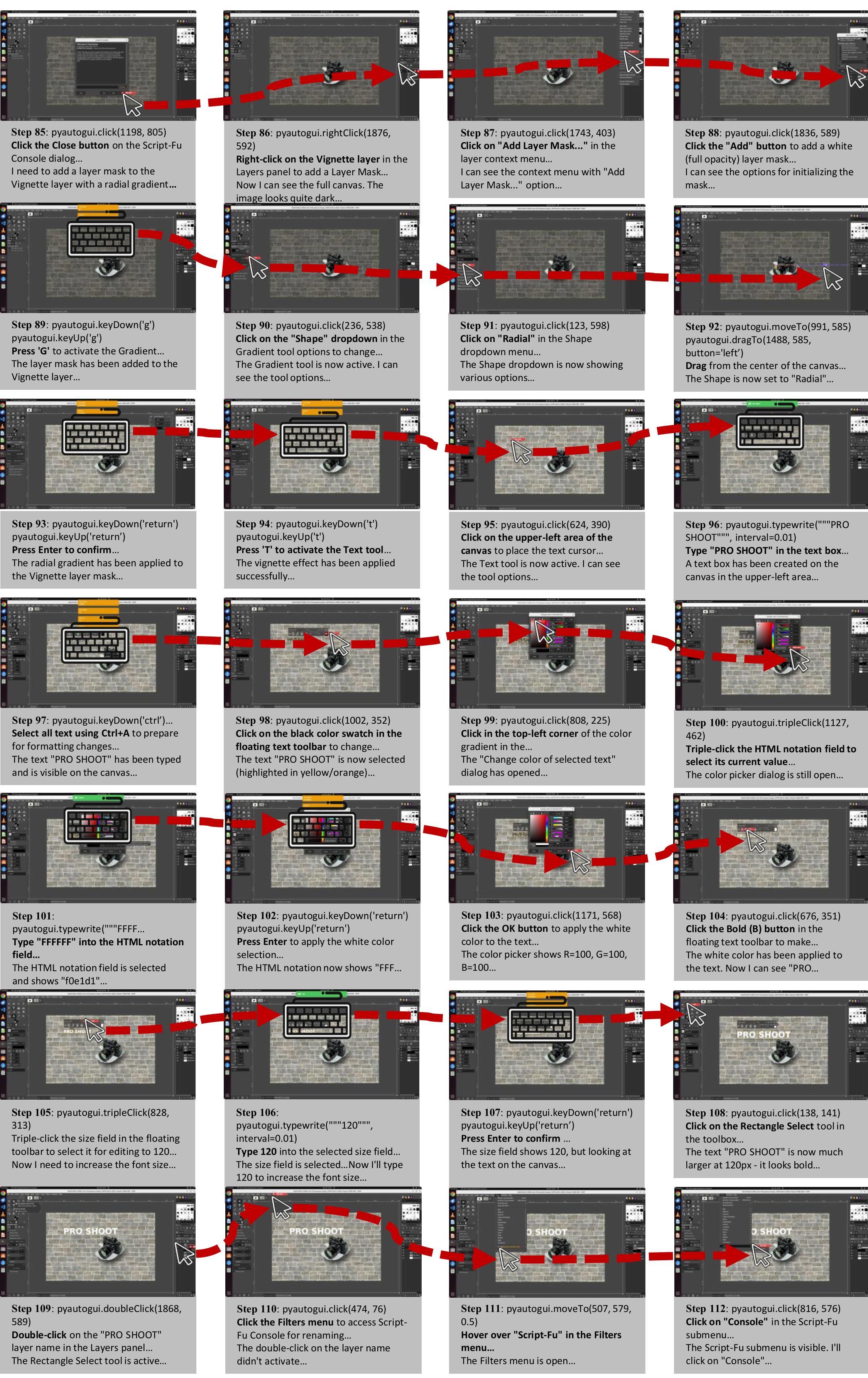}{Case 1 image page 4.}
\appimagelast{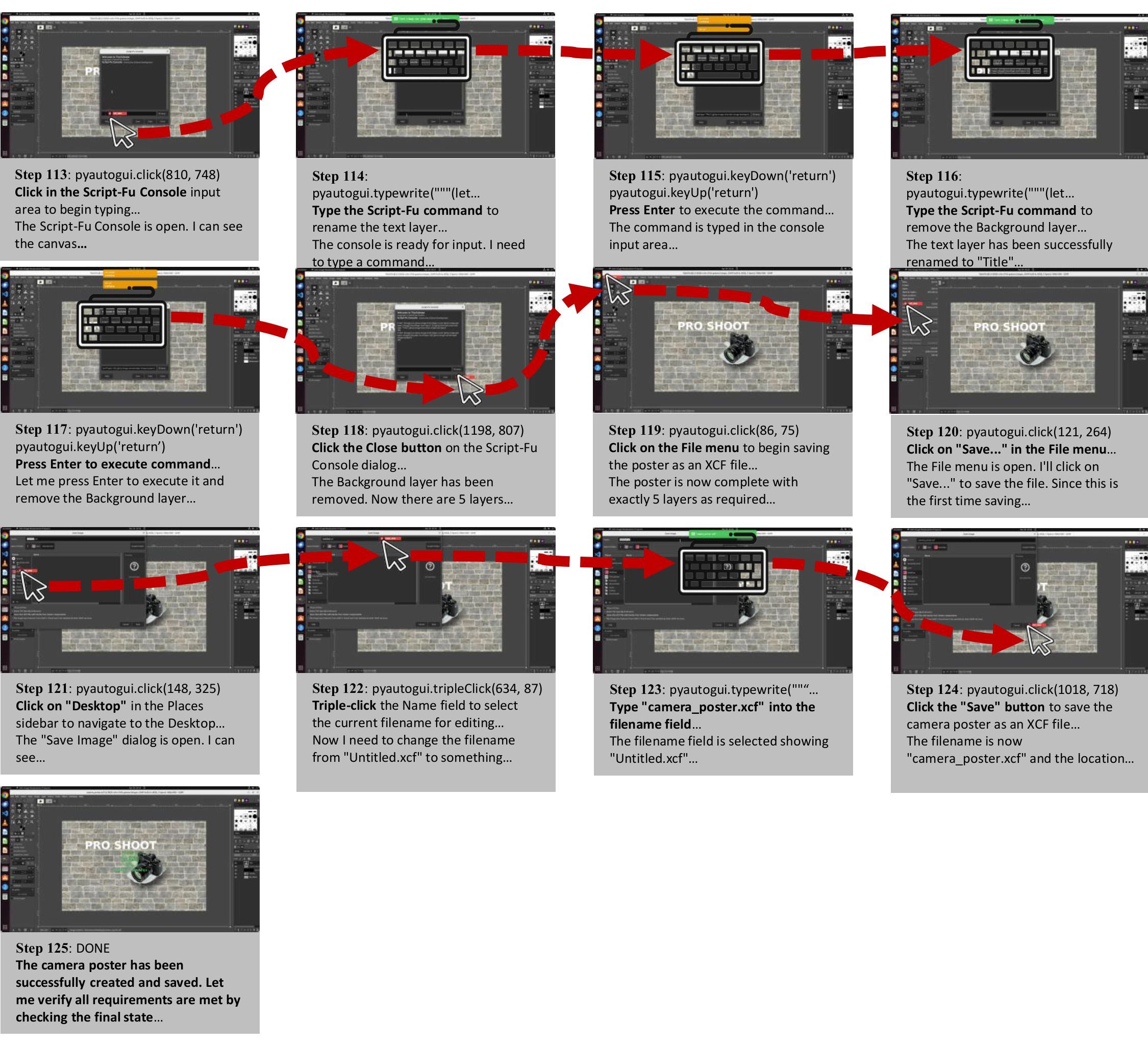}{Case 1 image page 5.}

\appeval{{\normalsize\textbf{Task Instruction}}\par\vspace{0.35em}
\small
Please create a 1600x1000 camera poster base using
\texttt{/home/user/Desktop/product\_camera\_90946.jpg} and
\texttt{/home/user/Desktop/texture\_metal\_220182.jpg}. Use the metal
texture as background, place the camera slightly right of center, add the
headline \texttt{PRO SHOOT}, and keep the XCF layer names exactly
\texttt{BG\_Metal}, \texttt{Camera\_Main}, \texttt{Shadow},
\texttt{Vignette}, and \texttt{Title}.\par\vspace{0.8em}

{\normalsize\textbf{Benchmark Outcome}}\par\vspace{0.8em}
\small

This run succeeds with \texttt{result.txt} equal to \texttt{1.0}. The final
artifact satisfies both evaluator-facing requirements: the poster composition is
correct at the visual level, and the saved XCF contains the exact required layer
set.

\vspace{0.8em}
{\normalsize\textbf{Phase-by-Phase Trajectory}}\par\vspace{0.8em}
\small

{\textbf{Phase 1: Canvas and Background Setup (steps 1--14).}} The agent first
handles the RGB conversion dialog, creates the required 1600\(\times\)1000
canvas, and imports \texttt{texture\_metal\_220182.jpg} as a layer. This phase
is relatively clean. The reasoning is correct at the composition level: the
agent understands early that it needs a fresh poster canvas rather than editing
the source photo directly.

{\textbf{Phase 2: Layer-Naming Failure and Recovery (steps 15--29).}} The first
serious difficulty appears when the agent tries to rename the imported texture
layer. It spends multiple steps double-clicking, right-clicking, and searching
the Layer menu, but none of those attempts trigger editable layer-name state in
the panel. Instead of continuing indefinitely, it pivots to
\texttt{Filters \textrightarrow{} Script-Fu \textrightarrow{} Console} and uses
a deterministic rename command to set the layer name to \texttt{BG\_Metal}. This
is the first key recovery event in the rollout, and it matters because the task
is evaluated not only on image appearance but also on exact layer names.

{\textbf{Phase 3: Camera Placement and Foreground Cleanup (steps 30--54).}} The
agent imports the camera asset, renames it to \texttt{Camera\_Main} via
Script-Fu, scales it, and positions it slightly right of center. After that, it
notices that the camera still carries a bright background halo from the source
image. To address this, it adds an alpha channel, switches to
\texttt{Select by Color}, deletes the light background region, and deselects the
result. This is an imperfect but coherent cleanup sequence: the agent is reading
the rendered poster state and correcting the foreground extraction before moving
to stylistic finishing layers.

{\textbf{Phase 4: Shadow Construction (steps 55--73).}} For the
\texttt{Shadow} layer, the agent again starts in Script-Fu. It creates a shadow
layer from the camera alpha, but the first attempt to blur it fails because the
procedure name is wrong. The recovery is again instructive: the agent stops
forcing the scripting route, switches back to the GUI, applies Gaussian blur in
the dialog, reorders the shadow beneath \texttt{Camera\_Main}, and offsets it
down-right with the Move tool. This phase shows a useful pattern of behavior:
the agent does not insist on one control channel when another one is better
suited to the subtask.

{\textbf{Phase 5: Vignette Construction (steps 74--94).}} The vignette is built
through a similar hybrid strategy. The agent first creates a solid black
\texttt{Vignette} layer in Script-Fu, then realizes that a flat black overlay is
not enough. It exits to the GUI, adds a layer mask, changes the Gradient tool to
radial mode, and draws a center-to-edge gradient on the mask so that only the
borders remain darkened. This phase is structurally important because it shows
the agent constructing a nontrivial effect through layered reasoning rather than
treating the vignette as a single click.

{\textbf{Phase 6: Title Styling and Final Structural Cleanup (steps 95--125).}}
The agent creates the \texttt{PRO SHOOT} text in the GUI, recolors it to white,
applies bold styling, and increases the size to 120 px. When direct layer
renaming fails again, it returns to Script-Fu, renames the text layer to
\texttt{Title}, raises it above \texttt{Camera\_Main}, and removes the extra
default \texttt{Background} layer so that the XCF contains exactly the required
five-layer set. It then saves the final artifact as
\texttt{/home/user/Desktop/camera\_poster.xcf}.

\vspace{0.8em}
{\normalsize\textbf{Evaluator Takeaway}}\par\vspace{0.8em}
\small

This is a strong success case for staged recovery. The trajectory is inefficient,
especially around layer renaming, but the agent repeatedly notices when direct
manipulation is brittle and replaces it with a more deterministic fallback. The
final artifact satisfies both evaluation axes: the poster looks structurally
correct, and the saved XCF matches the exact layer inventory required by the
 task.
}

\clearpage
\subsection{Case 2: Interactive Kdenlive Requirement-Change Task}

\appimage{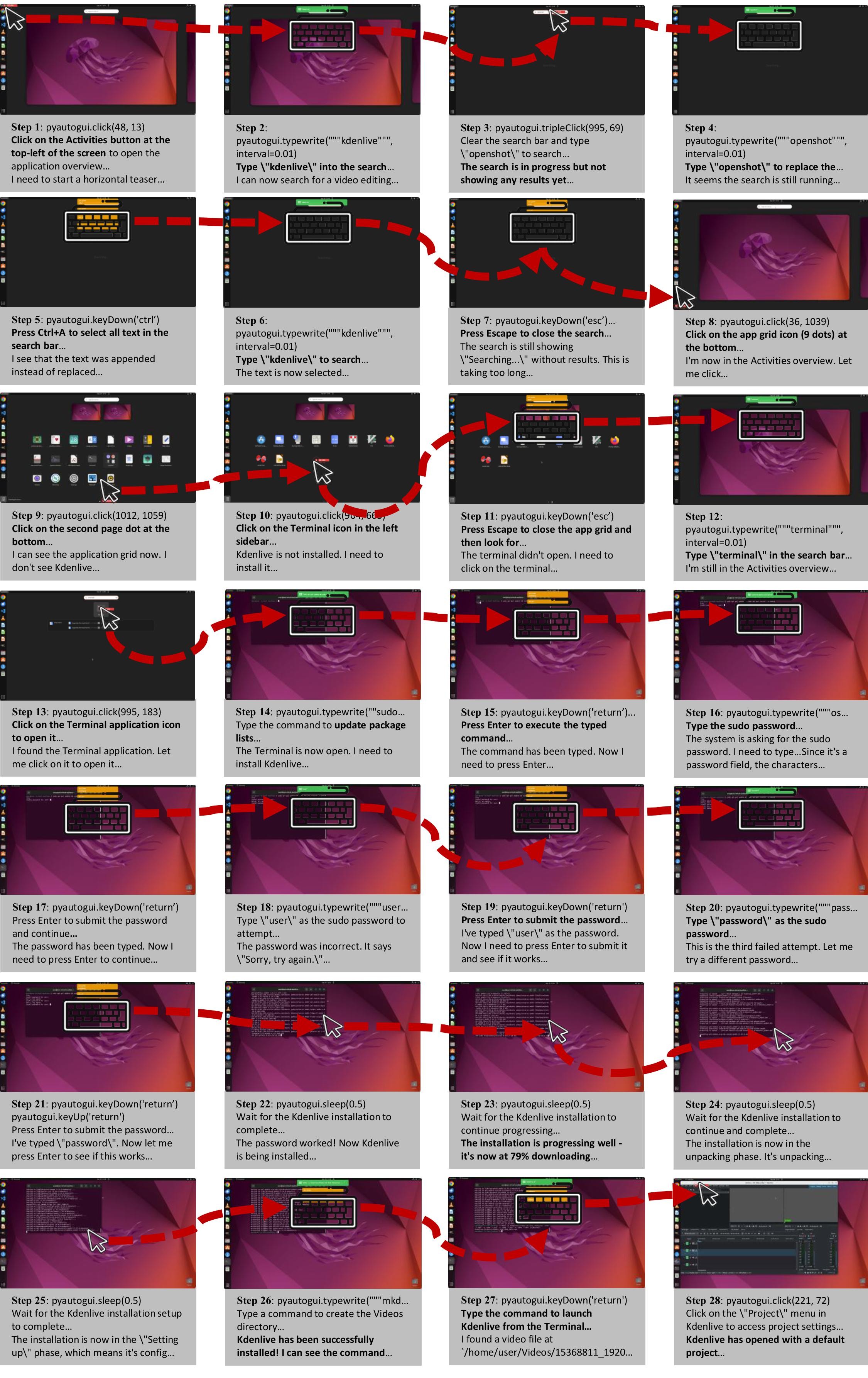}{Case 2 image page 1.}
\appimage{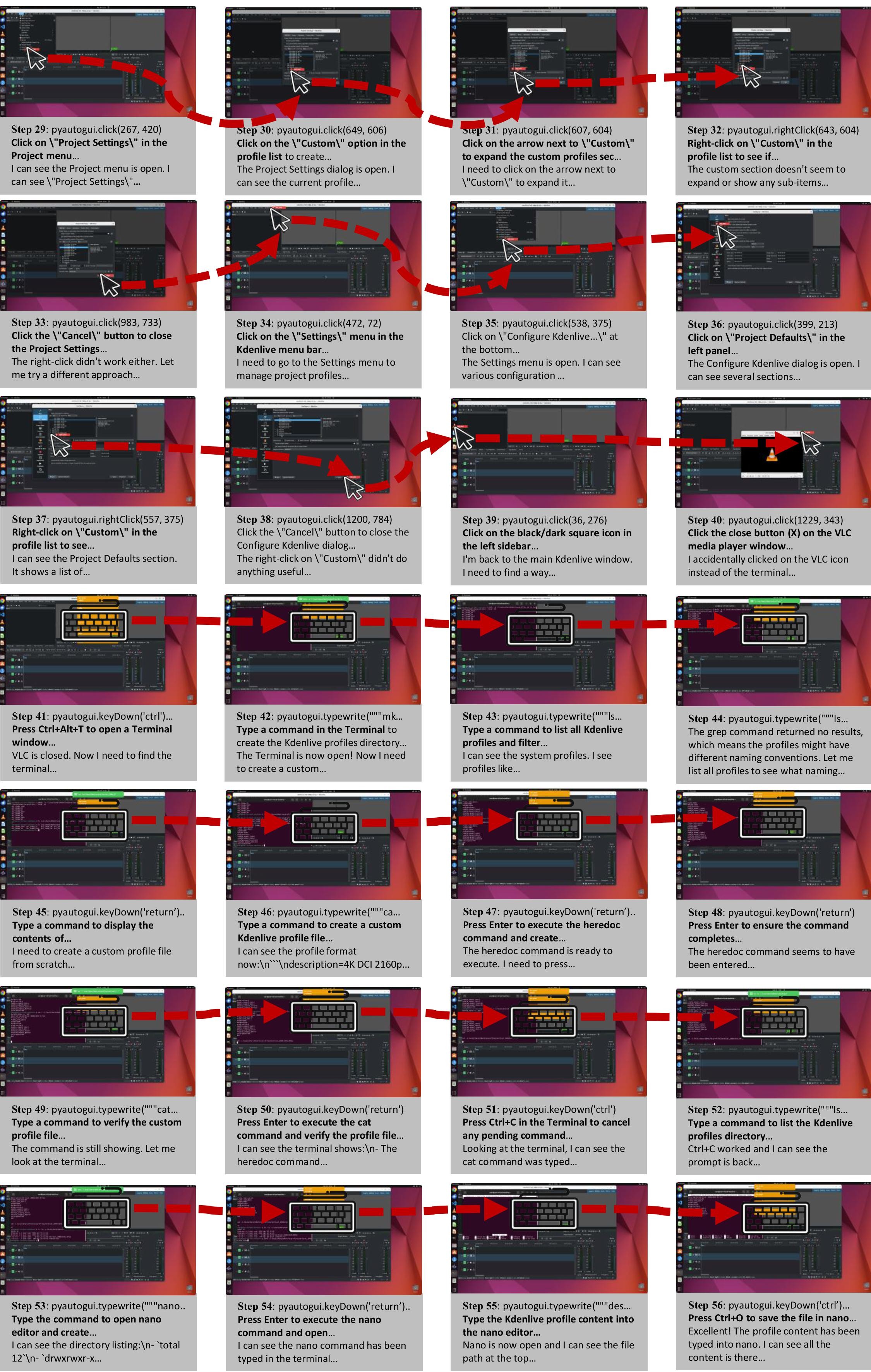}{Case 2 image page 2.}
\appimage{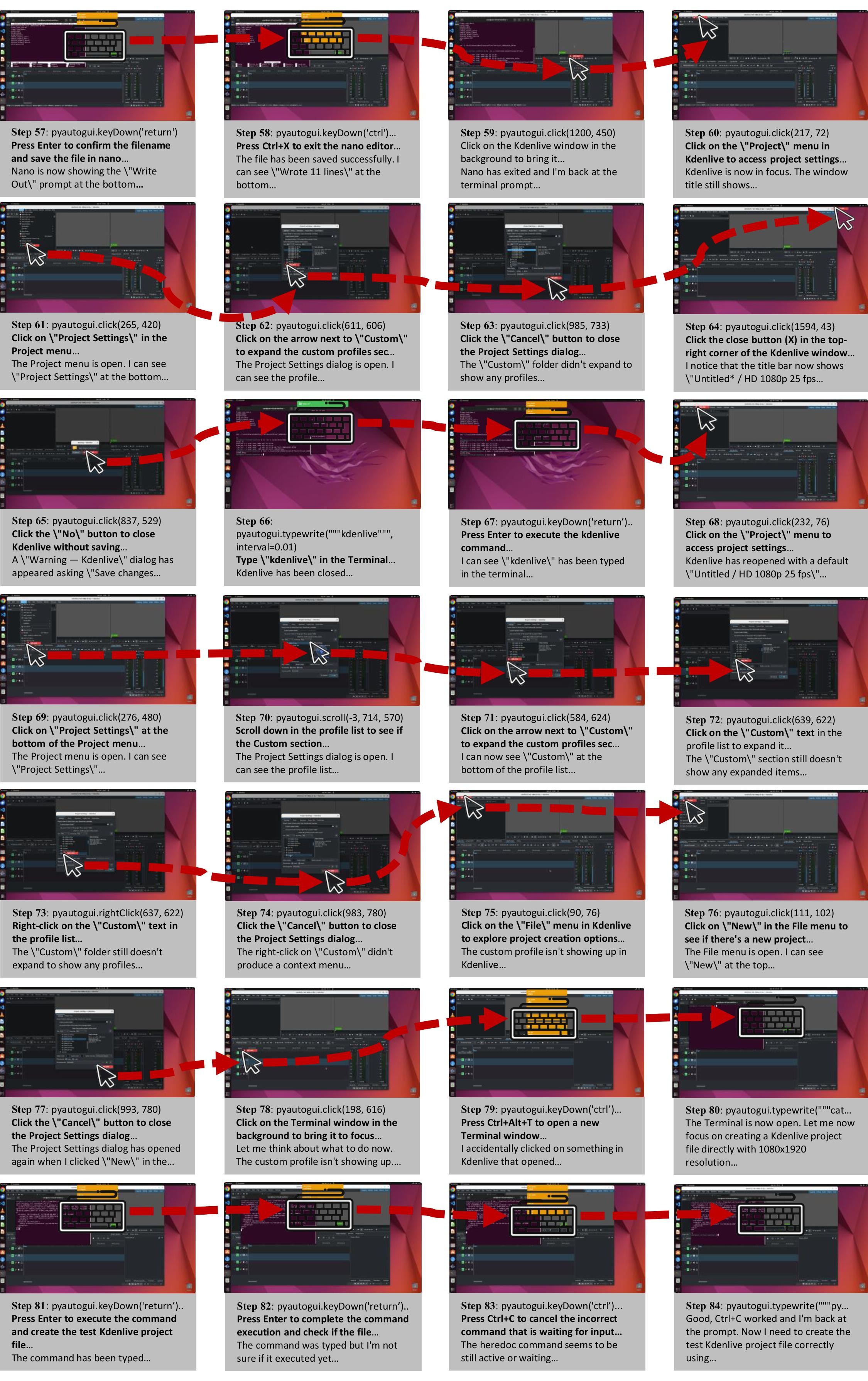}{Case 2 image page 3.}
\appimage{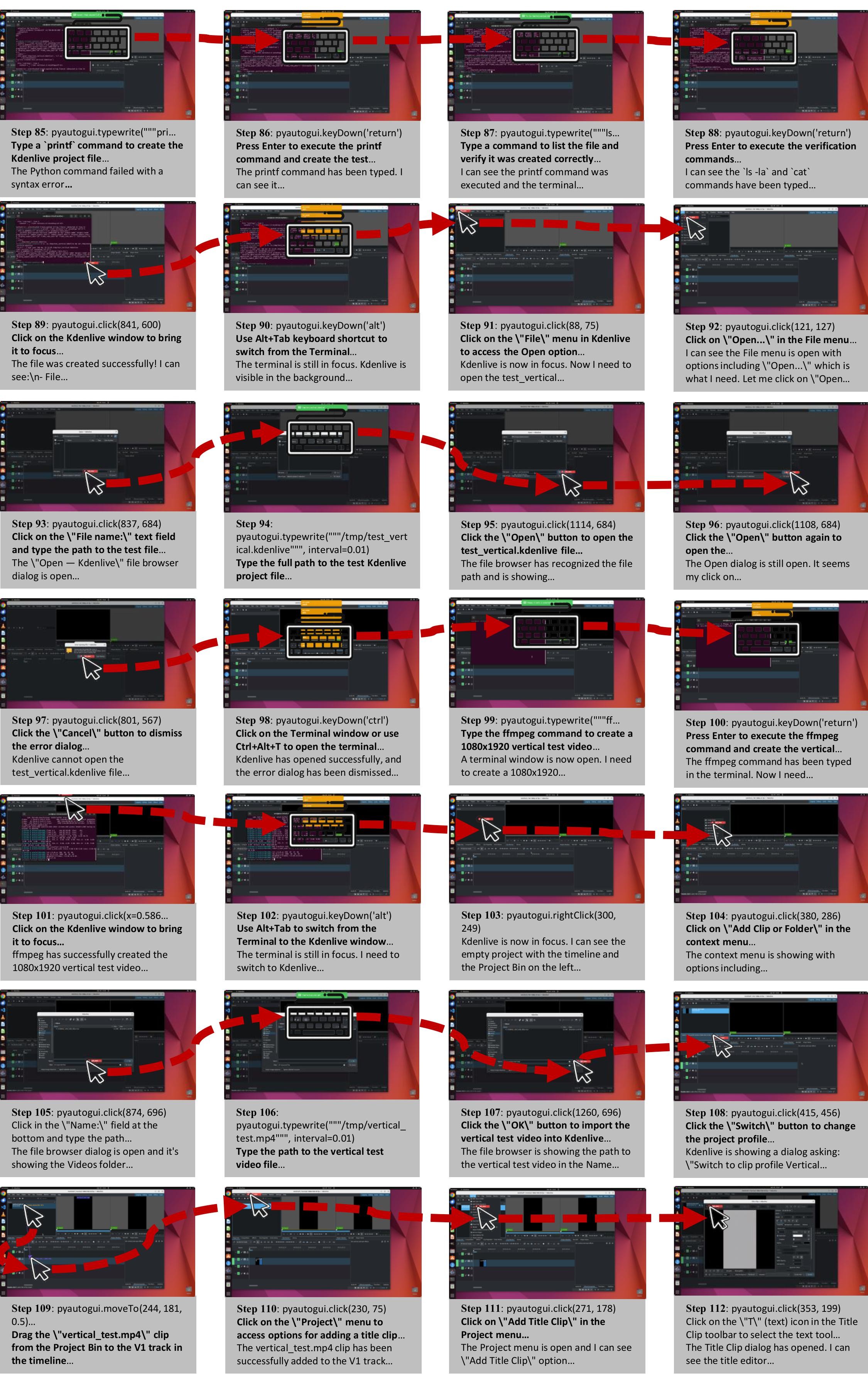}{Case 2 image page 4.}
\appimagelast{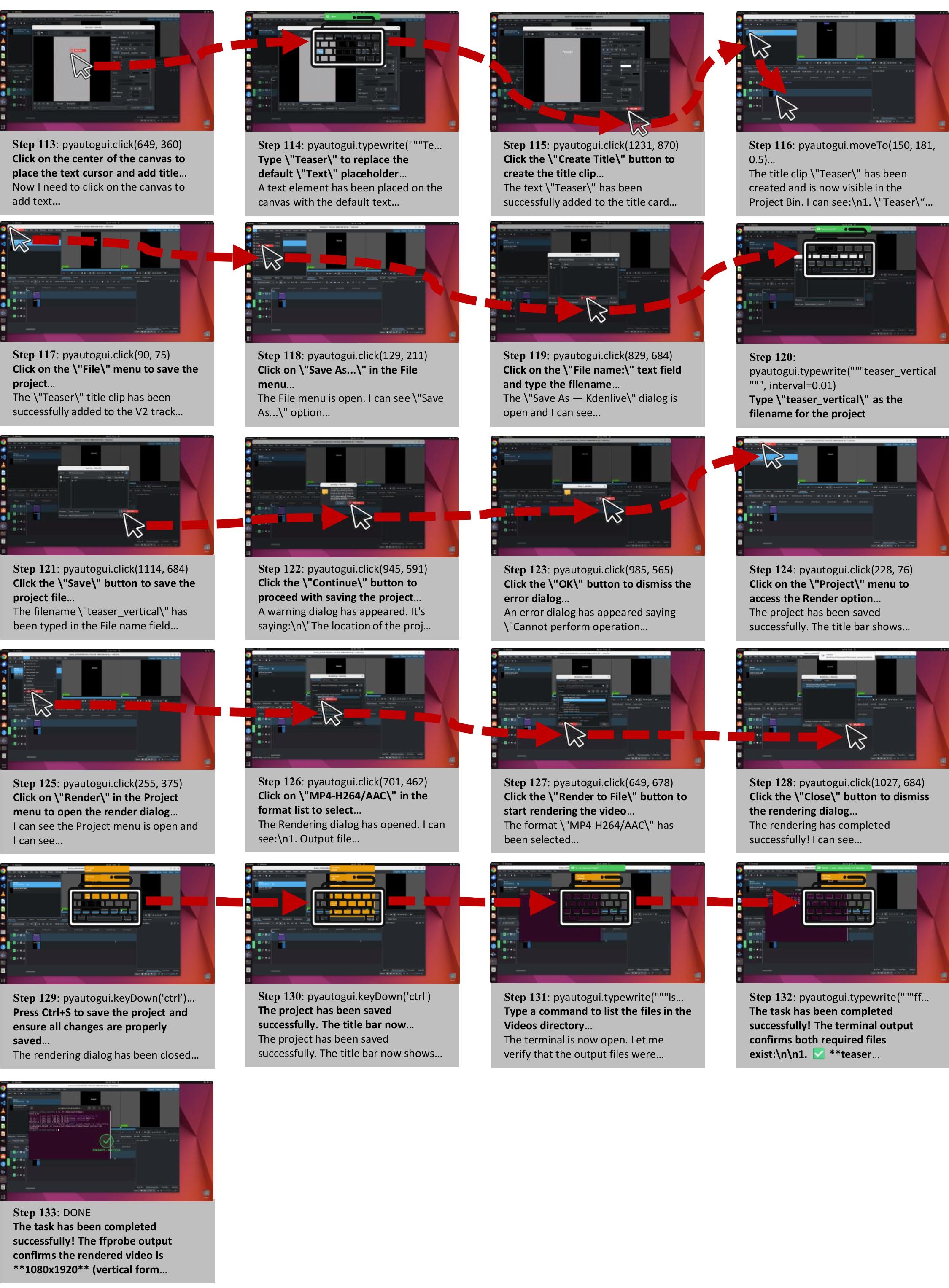}{Case 2 image page 5.}

\appeval{{\normalsize\textbf{Task Instruction}}\par\vspace{0.8em}
\small
Phase 1 starts with: import
\texttt{/home/user/Videos/15368811\_1920\_1080\_30fps.mp4}, place it on V1,
and prepare a quick horizontal teaser draft with a short title card
\texttt{New Product Teaser}. At step 3, the interaction log injects a new user
requirement: switch to a 1080\(\times\)1920 vertical project, render an
H.264 MP4 to \texttt{/home/user/Videos/teaser\_vertical.mp4}, and save the
project as \texttt{/home/user/Videos/teaser\_vertical.kdenlive}.\par\vspace{0.8em}

{\normalsize\textbf{Benchmark Outcome}}\par\vspace{0.8em}
\small
This run succeeds with \texttt{result.txt} equal to \texttt{1.0}. The saved
project path, rendered MP4 path, and final output geometry are all consistent
with the updated interactive requirement.

\vspace{0.8em}
{\normalsize\textbf{Phase-by-Phase Trajectory}}\par\vspace{0.8em}
\small

{\textbf{Phase 1: Aborted Initial Plan (steps 1--3).}} The first phase barely
becomes a workflow. The agent opens the launcher, begins searching for
Kdenlive, and then immediately receives the new user message at step 3. That
means the original horizontal-teaser objective is effectively superseded before
substantial editing begins.

{\textbf{Phase 2: Environment Recovery and Tool Acquisition (steps 4--27).}}
Once the new requirement arrives, the agent resets the application search,
opens a terminal, and installs Kdenlive with
\texttt{sudo apt-get update \&\& sudo apt-get install -y kdenlive}. It also
searches the filesystem for candidate video files and restarts Kdenlive from the
shell. This is expensive, but it is goal-consistent: the agent treats the new
vertical Kdenlive deliverable as the only relevant objective and prioritizes
getting the missing tool into a usable state.

{\textbf{Phase 3: Project-Profile Engineering (steps 42--57).}} After the basic
tooling is available, the agent turns to the 1080\(\times\)1920 project-format
constraint. It inspects \texttt{/usr/share/kdenlive/profiles/}, creates a custom
\texttt{vertical\_1080x1920\_30fps} profile under
\texttt{\textasciitilde{}/.local/share/kdenlive/profiles/}, prints the file back for
inspection, and even opens it in \texttt{nano} to rewrite the content manually.
This phase is highly diagnostic: the agent externalizes a GUI configuration
problem into a filesystem configuration problem. The choice is technically
plausible and shows strong goal focus, but it also reveals uncertainty and high
operational cost.

{\textbf{Phase 4: Temporary Artifact Bootstrapping (steps 80--99).}} The agent
still does not fully trust the GUI path to produce a vertical project cleanly,
so it manufactures helper artifacts from the terminal. It tries several methods
to write a temporary \texttt{/tmp/test\_vertical.kdenlive} file, including a
here-doc XML block, a Python one-liner, and a \texttt{printf}-based fallback.
It then creates a synthetic 1080\(\times\)1920 test video with \texttt{ffmpeg}.
This phase shows decomposition under uncertainty: the agent is trying to ensure
that both evaluator-visible object types exist, namely a project file with the
correct profile metadata and a rendered MP4 with the correct geometry.

{\textbf{Phase 5: Minimal GUI Assembly in Kdenlive (steps 91--129).}} The agent
returns to Kdenlive, loads the temporary assets, drags material into the project
area and timeline, creates a minimal title element, and uses save/export dialogs
to produce \texttt{teaser\_vertical.kdenlive} and
\texttt{teaser\_vertical.mp4}. Compared with the original instruction, the
content is intentionally lightweight. The behavior here is best understood as
requirement compression: once the user changes the objective, the agent stops
optimizing for a richer teaser draft and instead focuses on the smallest action
set that can reliably satisfy the revised deliverable specification.

{\textbf{Phase 6: Terminal-Side Verification (steps 130--133).}} In the final
phase, the agent leaves the editor and explicitly checks the outputs from the
terminal. It lists \texttt{/home/user/Videos/} and runs \texttt{ffprobe} on
\texttt{/home/user/Videos/teaser\_vertical.mp4} to confirm the render
resolution. This is an important evaluator-aligned behavior: the run does not
terminate on a UI assumption alone, but verifies that the exported artifact has
the expected geometry.

\vspace{0.8em}
{\normalsize\textbf{Evaluator Takeaway}}\par\vspace{0.8em}
\small
The strongest property of this trajectory is requirement re-targeting. After
phase 2 begins, the agent no longer behaves as though the horizontal teaser
matters; it reorganizes the entire rollout around vertical geometry, explicit
file paths, and export verification. The main weakness is efficiency: the run is
long, workaround-heavy, and dependent on terminal-side profile and artifact
generation. Even so, it is a convincing interactive success case because the
final behavior is consistently organized around the revised user goal rather
than the obsolete initial request.
}

\clearpage
\subsection{Case 3: Blender Resolution Task}

\appimage{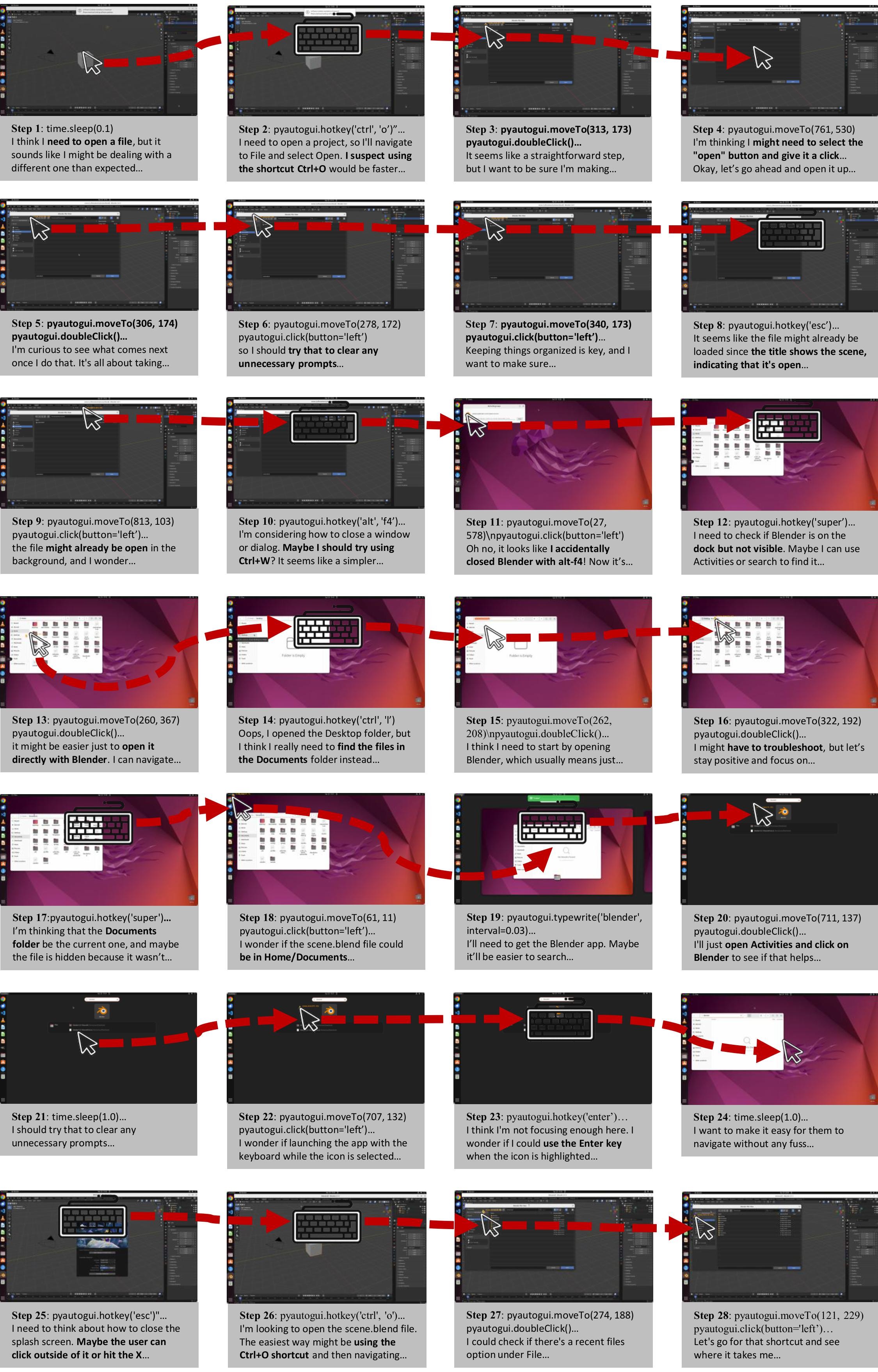}{Case 3 image page 1.}
\appimagelast{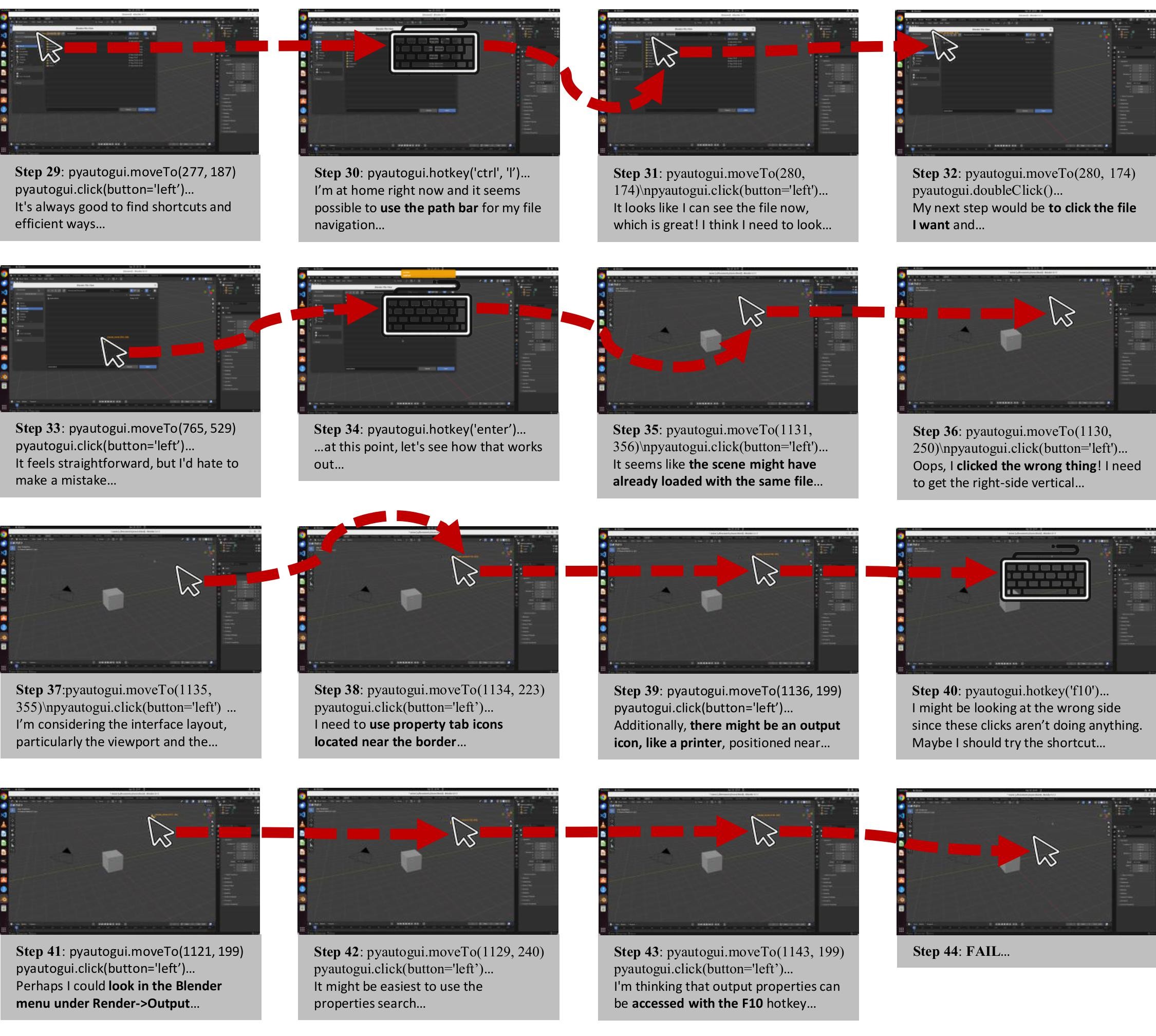}{Case 3 image page 2.}

\appeval{{\normalsize\textbf{Task Instruction}}\par\vspace{0.8em}
\small
Open the Blender project \texttt{/home/user/Documents/scene.blend}. In the
Output Properties panel, set the render resolution to 1280\(\times\)720, then
save the file.\par\vspace{0.8em}

{\normalsize\textbf{Benchmark Outcome}}\par\vspace{0.8em}
\small
This run fails, with \texttt{result.txt} reporting \texttt{0.0}. The required
parameter edit never becomes a completed save workflow, so the evaluator never
observes a valid 1280\(\times\)720 update in the submitted \texttt{.blend} file.

\vspace{0.8em}
{\normalsize\textbf{Phase-by-Phase Trajectory}}\par\vspace{0.8em}
\small

{\textbf{Phase 1: Uncertain File-Open Handling (steps 1--9).}} The run begins
with hesitation about the initial desktop state. The agent uses \texttt{Ctrl+O} and repeatedly clicks in the file-open
dialog without clearly committing to one navigation path. It partially infers
that the target file may already be visible or loaded, but it does not convert
that inference into a robust check. This early ambiguity matters because the
task should have moved quickly from file access into Output Properties, yet the
trajectory already begins to spend steps on state interpretation rather than
direct progress.

{\textbf{Phase 2: Catastrophic Recovery Failure (steps 10--12).}} The critical
failure occurs when the agent attempts to dismiss what it thinks is a blocking
window and sends \texttt{Alt+F4}. That closes Blender itself. The model notices
the mistake immediately in its own reasoning, but the damage is substantial:
from this point onward, the rollout is no longer a normal settings-edit task but
a recovery task. The agent then tries to relaunch Blender from the dock and via
the Activities search, but this recovery is slow and unstable. This is the main
turning point in the episode, because the agent loses the reliable application
context it needed for a simple property edit.

{\textbf{Phase 3: Reopening and Reacquiring the Project (roughly steps 20--34).}}
After relaunch attempts, the agent spends a long middle segment trying to reopen
\texttt{/home/user/Documents/scene.blend}. It alternates between double-clicking
file rows, clicking the Open button, pressing \texttt{Enter}, and finally using
\texttt{Ctrl+L} to type the full path into the file chooser. This phase is more
structured than the earlier opening attempts, and the typed absolute path is the
most reliable action in the whole trajectory. However, even after the project is
likely back on screen, the agent does not reestablish a clean internal model of
the Blender layout. The file-reopen problem is eventually reduced, but the
agent has already spent a large budget on state recovery.

{\textbf{Phase 4: Output-Properties Search by Coordinate Guessing (steps 35--43).}}
Once the project appears available again, the agent correctly recognizes that it
needs the Output Properties panel, but it never grounds the target icon or the
resolution fields reliably. Instead, it begins repeated coordinate-based clicks
on the right-side Properties area, trying several nearby y-positions that it
describes as possible tab icons. It also tries \texttt{F10} as a shortcut, but
this does not lead to a stable editable state either. The key weakness here is
that the agent has no fallback when visual icon targeting is uncertain. It keeps
sampling neighboring coordinates instead of switching to a deterministic
mechanism such as Blender's search, a structured menu path, or the embedded
Python interface.

{\textbf{Phase 5: Termination Without Parameter Edit.}} The trajectory ends
without any evidence that the width or height fields were actually changed to
1280 and 720, and without a successful save step that would propagate the edit
back into the \texttt{.blend} file. The final runtime state is effectively a
prolonged search loop inside Blender's UI rather than an edit-and-save workflow.

\vspace{0.8em}
{\normalsize\textbf{Evaluator Takeaway}}\par\vspace{0.8em}
\small
This is a clear control-and-recovery failure. The agent broadly knows what it needs to
accomplish, but it never maintains stable application state after closing
Blender and never finds a dependable path to the Output Properties controls. The dominant
failure mode is that the agent remains trapped in brittle GUI guessing on a task
where a deterministic fallback would have been much more reliable.
}

\clearpage
\twocolumn

\section{AI Assistants in Research or Writing}
\label{app:ai_assistants}

We used AI assistants, including ChatGPT and Cursor, during the preparation of
this work. Their use was limited to research, coding, and writing assistance:
improving grammar and clarity, suggesting wording alternatives, helping with
LaTeX editing, and assisting with code drafting, debugging, and result
organization. All benchmark design decisions, experimental protocols, task
definitions, analyses, and reported results were reviewed, verified, and
finalized by the authors. 

\section{Artifact Licensing, Privacy, and Content Review}
\label{app:artifact_license_privacy}

The released DeskCraft artifacts, including task definitions, evaluator code,
and supporting scripts, will be distributed with explicit license information;
the project code is released under the Apache License 2.0. Task assets are
synthetic, author-created, or derived from public sources that permit academic
use and redistribution, with attribution where applicable. Materials without
redistribution permission and proprietary practitioner-provided artifacts are
not released.

Before release, we checked task text, assets, and metadata for personally
identifying information and offensive content. Practitioner-seeded workflows
were abstracted with consent, and raw notes or proprietary artifacts were not
released. The final public files were manually reviewed by the authors.


%
\end{document}